\documentclass[letterpaper]{article} 
\usepackage{aaai25}  
\usepackage{times}  
\usepackage{helvet}  
\usepackage{courier}  
\usepackage[hyphens]{url}  
\usepackage{graphicx} 
\usepackage{natbib}  
\usepackage{caption} 
\usepackage{algorithm}
\usepackage{algorithmic}
\usepackage{subfigure}

\usepackage{newfloat}
\usepackage{listings}
\DeclareCaptionStyle{ruled}{labelfont=normalfont,labelsep=colon,strut=off} 
\floatstyle{ruled}
\newfloat{listing}{tb}{lst}{}
\floatname{listing}{Listing}
\usepackage{amsmath}
\DeclareMathOperator*{\argmin}{argmin}
\DeclareMathOperator*{\argmax}{argmax}
\usepackage{amssymb}
\usepackage{adjustbox}
\usepackage{multicol}
\usepackage{multirow}
\usepackage{comment}
\usepackage{tabularx}
\usepackage{colortbl}
\usepackage[capitalize,noabbrev]{cleveref}
\newcommand{\fhbig}{1.2cm}
\newcommand{\fwbig}{1.4cm}
\newcommand{\kernpic}[1]{\includegraphics[height=\fhbig,width=\fwbig]{#1}}

\newcommand{\name}{\texttt{KOBO}}
\usepackage[textsize=tiny]{todonotes}
\usepackage{enumitem}
\setlist[itemize]{align=parleft,left=0pt..1.5em}

\usepackage{newfloat}
\usepackage{listings}
\DeclareCaptionStyle{ruled}{labelfont=normalfont,labelsep=colon,strut=off} 
\floatstyle{ruled}
\newfloat{listing}{tb}{lst}{}
\floatname{listing}{Listing}
\title{Kernel Learning for Sample Constrained Black-Box Optimization}
\author{Rajalaxmi Rajagopalan, Yu-Lin Wei, Romit Roy Choudhury\\
}
\affiliations {
    Department of Electrical \& Computer Engineering\\
University of Illinois Urbana-Champaign\\
    \{rr30,yulinlw2,croy\}@illinois.edu
}
\vspace{-7.5ex}

\usepackage{bibentry}

\usepackage[utf8]{inputenc}

\begin{document}
\maketitle

\begin{abstract}
Black box optimization (BBO) focuses on optimizing unknown functions in high-dimensional spaces.
In many applications, sampling the unknown function is expensive, imposing a tight sample budget.
Ongoing work is making progress on reducing the sample budget by learning the shape/structure of the function, known as kernel learning.
We propose a new method to learn the kernel of a Gaussian Process.
Our idea is to create a continuous kernel space in the latent space of a variational autoencoder, and run an auxiliary optimization to identify the best kernel.
Results show that the proposed method, {\em Kernel Optimized Blackbox Optimization} ({\name}), outperforms state of the art by estimating the optimal at considerably lower sample budgets.
Results hold not only across synthetic benchmark functions but also in real applications.
We show that a hearing aid may be personalized with fewer audio queries to the user, or a generative model could converge to desirable images from limited user ratings. 
\end{abstract}

\section{Introduction}
\label{sec:intro}
Many problems involve the optimization of an {\em unknown} objective function. 
Examples include personalizing content $x$ to maximize a user's satisfaction $f(x)$, or training deep learning models with hyperparameters $x$ to maximize their performance $f(x)$.
Function $f(x)$ is unknown in these cases because it is embedded inside the human brain (for the case of personalization) or too complex to derive (for hyper-parameter tuning).
However, for any chosen sample $x_i$, the value of $f(x_i)$ can be evaluated. 
For hearing-aid personalization, say, evaluating the function would entail playing audio with some hearing-compensation filter $x_i$ and obtaining the audio clarity score $f(x_i)$ from the user.

\begin{figure}
\centerline{\includegraphics[width=\columnwidth]{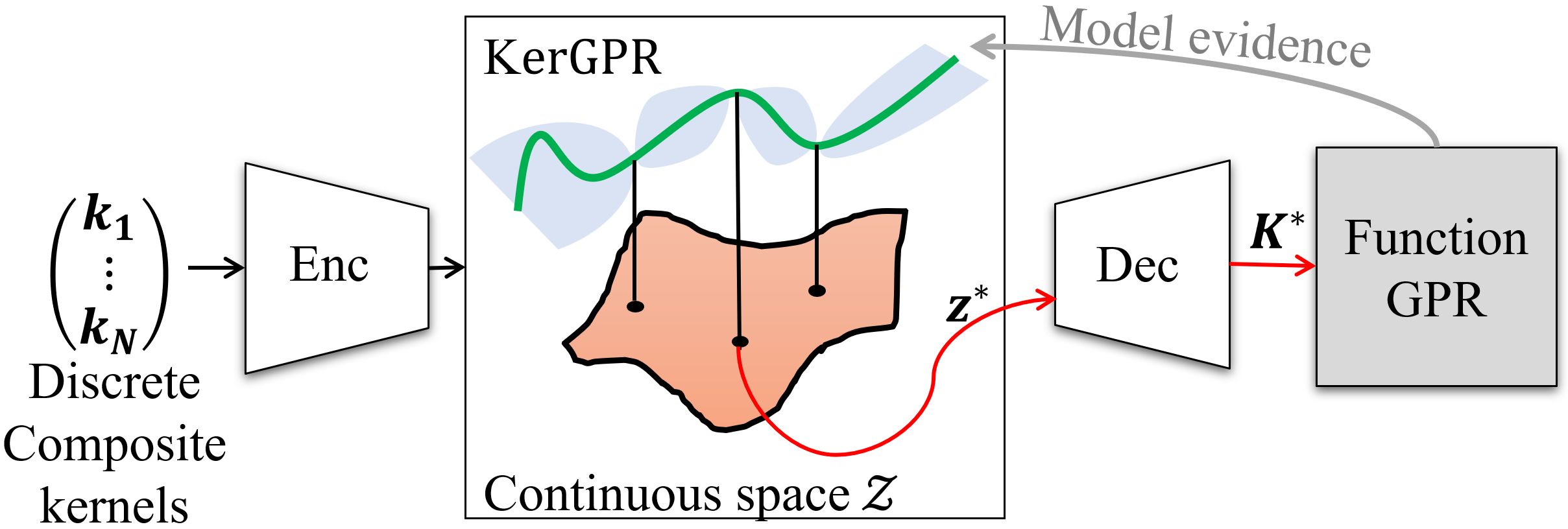}}
\caption{KerGPR in VAE latent space gives $K^*$ to $f$GPR}
\label{fig:kerVAE1}
\end{figure}

Bayesian methods like Gaussian Process Regression (GPR) are de facto approaches to black-box optimization (BBO). 
Using a set of function evaluations, conventional GPR \cite{frazier2018tutorial} learns a probabilistic surrogate model $\hat{f}(x)$ for $f(x)$. 
The optimum is estimated on this surrogate as $\hat{x}^* = \argmin -\hat{f}(x)$.
In most BBO problems, $f(x)$ is expensive to evaluate, hence a strict sample or query budget $B$ is of interest.
Techniques that lower this budget have garnered recent attention. 
One idea is to exploit domain knowledge about the rough shape of $f(x)$, i.e., select a GPR kernel that models this shape.
With humans, for example, $f(x)$ may have a staircase structure as they may not perceive differences in certain neighborhoods of $x$, but their ratings may change just outside that neighborhood. If GPR's surrogate model $\hat{f}(x)$ captures this staircase structure in its kernel, sample efficiency can improve.
However, in the absence of domain knowledge, {\em can the optimal GPR kernel $\mathbf{K}^*$ be learnt, using the same sample queries used to estimate $x^*$?}

A growing body of research \cite{grosse2012exploiting}\cite{kim2016scalable}\cite{teng2020scalable} is concentrating on kernel learning. 
One effective approach is Automatic Statistician (AS) \cite{duvenaud2013structure} where authors compose complex kernels by combining simple ones through a context-free grammar and design a search method over the countably infinite complex kernels (more in Section \ref{sec:related-work}). 
Subsequent improvements over AS have used \textit{Hellinger distance} as a measure of kernel similarity \cite{malkomes2016bayesian}. 
This similarity metric guides an optimization-based search over the space of composite kernels. 
To reduce search complexity, \cite{gardner2017discovering} exploits additive structures in the search space and employs MCMC methods to discover kernels.
All these techniques operate on discrete spaces, delivering a categorical composition of simple kernels.

A {\em continuous} space of kernels naturally lends itself to optimization methods. 
Our contribution is in designing such a continuous kernel space and running an auxiliary optimization on it to discover $\mathbf{K}^*$. 
To this end, we first synthesize many discrete kernels by adding or multiplying a set of simple ``basis'' kernels, and then use a variational autoencoder (VAE) to learn a low-dimensional continuous manifold for the discrete kernels. 
This manifold lives in the latent space of the VAE, as shown by the orange region in Figure \ref{fig:kerVAE1}.
Conventional optimization on this latent kernel space is difficult since we lack an objective function, but 
given a kernel, we can evaluate its effectiveness using {\em model evidence} (i.e., how well a given kernel agrees with  available samples from $f(x)$).
Thus, optimizing over the kernel space can also be designed as a blackbox optimization problem, hence 
we run a kernel space GPR {\em (KerGPR)} to output an optimal $\mathbf{K^*}$.
The main GPR module, {\em Function GPR}, uses $\mathbf{K}^*$ to model the surrogate function $\hat{f}(x)$, and queries the surrogate at more points. 
The new model evidence is then passed back to KerGPR (see Fig. \ref{fig:kerVAE1}) to update the optimal kernel.
The iteration terminates when the query budget is expended. {\em Function GPR} then outputs the minimum of the surrogate $\hat{f}(x)$.

Results show that {\name} consistently outperforms SOTA methods (namely MCMC \cite{gardner2017discovering}, BOMS \cite{malkomes2016bayesian}, and CKS \cite{duvenaud2013structure}) in terms of the number of queries needed to reach the optimal.
The gain from kernel learning is also significant compared to the best static kernels.
Experiments are reported across synthetic benchmark functions and from real-world audio experiments with $U$=$6$ users.
Volunteers were asked to rate the clarity of audio clips, and using $B \leq 25$ ratings, {\name} prescribed a personalized filter that maximized that user's personal satisfaction.
The performance gain is robust across extensive experiments, suggesting that {\name} could be deployed in real-world black-box applications where sample-budget is of prime concern.

\section{Problem Formulation and Background}

\subsection{Problem Formulation} 
Consider an {\em unknown} real-valued function $f: \mathcal{H} \to \mathbf{R}$ where $\mathcal{H} \subseteq \mathbf{R}^N$, $N \ge 500$. Let $x^*$ be the minimizer of $f(x)$. 
We aim to estimate $x^*$ using a budget of $B$ queries.
Thus, the optimization problem is,
\begin{equation}
\begin{aligned}
\argmin_{\hat{x} \in \mathcal{H}} \quad &||f(\hat{x})-f(x^*)||_2  \quad \quad \textrm{s.t.} \quad &Q \leq B
\end{aligned}
\label{eqn:opti-prob}
\end{equation}
where $Q$ is the number of times the objective function is evaluated/queried, and the sample budget is $B \ll N$. 
Function $f$ may be non-convex, may not have a closed-form expression, and its gradient is unavailable (hence, a black-box optimization problem).
Bayesian approaches like GPR suit such problems but require choosing a kernel to model the function structure; a poor choice incurs more queries for optimization.
Since queries can be expensive (e.g., when users need to answer many queries, or a NeuralNet needs re-training for each hyper-parameter configuration), lowering $Q$ is of growing interest.
Kernel learning aims to address this problem.

\subsection{Background: Gaussian Process Regression (GPR)}
Bayesian optimization (BO) \cite{frazier2018tutorial} broadly consists of two modules: 
(1) {\em Gaussian Process Regression} that learns a Gaussian posterior distribution of the likely values of $f(x)$ at any point of interest $x$. 
(2) {\em Acquisition function}, a sampling strategy that prescribes the point at which $f$ should be evaluated (or sampled) next. 
We briefly discuss GPR to motivate the kernel learning problem. 


\textbf{GPR Prior \& Posterior}: 
GPR generates a probabilistic surrogate model by defining a Gaussian distribution ($\mathcal{N}(\boldsymbol\mu,\mathbf{K})$) over infinite candidate functions. 
At initialization, i.e., when no function-sample are available, the prior distribution over the candidate functions is defined by $\boldsymbol\mu = \mathbf{0}$ and a covariance matrix $\mathbf{K}$.
This matrix is computed using a kernel function $k$ as, $\mathbf{K}_{ij}$=$k(x_i,x_j)$. 
The kernel essentially favors 
candidate functions that are similar to the kernel's own shape/structure; these candidates are assigned a higher likelihood.
An expert with domain knowledge about the structure of $f(x)$ can choose the kernel judiciously, resulting in better surrogate functions $\hat{f}(x)$.
Better $\hat{f}(x)$ will ultimately reduce the number of queries needed to optimize the objective $f(x)$.

\begin{equation}
\begin{aligned}
P(\mathcal{F}|\mathcal{X}) \sim \mathcal{N}(\mathcal{F}|\boldsymbol\mu,\mathbf{K})
\end{aligned}
\label{eqn:posterior}
\end{equation}
where, $\boldsymbol\mu$ = $\{\mu(x_1),\mu(x_2),\dots,\mu(x_K)\}$,  $\mathbf{K}_{ij}$=$k(x_i,x_j)$, and $k$ represents a kernel function.

\textbf{Predictions}: To make predictions $\hat{\mathcal{F}}=f(\hat{\mathcal{X}})$ at new points $\hat{\mathcal{X}}$, GPR uses the current posterior (Eqn. \ref{eqn:posterior}) to define
the conditional distribution of $\hat{\mathcal{F}}$ as: $P(\hat{\mathcal{F}}|\mathcal{F},\mathcal{X},\hat{\mathcal{X}}) \sim \mathcal{N}(\hat{\mathbf{K}}^T\mathbf{K}^{-1}\mathcal{F},\hat{\hat{\mathbf{K}}} - \hat{\mathbf{K}}^T\mathbf{K}^{-1}\hat{\mathbf{K}})$
The details and proof of all the above are clearly explained in \cite{wang2020intuitive}).

\subsection{Kernel Selection}

\begin{figure}
\centering
\renewcommand{\tabularxcolumn}[1]{>{\arraybackslash}m{#1}}
\begin{tabularx}{0.9\columnwidth}{XXXX}
  \kernpic{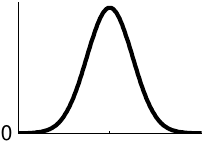} & \kernpic{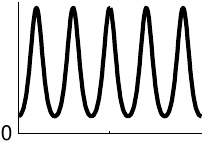}   
& \kernpic{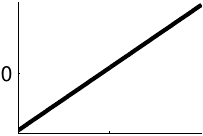} & \kernpic{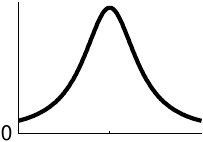}  
\\
  {\small Squared-exp (SE)} & {\small Periodic (PER)} 
& {\small Linear \newline (LIN)} & {\small Rational- quad. (RQ)}
\\
  \kernpic{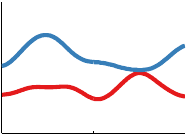} & \kernpic{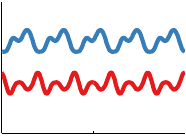}
& \kernpic{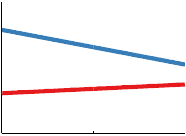} & \kernpic{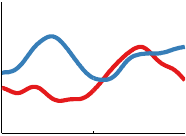}
\\
  {\small local \newline variation} & {\small repeating structure} 
& {\small linear \newline functions} & {\small multi-scale \phantom{iii}variation}
\end{tabularx}
\caption{Row 1 shows simple (or base) kernels \cite{duvenaud2013structure}. Row 2 shows corresponding surrogates drawn from a GP with the above kernel.}
\label{fig:basic_kernels}
\end{figure}

Kernels model the possible shape of the unknown function based on a set of observations $(\mathcal{X},\mathcal{F})$ of the unknown function.
Figure \ref{fig:basic_kernels} illustrates example kernels on the top row; the bottom row shows candidate functions that GPR can derive using the corresponding kernel. 
In general, a class of kernels $\mathcal{K}$ produces a family of (GPR) surrogates that fit the function observations $(\mathcal{X},\mathcal{F})$.
Of course, each surrogate is associated to a likelihood which can improve  with additional observations.

The goal of kernel selection is to select one kernel $\mathbf{K}^* \in \mathcal{K}$ that best explains the function observations $(\mathcal{X},\mathcal{F})$.  Let's denote $\mathcal{L}: \mathcal{K} \to \mathbf{R}$ to be a model evidence that measures how well a kernel $\mathbf{K}$ fits the observations. We assume that evaluating $\mathcal{L}(\mathbf{K})$ for all kernels in $\mathcal{K}$ is too expensive. The kernel selection problem is then,
\begin{equation}
\begin{aligned}
\mathbf{K}^* = \argmax_{\mathbf{K} \in \mathcal{K}} \mathcal{L}(\mathbf{K})
\end{aligned}
\label{eqn:ker-learn}
\end{equation}
This problem is difficult to optimize with Bayesian approaches when the kernel space $\mathcal{K}$ is discrete \cite{parker2014discrete}. 
This is because the function $\mathcal{L}(\mathbf{K})$ is only defined at the feasible points and cannot be queried arbitrarily. 
In contrast, it is possible to deduce a continuous function's behavior in a neighborhood of a point; in the discrete case, the behavior of the objective may change significantly as we move from one feasible point to another.
This motivates transforming the problem in Eqn. \ref{eqn:ker-learn} into a problem in continuous space on which optimization can be applied.

In this paper, the ``model evidence" $\mathcal{L}$ is chosen to be GPR posterior in Eqn. \ref{eqn:posterior} as it generates the surrogate that best describes the observations informed by the chosen kernel.
\begin{equation}
\begin{aligned}
\mathcal{L}(\mathbf{K}) = P(\mathcal{F}|\mathcal{X}, \mathbf{K})
\end{aligned}
\label{eqn:ker-evidence}
\end{equation}

\section{Kernel Learning in {\name}}

\textbf{Intuition and Overview:} 
Our prime objective is to create a continuous space $\mathcal{Z}$ corresponding to the discrete space $\mathcal{K}$, thus simplifying the optimization in Eqn. \ref{eqn:ker-learn}. We propose to achieve this using a Variational Autoencoder (VAE) which can take discrete inputs ($\mathbf{K} \in \mathcal{K}$) and learn (encoder) a continuous latent space $\mathcal{Z}$ from which the inputs can be faithfully reconstructed (decoder). 
If a large number of discrete kernels are created and represented sufficiently using a scheme that offers some notion of mutual similarity between kernels (i.e., representations defining a kernel space $\mathcal{K}$), then we expect the VAE to give us a continuous representation of such kernels $\mathcal{Z}$. This approach satisfies our objective since VAEs are expected to ensure \textit{continuity} and \textit{completeness} of their latent space, i.e., (i) two close points in the latent space cannot decode to completely different results, and (ii) a point sampled from the latent space must decode to a valid result. When the VAE is trained, we have successfully transformed the discrete optimization in $\mathcal{K}$ (Eqn. \ref{eqn:ker-evidence}) to an easier continuous one in $\mathcal{Z}$. 

\begin{figure}
\centering
{\includegraphics[width=1\columnwidth]{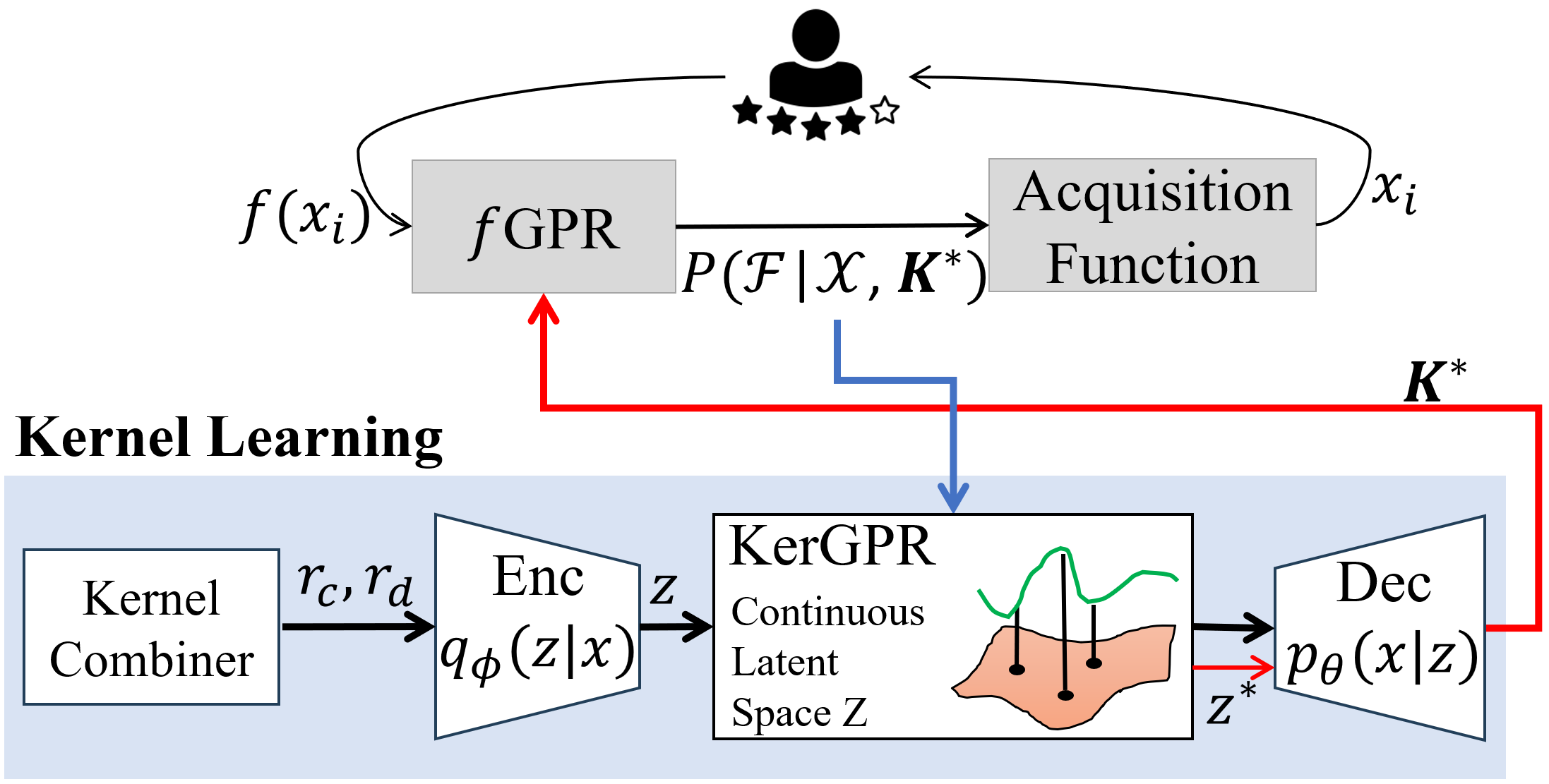}}
\caption{\small{System flow: \name\ iterates across \textit{$f$GPR} on top and \textit{{KerGPR}} below that runs in the KerVAE latent space. The blue arrow denotes the model evidence input to KerGPR, and the red arrow denotes the optimal kernel $\mathbf{K}^*$ supplied by KerGPR to $f$GPR.}}
\label{fig:kerGPR}
\end{figure}

Building on this intuition, {\name}'s kernel learner is composed of $3$ modules as shown in Figure \ref{fig:kerGPR}:

\begin{itemize}
    \item[(1)] \textit{\textbf{Kernel Combiner}} creates composite kernels $\mathbf{K} \in \mathcal{K}$ that form the discrete kernel space $\mathcal{K}$.
    \item[(2)] \textit{\textbf{Kernel Space Variational Autoencoder (KerVAE)}} trained on kernels generated by Kernel Combiner transforms discrete kernel space $\mathcal{K}$ to continuous space $\mathcal{Z}$.
    \item[(3)] \textit{\textbf{Kernel Space GPR (KerGPR)}}: Optimizes model evidence (Eqn. \ref{eqn:ker-evidence}) on $\mathcal{Z}$; this gives $z^*$ which decodes to the optimal kernel $\mathbf{K}^*$. 
\end{itemize}

Figure \ref{fig:kerGPR} connects all the modules to give a complete overview of {\name}.
The main objective function (Eqn. \ref{eqn:opti-prob}) is optimized with a \textbf{Function GPR ($f$GPR)}. $f$GPR uses a simple Square-Exponential (SE) kernel to obtain a batch of observations $\mathcal{D}_n$. 
The model evidence is then passed to the kernel learning pipeline.
The Kernel Combiner takes simple kernels and observations $\mathcal{D}_n$ as inputs, and outputs a discrete space of composite kernels $\mathbf{K}_{\mathcal{C}} \in \mathcal{K}$. 
This is guided by a context-free-grammar described later.
The KerVAE is trained on this discrete space and generates the corresponding continuous latent space $\mathcal{Z}$. 
KerGPR running on $\mathcal{Z}$ optimizes the model evidence (from $f$GPR) to find the optimal kernel $\mathbf{K}^*$ which is prescribed to $f$GPR. 
$f$GPR uses this $\mathbf{K}^*$ to obtain a new batch of observations $\mathcal{D}_{n+1}$ and update model evidence, which is again passed to the kernel learning pipeline.
The cycle iterates until $f$GPR has expended the sample budget $B$. At this point, $f$GPR outputs the minimum of its surrogate model.
The following discussions expand on Kernel Combiner, KerVAE, and KerGPR.

\subsection{Kernel Combiner}
Complex kernels $\mathbf{k}_{\mathcal{C}}$ can be expressed as operations on the context-free grammar of base kernels $\mathcal{B}$ \cite{hopcroft2001introduction,duvenaud2013structure}.
Given $\mathcal{B} = \{ A, B, C, D, E \}$, and a set of operators $\mathcal{O} = \{ \text{add}, \text{multiply}, \text{end},\dots \}$, the Kernel Combiner generates a composite kernel $\mathbf{k}_{\mathcal{C}}$ by drawing kernels from $\mathcal{B}$ and operators from $\mathcal{O}$ with probabilities $p_{\mathcal{B}}, p_{\mathcal{O}}$.
An example $\mathbf{k}_{\mathcal{C}} = A*C + B*D$.
In general, to form a kernel space $\mathcal{K}$, the Kernel Combiner develops a unique representation for each $\mathbf{k}_{\mathcal{C}}$.

\textbf{Grammar-based Representation}:
Given a composite kernel $\mathbf{k}_{\mathcal{C}}$, its grammar-based representation is a vector $r_c$, designed as follows.
Let $\mathbf{A}, \mathbf{B}, \mathbf{C}, \mathbf{D}, \mathbf{E}$ be five base kernels in $\mathcal{B}$. 
These are simple kernels like Square-exponential, Periodic, Rational Quadratic, etc. 
Any composite kernel $\mathbf{k}_{\mathcal{C}}$ created from the base kernels is expressed in the form of Eqn. \ref{eqn:frac_code}. 
The code $r_c$ is then the vector of indices, i.e., 
$r_c = [a_1,b_1,c_1,d_1,e_1,a_2,b_2,c_2,d_2,e_2,a_3,b_3,c_3,d_3,e_3]$.

\begin{equation}
\begin{aligned}
 \mathbf{k}_{\mathcal{C}} &= \mathbf{A}^{a_1} * \mathbf{B}^{b_1} * \mathbf{C}^{c_1} * \mathbf{D}^{d_1} * \mathbf{E}^{e_1} \\  
&+ \mathbf{A}^{a_2} * \mathbf{B}^{b_2} * \mathbf{C}^{c_2} * \mathbf{D}^{d_2} * \mathbf{E}^{e_2} \\
&+ \mathbf{A}^{a_3} * \mathbf{B}^{b_3} * \mathbf{C}^{c_3} * \mathbf{D}^{d_3} * \mathbf{E}^{e_3} \quad \dots
   \end{aligned}
\label{eqn:frac_code}
\end{equation}
If a composite kernel is, say, $\mathbf{k}_{\mathcal{C}}' = \mathbf{A}^{2} * \mathbf{E} + \mathbf{C} * \mathbf{D}$, then its Grammar-based representation would be $r_c' = [2,0,0,0,1,0,0,1,1,0,0,0,0,0,0]$. 
Note: elements of the code vectors can also be fractions.

This encoding scheme has two advantages: (1)  
Each code $r_c$ preserves its composition, i.e., given the code vector $r_c$, the base kernels and the operators used to construct $\mathbf{k}_{\mathcal{C}}$ can be interpreted.
(2) The code space is continuous, hence, a code $r_c'' = [2,1,0,0,1,0,0,1,1,0,0,0,0,0,0]$ -- which is only a flip of the second element in $r_c'$ -- results in $\mathbf{k}_{\mathcal{C}}'' = \mathbf{A}^2 * \mathbf{B} * \mathbf{E} + \mathbf{C} * \mathbf{D}$.
In general, a small change in the code produces a small modification to the kernel composition (which makes the VAE's task: $\mathcal{K} \to \mathcal{Z}$ easier).
Past work \cite{garrido2020dealing}\cite{lu2018structured} have used one-hot encoding to represent kernel matrices, however, 
such one-hot schemes suffer from the lack of continuity (as shown in Figure 5(a) and (b) in the Appendix).

However, the context-free grammar $r_c$ does not encode any information from the objective function to be modeled by the complex kernels. 
We add this to the representation next.

\textbf{Data-based Representation}:
Given the available function observations $(\mathcal{X},\mathcal{F})$, for each $\mathbf{k}_{\mathcal{C}}$, we compute the ``distances'' between its GPR covariance matrix $M_{\mathcal{C}}$ and the covariance matrix of each base kernel, $M_{b \in \mathcal{B}}$ (the covariance matrix is computed as $\mathbf{K}_{ij} = k(x_i,x_j)$, $k(\cdot)$ is the kernel function and $x_i,x_j$ are any two observations).
We use the Forbenius norm to compute the matrix distances. 
This representation is denoted as $r_d \in \mathcal{R}^{|\mathcal{B}|}$. 
\begin{equation}
\begin{aligned}
r_d = ||M_{\mathcal{C}} - M_{b \in \mathcal{B}}||_F
\end{aligned}
\label{eqn:ker-norm}
\end{equation}
The function information is now encoded in $r_d$ as the covariance matrix is computed using the kernel and the function observations/samples. 
Kernels that model the function's data similarly need to be in the same neighborhood so that the model evidence is sufficiently smooth. Figure 5(c) and (d) (in the Appendix) shows the advantage of encoding the objective function's data in the kernel code.

Thus, the kernel space $\mathcal{K}$ consisting of complex kernels is a subset of the space of all positive semi-definite (PSD) matrices $\mathbf{S}$, $\mathcal{K} \subseteq \mathbf{S}$. By restricting our search to $\mathcal{K}$-- not $\mathbf{S}$ -- the kernels generated through context-free grammar compositions, we can scale the GPR covariance matrix as new function observations arrive (simple kernel functions have closed-form expressions that are expanded to complex kernels for any number of observations). Therefore, our kernel learning problem in Eqn. \ref{eqn:ker-learn} becomes a kernel selection problem. The final representation of a composite kernel $\mathbf{k}_{\mathcal{C}}$ is $r=[r_c,r_d]$. This is used to train the KerVAE to generate the continuous kernel space $\mathcal{Z}$.

\subsection{Kernel Space Variational Autoencoder (KerVAE) and GPR (KerGPR)}

The KerVAE learns a continuous latent space $\mathcal{Z}$ of the discrete kernel space $\mathcal{K}$ and has two main components: (1) a probabilistic encoder that models $q_{\phi}(z|x) \sim p_{\theta}(x|z)p(z)$ parameterized by $\phi$ where $p(z)$ is the prior over the latent space, and (2) a decoder that models the likelihood $p_{\theta}(x|z)$ parameterized by $\theta$. 
The parameters of $q_{\phi}(z|x), p_{\theta}(x|z)$ are optimized by joint maximization of the ELBO loss \cite{kingma2013auto},
\begin{equation}
\begin{aligned}
\mathcal{L}(\phi,\theta,x) = \mathbf{E}_{q_{\phi}(z|x)}[\text{log}p_{\theta}(x,z) - \text{log}q_{\phi}(z|x)]
\end{aligned}
\label{eqn:ker-vae}
\end{equation}
The KerVAE is re-trained after accumulating $v$ function observations as  
the data representation $r_d$ is re-computed for every new set of observations $\mathcal{D}_v = (\mathcal{X}_v,\mathcal{F}_v)$. 
Once KerVAE is trained, KerGPR is used to determine the optimal kernel $z^* \in \mathcal{Z}$.
The KerVAE decoder decodes $z^*$ to $r_c^*$ and $r_c^*$ is easily mapped to $\mathbf{K}^* \in \mathcal{K}$ due to the grammar's interpretability.

\textbf{KerGPR:}
Optimizing on KerVAE's latent kernel space $\mathcal{Z}$ is also a black-box problem (similar to Eqn. \ref{eqn:opti-prob}) because the objective can {\em only} be evaluated for a given kernel $k$ (i.e., by first decoding a given $z$ to $k$, and computing $k$'s model evidence $\mathcal{L}(\mathbf{K})$ in Eqn. \ref{eqn:ker-evidence}).
We use GPR to find $z^*$ in the latent space and decode to the optimal kernel $\mathbf{K}^*$. Thus, our optimization objective is:
\begin{equation}
\begin{aligned}
\mathbf{K}^* = \text{Dec}(\argmax_{z \in \mathcal{Z}} P(\mathcal{F}|\mathcal{X}, \text{Dec}(z)))
\end{aligned}
\label{eqn:ker-gpr}
\end{equation}

where, Dec(.) is the KerVAE decoder that maps a point from $\mathcal{Z}$ to the $\mathcal{K}$ space.
We use the simple \texttt{SE} kernel in KerGPR as it does not benefit from recursive kernel learning (explained in Technical Appendix). 
The optimal kernel $\mathbf{K}^*$ is then used by Function GPR ($f$GPR) --- the GPR posterior in Eqn. \ref{eqn:posterior} --- to generate surrogates that closely model the unknown function structure.

\begin{figure*}
\centering
\hfill%
{\includegraphics[width=0.19\columnwidth]{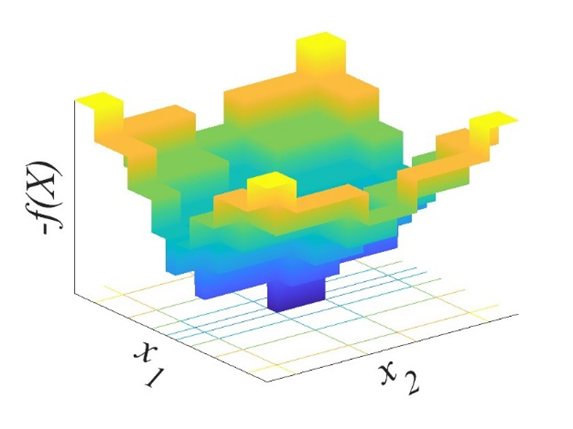}}
\hfill%
\hfill%
{\includegraphics[width=0.19\columnwidth]{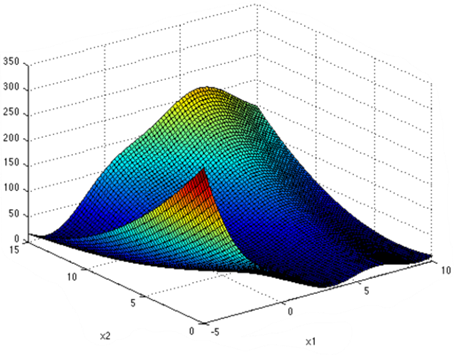}}
\hfill%
\hfill%
{\includegraphics[width=0.19\columnwidth]{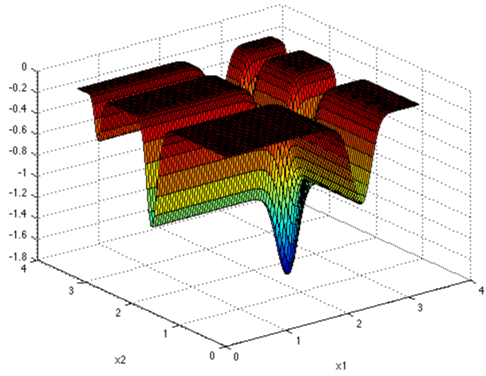}}
\hfill%
\hfill%
\hfill
\\
{\includegraphics[width=0.32\textwidth]{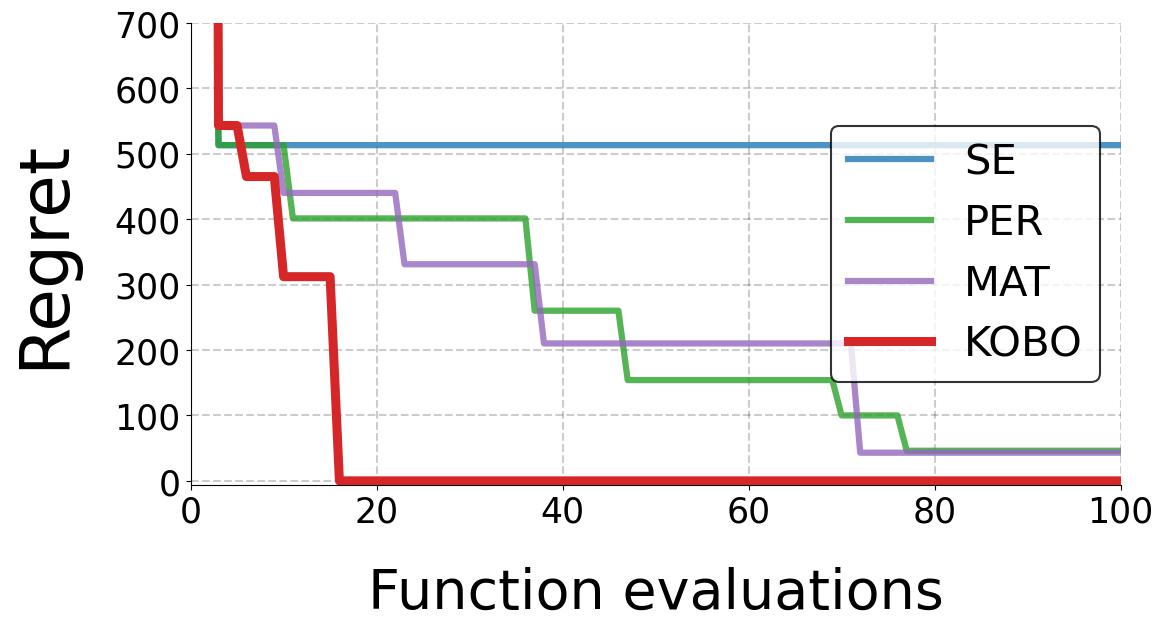}}
{\includegraphics[width=0.32\textwidth]{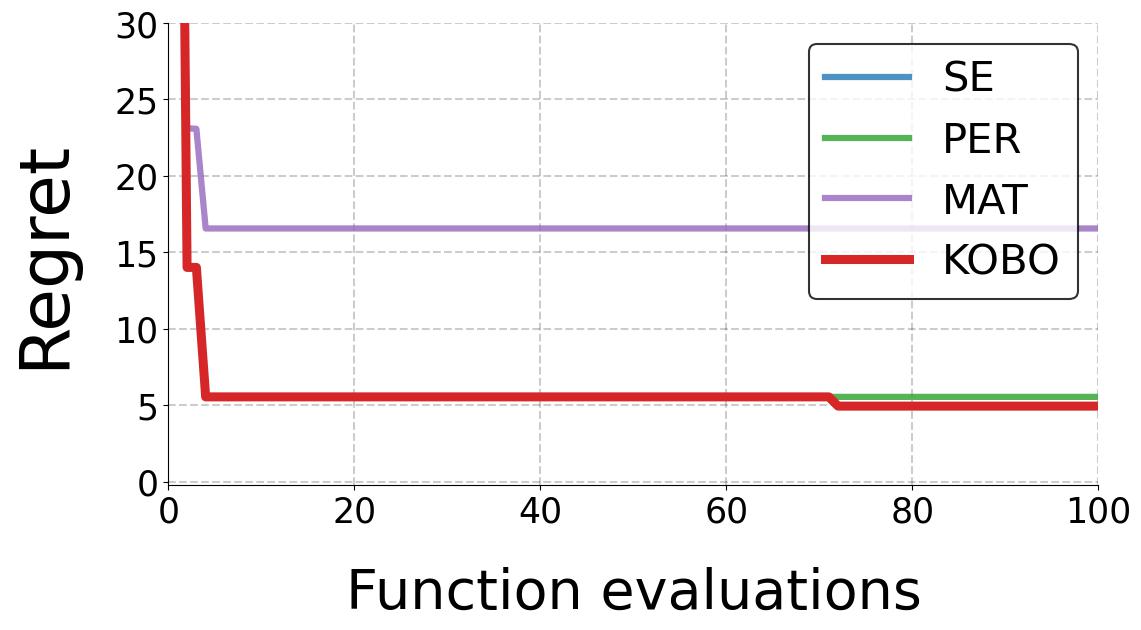}}
{\includegraphics[width=0.32\textwidth]{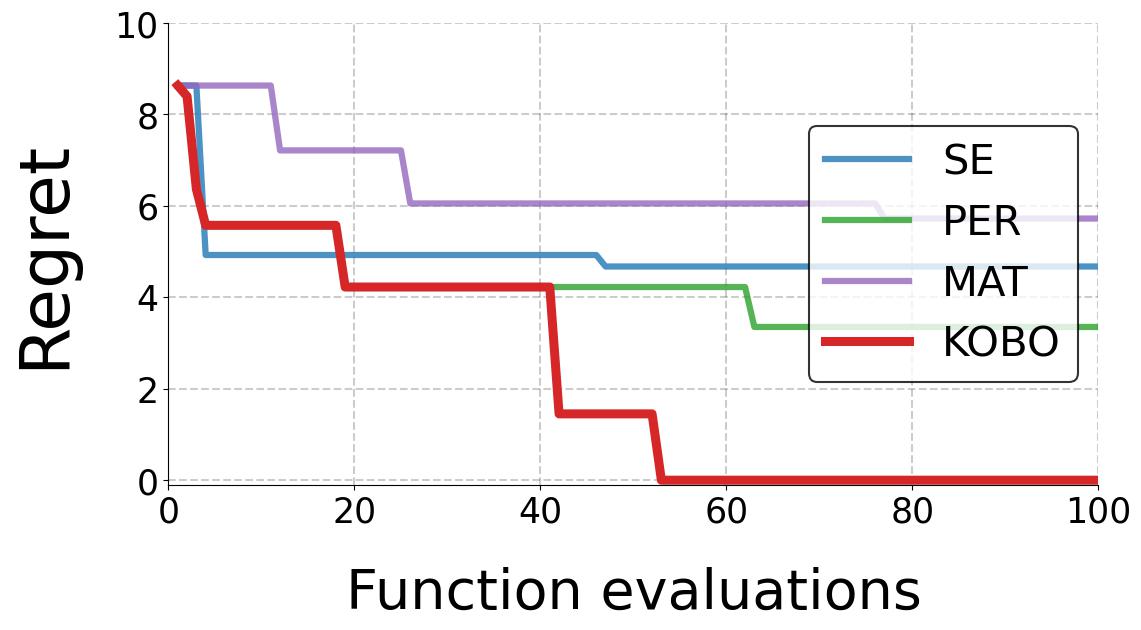}}
\caption{Comparison of {\name} and conventional BO using SE, PER, RQ, and Mat\'{e}rn kernels (Bottom Row) for (a) Staircase, (b) Smooth Branin, and, (c) Periodic Michalewicz functions (Top Row).} %
\label{fig:R2}
\end{figure*}

\begin{table}
    \centering
    \begin{adjustbox}{width=\columnwidth}
    \begin{tabular}{|c|c|c|c|c|c|}
    \hline
    Q & $f_1(x) \sim A*A*B+C$ & $f_2(x) \sim C+D$& $f_3(x) \sim D*B+D$\\
    \hline
    $5$ & $A*B$ & $A$& $A$\\
    \hline
    $10$ & $A*B+C$ & $A+D$& $A*B*D+D$\\
    \hline
    $15$ & $A*A*B+D$ & $A*C+D$& $A*B+D$\\
    \hline
    $20$ & $A*B+C*D$& $A*C+D$& $B*D+D$\\
    \hline
    $25$ & $A*A*B+C*D$ & $A*C+D$& $B*D+D$\\
    \hline
    \end{tabular}
    \end{adjustbox}
    \caption{{\name} learning the ground truth kernel. $\{A,B,C,D,E\} = \{\text{SE},\text{PER},\text{RQ},\text{MAT},\text{LIN}\}$}
    \label{tab:synthetic}
    
\end{table}

\section{Evaluation and Results}

\subsection{{\name} Versus Simple Kernels}
\textbf{Metric:} We use the metric of \textbf{\texttt{Regret}}, defined as the difference between the predicted and true minimum, $({f}(\hat{x}^*) - f(x^*))$. 
We compare {\name}'s regret against $5$ popular base kernels $\mathcal{B} = \{\text{SE},\text{PER},\text{RQ},\text{MAT},\text{LIN}\}$ which respectively denote Square-Exponential (SE), Periodic (PER), Rational Quadratic (RQ), Mat\'{e}rn (MAT), and Linear (LIN) kernels. 
All {\name} experiments are initialized with the SE kernel.

\textbf{Synthetic Baseline Functions:} We report results from $3$ types of synthetic functions $f(x)$ that are popular benchmarks \cite{kim2020benchmark} for black-box optimization:
$\blacksquare$ \texttt{Staircase} functions in $N=2000$ dimensions; they exhibit non-smooth structures \cite{Al-Roomi2015}. 
$\blacksquare$ \texttt{Smooth} benchmark functions such as \texttt{BRANIN} commonly used in Bayesian optimization research~\cite{ssurjano}. 
$\blacksquare$ \texttt{Periodic} functions such as \texttt{MICHALEWICZ} that exhibit repetitions in their shape \cite{ssurjano}.
The top row of Figure \ref{fig:R2} visualizes these functions (more details on evaluation parameters in the Technical Appendix).
All reported results are an average of $10$ runs.

\textbf{Results:} Figures \ref{fig:R2}(bottom row) plots \texttt{Regret} for the \texttt{Staircase}, \texttt{Smooth(BRANIN)}, and the \texttt{MICHALEWICZ} functions, respectively.
{\name} minimizes \texttt{Regret} at significantly fewer function evaluations (or samples), especially for \texttt{Staircase} and \texttt{MICHALEWICZ}.
For a smooth function like \texttt{BRANIN}, {\name}'s gain is understandably less since the $SE$ and $PER$ kernels naturally fit the smooth shape.
When real world functions exhibit complex (non-smooth) structures and when function evaluations are expensive, {\name}'s advantage is desirable.

\subsection{{\name} Versus SOTA Baselines} 

\textbf{Another Metric:} 
Since {\name} learns the kernel in the latent space, we will use \texttt{Model Evidence} in addition to \texttt{Regret}. 
\texttt{Model Evidence} is the normalized probability of generating the observed data $\mathcal{D}$ given a kernel model  $\mathbf{K}$, i.e., $\text{log}(P(\mathbf{f}|\mathcal{X}, \mathbf{K}))/|\mathcal{D}|$ \cite{malkomes2016bayesian}. 
Computing the exact model evidence is generally intractable in GPs \cite{rasmussen2006gaussian}\cite{mackay1998introduction}. We use the Bayesian Information Criterion (BIC) to approximate the model evidence as $\text{log}(P(\mathbf{f}|\mathcal{X}, \mathbf{K})) = -\frac{1}{2}\mathbf{f}^T\mathbf{K}^{-1}\mathbf{f} - \frac{1}{2}\text{log}((2\pi)^N |\mathbf{K}|)$, where $N$ is the dimensions of the input space $\mathcal{H} \subseteq \mathbf{R}^N$.

We will plot \texttt{Regret} against the number of ``Function Evaluations'' (on the X axis), but for \texttt{Model Evidence}, we will plot it against the number of ``Latent Evaluations".
Recall that \texttt{Model Evidence} is the metric used in the latent space of KerVAE to find the ``best" kernel $\mathbf{K}^*$. 
Hence ``Latent Evaluations" denotes the number of latent space samples $z$ and the corresponding kernels $\text{Dec}(z) = \mathbf{K}$ sampled by KerGPR to find $\mathbf{K}^*$. 
This reflects the computation overhead of {\name}.

\textbf{SOTA Baselines} (details in Technical Appendix): \\
\textbf{(1) MCMC:}
The MCMC kernel search \cite{gardner2017discovering,abdessalem2017automatic} applies the Metropolis-Hastings algorithm \cite{gardner2017discovering} on the space of composite kernels $\mathbf{k}_{\mathcal{C}}$, using model evidence as the function.
The proposal distribution is defined as: 
given a kernel $\mathbf{k}$, it is either added or multiplied to a kernel from $\mathcal{B}$ (chosen with $p$). 

\textbf{(2) CKS:}
The Automatic Statistician/Compositional Kernel Search (CKS) \cite{duvenaud2013structure} method takes advantage of the fact that complex kernels are generated as context-free grammar compositions of positive semi-definite matrices (closed under addition and multiplication); the kernel selection is then a tree search guided by model evidence. 
CKS searches over the discrete kernel space $\mathcal{K}$ using a greedy strategy that, at each iteration, chooses the kernel with the highest model evidence. This kernel is then expanded by composition to a set of new kernels. The search process repeats on this expanded list.   

\textbf{(3) BOMS:} The Bayesian Optimization for Model Search (BOMS) \cite{malkomes2016bayesian}, unlike CKS' greedy strategy, is a meta-learning technique, which, conditioned on observations $\mathcal{D}$, establishes similarities among the kernels in $\mathcal{K}$, i.e., BOMS constructs a kernel between the kernels (``kernel kernel”). Like {\name}, BOMS performs BO in $\mathcal{K}$ by defining a Gaussian distribution: $P(g) = \mathcal{N}(g;\mu_g,\mathbf{K}_g)$, where $g$ is the model evidence, $\mu_g$ is the mean, and $\mathbf{K}_g$ is the covariance (defined by "kernel kernel" function). $\mathbf{K}_g$ is constructed by defining a heuristic similarity measure between two kernels: {\em Hellinger distance}. 

\begin{figure*}[!htbp]
\centering
{\includegraphics[width=0.65\columnwidth]{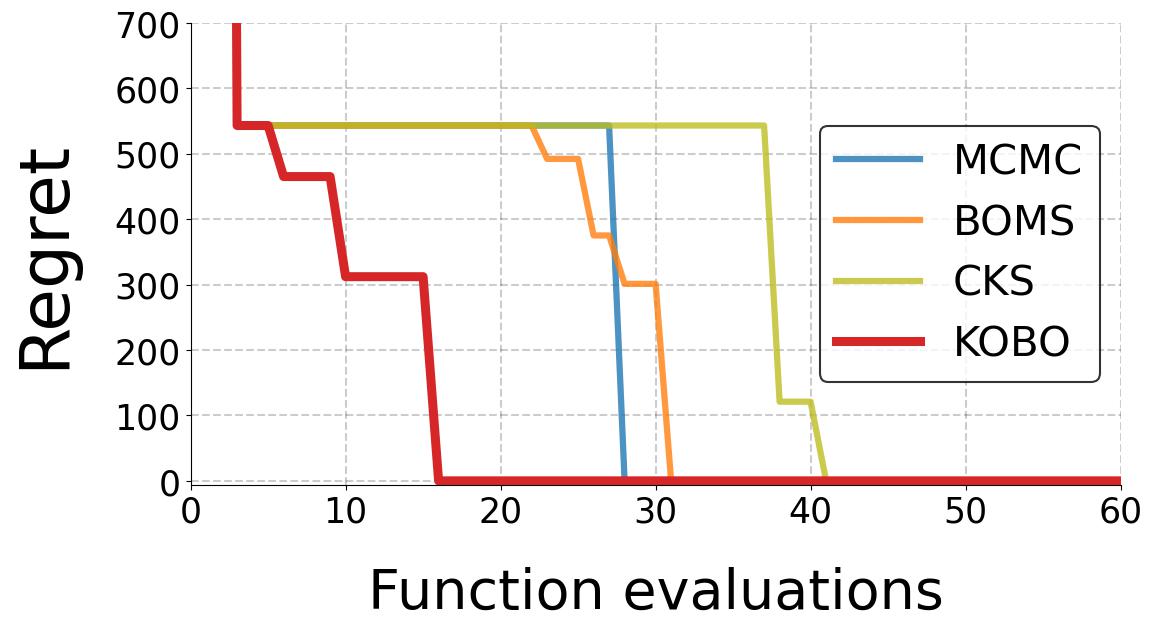}}
{\includegraphics[width=0.65\columnwidth]{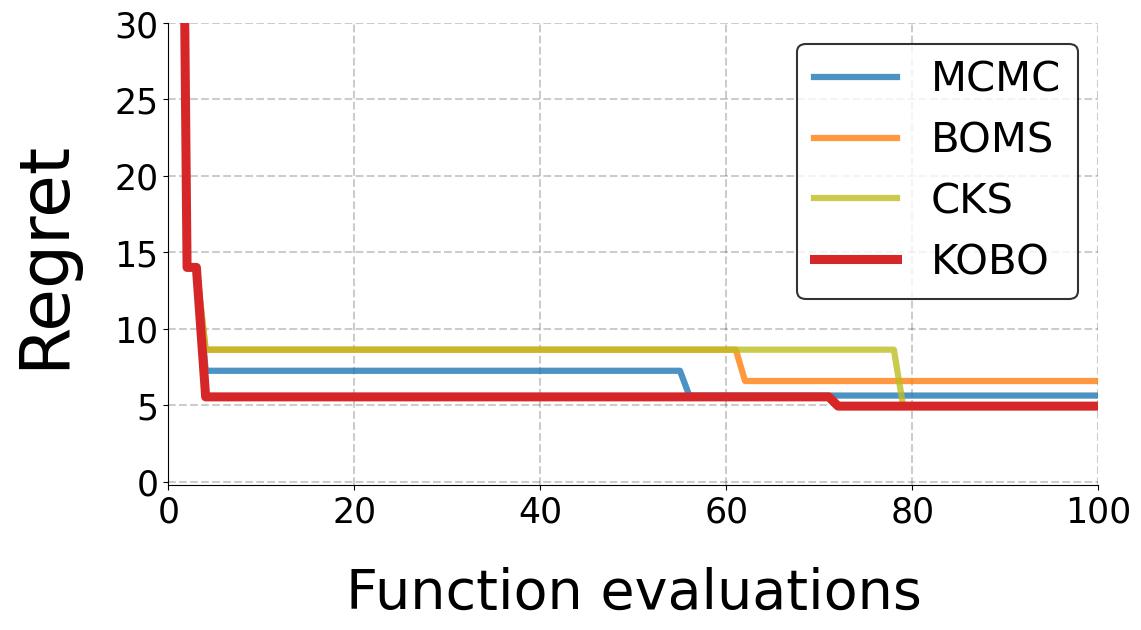}}
{\includegraphics[width=0.65\columnwidth]{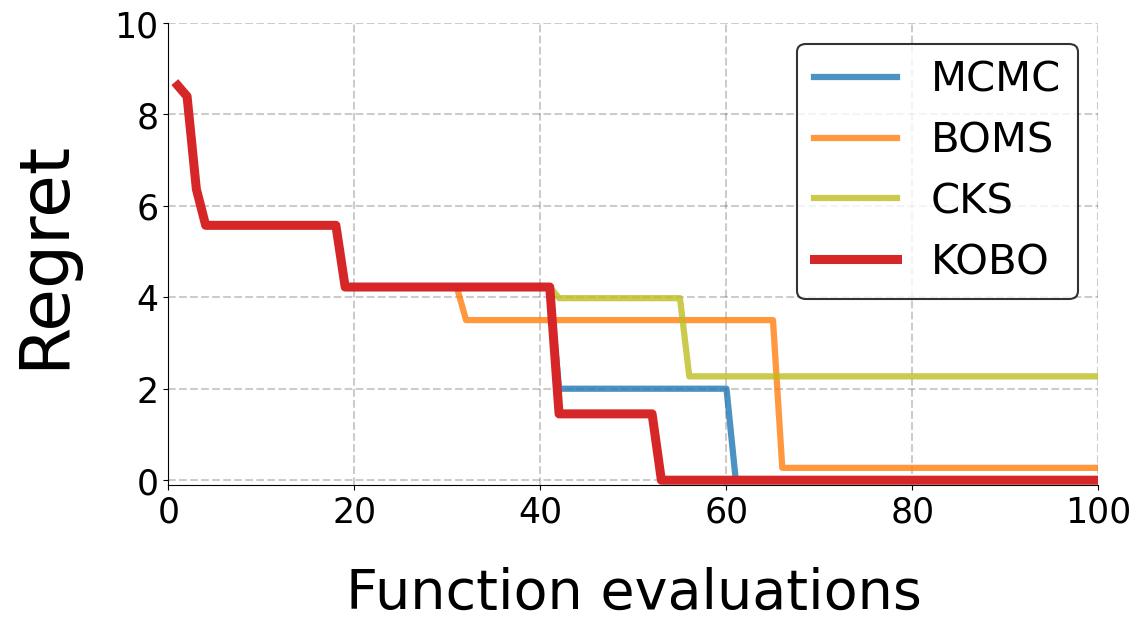}}
{\includegraphics[width=0.65\columnwidth]{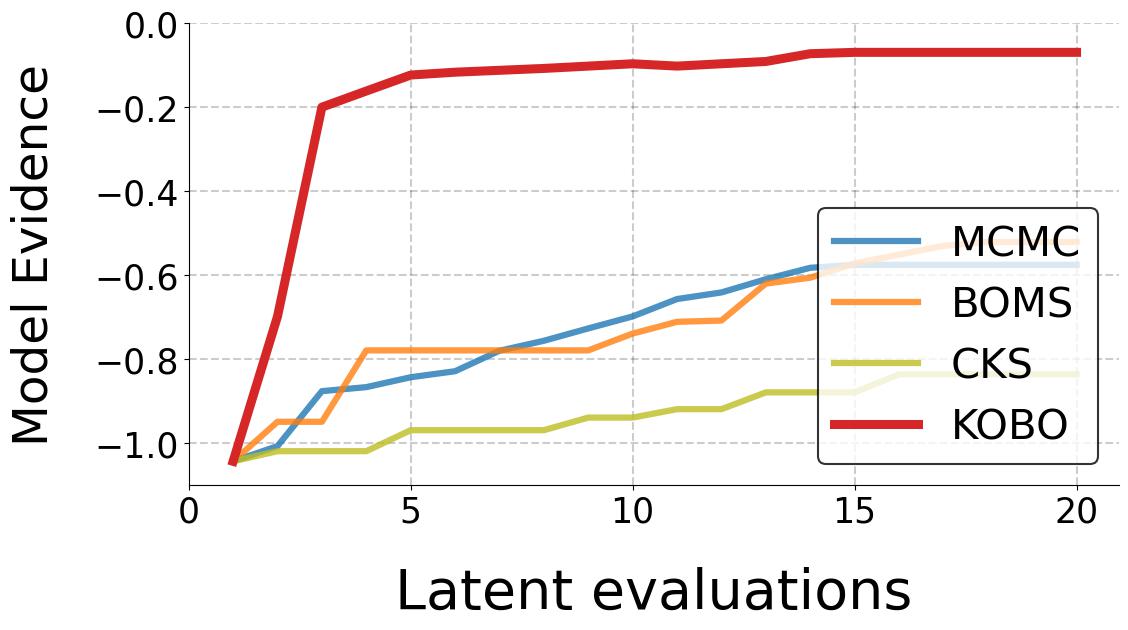}}
{\includegraphics[width=0.65\columnwidth]{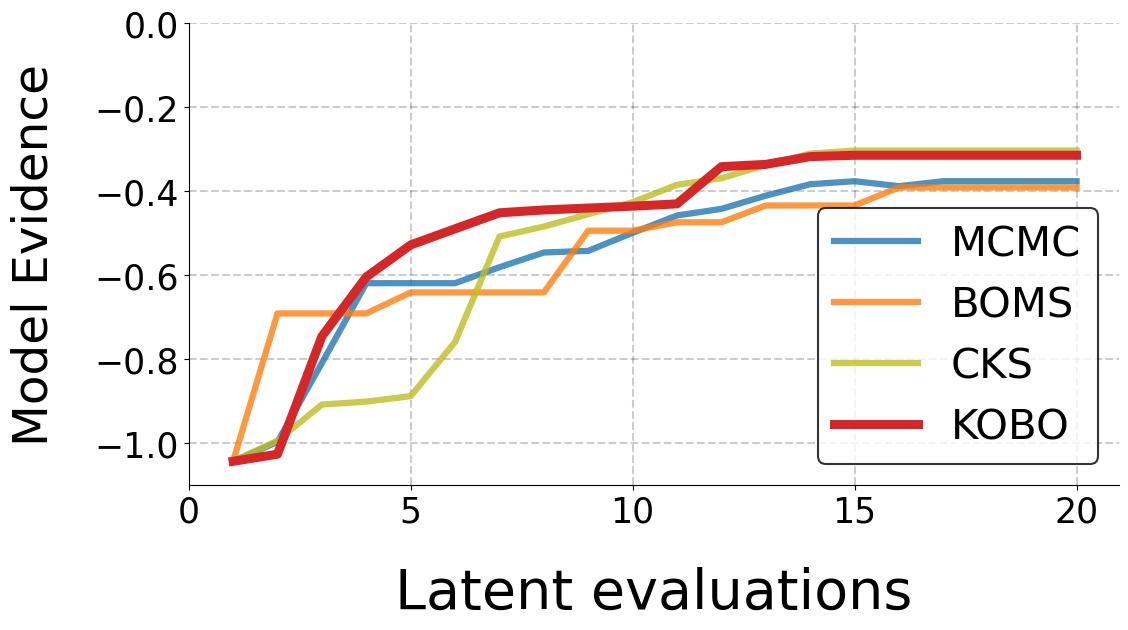}}
{\includegraphics[width=0.65\columnwidth]{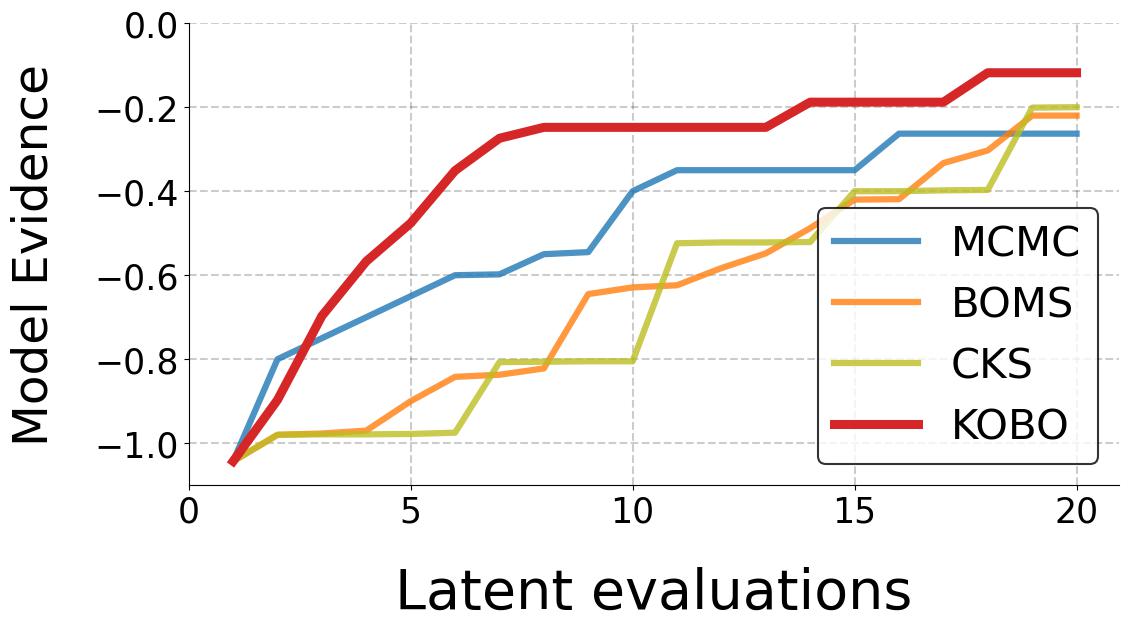}}
\caption{\textit{Comparison of {\name}, MCMC, CKS, and BOMS for Staircase (Col 1), Branin (Col 2), and Michalewiez (Col 3) functions:} (Row 1) \texttt{Regret} (Row 2) \texttt{Model Evidence}.}
\label{fig:R1}
\end{figure*}

\textbf{Results:}
Figure \ref{fig:R1}(Row 1) shows that {\name} lowers \texttt{Regret} faster than all SOTA baselines for the three benchmark functions. 
For \texttt{Staircase}, {\name} attains the global minimum in about $17$ function evaluations in contrast to MCMC, which incurs $28$, BOMS $32$, and CKS $43$. 
For \texttt{Michalewiez}, {\name} attains the minimum in about $10$ fewer samples than MCMC. While BOMS and CKS do not attain the minimum but get close to it. 
However, {\name}'s performance gain is not as pronounced for \texttt{Branin} due to its smooth structure as evidenced in Figure \ref{fig:R2}.

Figure \ref{fig:R1}(Row 2) compares \texttt{Model Evidence} for the same benchmarks.
For \texttt{Staircase}, {\name}'s KerGPR achieves significantly higher \texttt{Model Evidence} in $20$ iterations compared to MCMC, i.e., {\name}'s optimal kernel $\mathbf{K}^*$ better explains the observed data. For \texttt{Branin}, {\name} can match the \texttt{Model Evidence} of baselines, and performs modestly better for \texttt{Michalewiez}. 
The results illustrate that {\name}’s performance is superior because a continuous search space learned by KerVAE simplifies the KerGPR optimization to determine $\mathbf{K}^*$, implying that KOBO presents an efficient search of the discrete kernel space $\mathcal{K}$ in contrast to sub-optimal search techniques like greedy search (CKS), or heuristic metrics for kernel optimization (BOMS).

\subsection{Is $K^*$ indeed learning the structure of $f(x)$?}
\label{sec:struct}

\textbf{Synthetic Functions:} If we knew the objective function $f(x)$, we could verify if $\mathbf{K}^*$ has learnt its structure.
To test this, we sample a function from a GP with a known kernel $K^+$ and pretend that to be $f(x)$; we check if {\name}'s $\mathbf{K}^*$ converges to $K^+$.
Table \ref{tab:synthetic} reports results from $3$ $N$-dimensional synthetic functions, shown in the top row ($N=2000$).
These synthetic objective functions were sampled from a GP that uses different but {\em known} kernels. 
The subsequent rows show KerVAE's learnt kernel after $Q$ observations/queries.
With more $Q$, KerGPR closely matches the ground truth kernel.

\begin{figure*}
{\includegraphics[width=0.63\columnwidth]{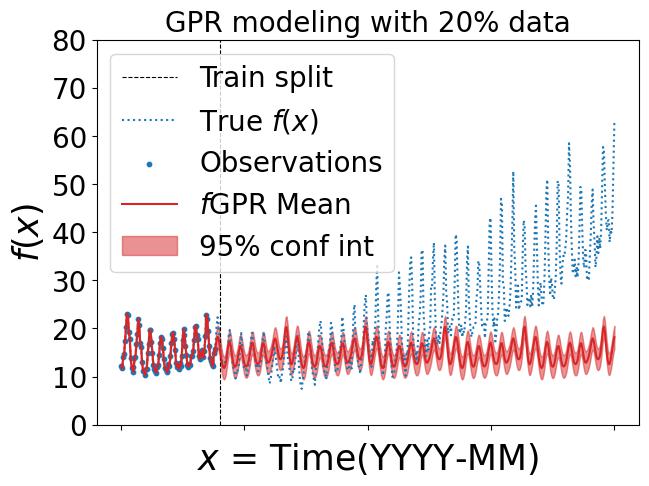}}
\hfill
{\includegraphics[width=0.63\columnwidth]{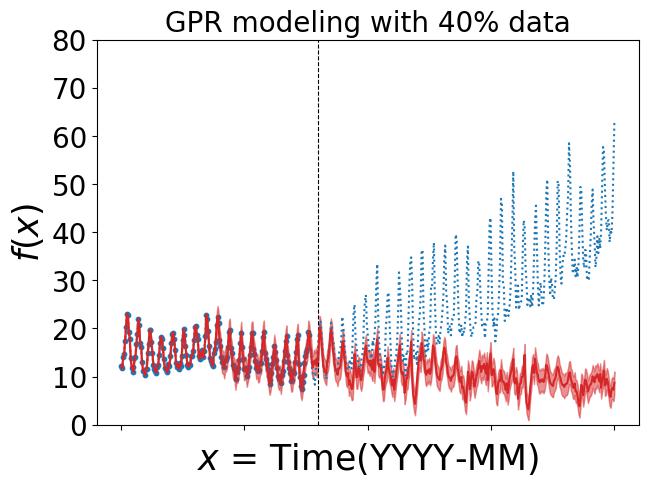}}
\hfill
{\includegraphics[width=0.63\columnwidth]{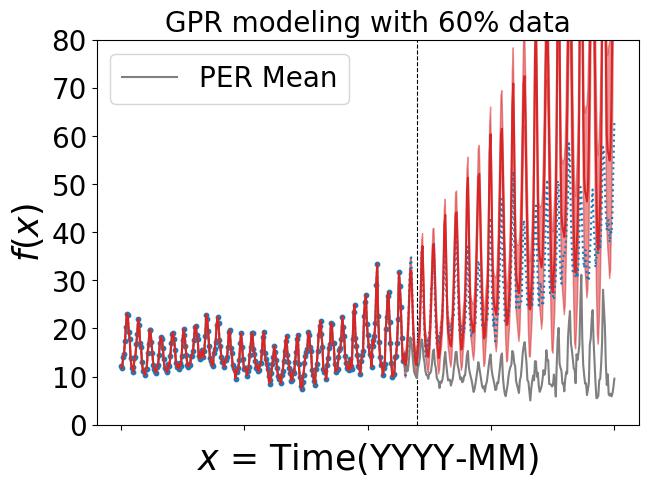}}
\caption{The blue curve is real-world $CO_2$ emissions data from \cite{thoning1989atmospheric}. The red curve is {\name}'s prediction of the blue curve after observing (a) 20\% (b) 40\% (c) 60\% of the blue data.}
\label{fig:R5}
\end{figure*}

\textbf{Learning Real-world $CO_2$ Emission Data:}
Figure \ref{fig:R5}'s blue curve plots real $CO_2$ emissions data over time \cite{thoning1989atmospheric}.
We test if {\name}'s $\mathbf{K}^*$ can learn the structure of this blue curve from partial data.
Figure \ref{fig:R5}(a,b,c) show results when {\name} 
has observed the first $20\%$, $40\%$, and $60\%$ of the data, respectively.
With the first $20\%$, $\mathbf{K}^* = \text{SE} * \text{PER} + \text{RQ}$, hence {\name}'s red curve captures the periodic structure of the early data.
When the first $40\%$ of the data is observed, {\name} captures the downward linear trend of the $CO_2$ data  resulting in $\mathbf{K}^* = \text{SE} * \text{PER} + \text{PER} + \text{LIN}$. 
With $60\%$ of the data, $\mathbf{K}^* = \text{SE} * \text{PER} * \text{RQ} + \text{PER} * \text{LIN} + \text{LIN}$ models the interplay between the function's periodic structure and linear trends. 
A conventional Periodic kernel (PER), shown in black in Figure \ref{fig:R5}(c), is only able to capture the periodic structure, not the upward linear trend, even with $60\%$ of the data.

\subsection{User Experiments: (1) Audio Personalization, and (2) Image Recommendation}
\label{sec:audio}

\textbf{(1) Audio:} We apply {\name} to audio personalization for real volunteers.
We deliberately corrupt audio played to the user with the aim of helping the user pick a filter $h^*$ that cancels the effect of the corruption -- equalization -- and recovers the original audio; hence maximizing the user's audio satisfaction. 
Therefore, a GPR employed in the space of all audio filters $\mathcal{H}$, maximizes the user's satisfaction $f(h)$ at $h^*$. 
At each iteration, the corrupted audio is filtered with a new $h'$ (recommended by GPR) and played to the user. 
The user's rating ($0$ to $10$) of the perceived audio quality serves as the function observations $f(h')$. 
User feedback is finite and the frequency selective nature of human hearing \cite{pitch} makes optimizing $f(h), h \in \mathcal{R}^{4000}$, well suited for kernel learning methods like {\name}. 

We invited $6$ volunteers to two experiment sessions.
In the first, the audio was corrupted with a ``hearing loss" audiogram \cite{nhanes}; in the second, a ``random'' corruption filter was used (more details in Technical Appendix).
By querying the user $Q$ times, each time with a new filter $h'$, {\name} expects to learn $K^*$, and in turn, maximize the user's satisfaction.
We report \texttt{Regret} against increasing $Q$, and compare to conventional GPR optimizers with simple base kernels $\{\text{SE},\text{PER}\}$.
The audio demos at various stages of the optimization is made public: \url{https://keroptbo.github.io/}.

\begin{table}
\centering
\begin{adjustbox}{width=1\columnwidth}
\begin{tabular}{|c|c|c|c|c|c|c|c|c|c|c|c|c|c|c|c|}
\hline
\multicolumn{10}{|c|}{Hearing Loss} \\
\hline
\multirow{2}{*}{Q} & \multicolumn{3}{c|}{U1} & \multicolumn{3}{c|}{U2} & \multicolumn{3}{c|}{U3} \\ 
\cline{2-10} 
& \multicolumn{1}{c|}{SE} & \multicolumn{1}{c|}{\name} & 
\multicolumn{1}{c|}{PER} & 
\multicolumn{1}{c|}{SE} & \multicolumn{1}{c|}{\name} & 
\multicolumn{1}{c|}{PER} & \multicolumn{1}{c|}{SE} & \multicolumn{1}{c|}{\name} & 
\multicolumn{1}{c|}{PER} \\ \hline 
$5$ &
$6$ &  \textbf{6} & 6 &
$8$ &  \textbf{8} & 8 &
$6$ &  \textbf{6} & 6 \\\hline 
 
$10$ &
$6$ & \textbf{8} & 6 &
$8$ & \textbf{8} & 8 &
$6$ & \textbf{7} & 7 \\\hline
 
$15$ &
$6$ & \textbf{10} &7 &
$8$ & \textbf{10} & 8 &
$7$ & \textbf{9} & 7 \\ \hline

$20$ &
$10$ & \textbf{10} & $10$ &
$9$ & \textbf{10} & 9 &
$7$ & \textbf{10} & 10 \\ \hline
 
 $25$ & 
 $10$ & \textbf{10} & 10 &
 $10$ & \textbf{10} & 10&
 $9$ & \textbf{10} & 10 \\ \hline \hline

\multicolumn{10}{|c|}{Random Audio Corruption} \\
\hline
\multirow{2}{*}{Q} & \multicolumn{3}{c|}{U1} & \multicolumn{3}{c|}{U2} & \multicolumn{3}{c|}{U3}\\ 
\cline{2-10} 
& \multicolumn{1}{c|}{SE} & \multicolumn{1}{c|}{\name} & 
\multicolumn{1}{c|}{PER} & 
\multicolumn{1}{c|}{SE} & \multicolumn{1}{c|}{\name} & 
\multicolumn{1}{c|}{PER} & \multicolumn{1}{c|}{SE} & \multicolumn{1}{c|}{\name} & 
\multicolumn{1}{c|}{PER} \\ \hline

$5$ &
$1$ & \textbf{1} & 1 &
$3$ & \textbf{3} & 3 &
$1$ & \textbf{1} & 1 \\\hline 
 
$10$ &
$2$ & \textbf{3} & 1 &
$4$ & \textbf{4} & 4 &
$2$ & \textbf{2} & 3 \\\hline
 
$15$ &
$2$ & 4 & 5 &
$4$ & \textbf{10} & 4 &
$2$ & \textbf{2} & 3 \\ \hline

$20$ &
$4$ & \textbf{10} & 5 &
$4$ & \textbf{10} & 9 &
$5$ & \textbf{8} & 4 \\ \hline
 
 $25$ & 
 $8$ & \textbf{10} & 9 &
 $8$ & \textbf{10} & 9 &
 $10$ & \textbf{8} & 7 \\ \hline

\end{tabular}
\end{adjustbox}
\caption{Audio personalization results (the $U$= $3$ volunteers (rest in Appendix) did not know which kernel was in use).}
\label{tab:audio}
\end{table}

\begin{table}
\centering
\begin{tabular}{|c|c|c|c|}
\hline
$x_{start}$ & $x_{Q=5}$ & $x_{Q=15}$ & $x_{Q=25}$\\\hline

\multicolumn{4}{|c|}{Prompt: Office room with a desk, a blue chair, a lamp } \\
\multicolumn{4}{|c|}{on the desk, and a green couch on the side} \\\hline

\includegraphics[width=0.2\columnwidth]{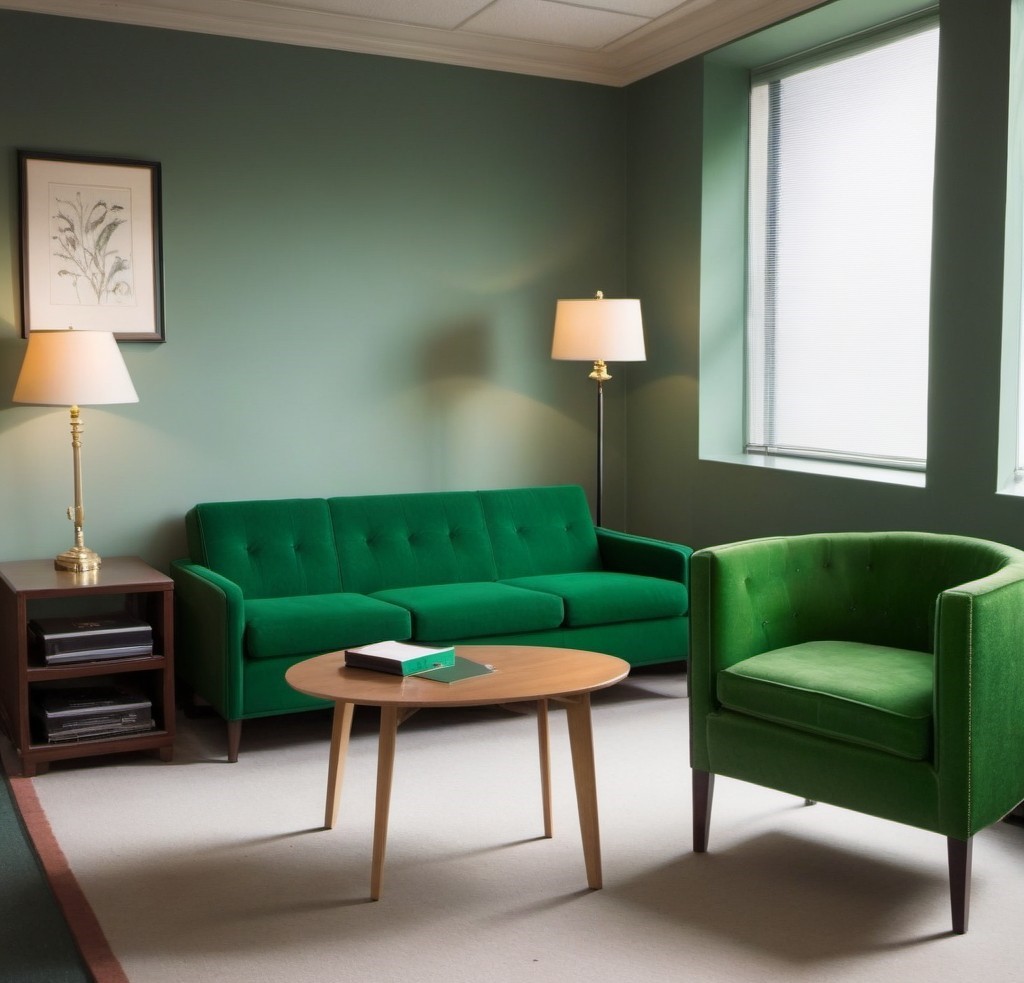} & \includegraphics[width=0.2\columnwidth]{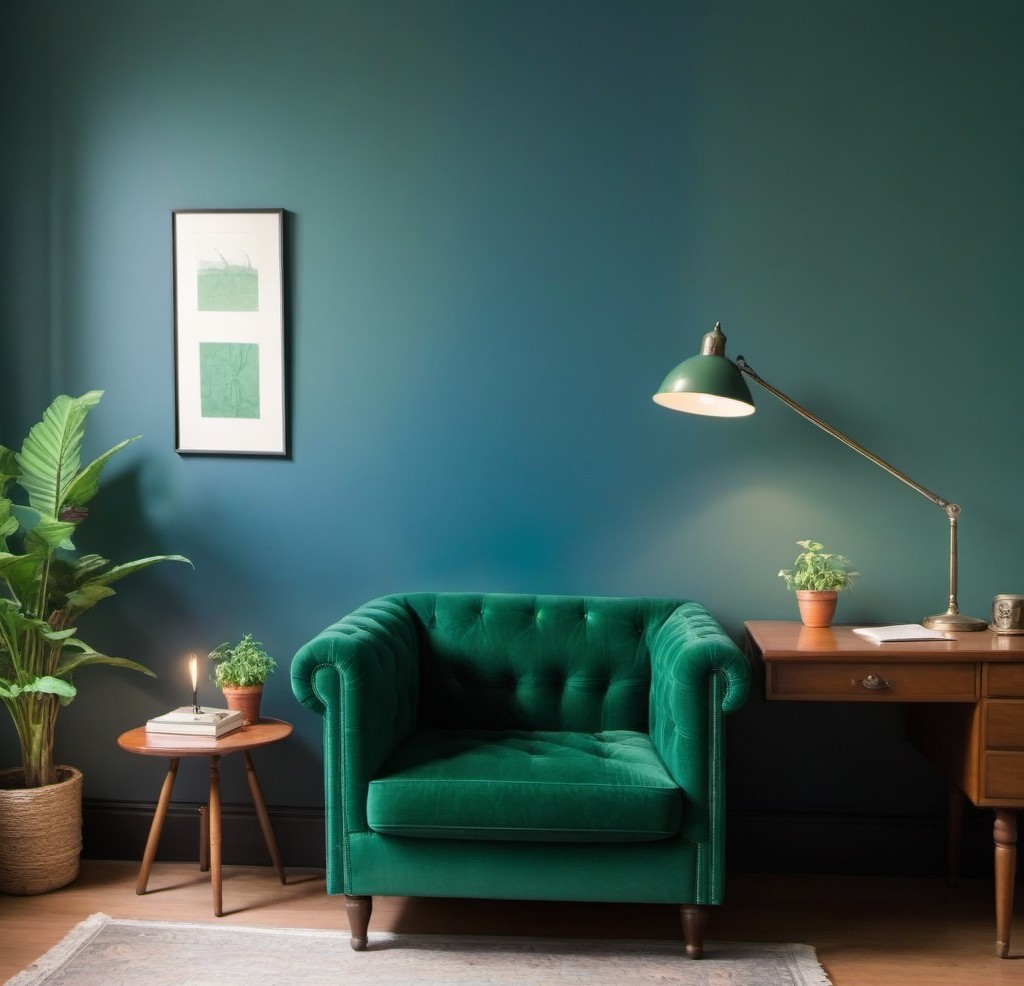}& \includegraphics[width=0.2\columnwidth]{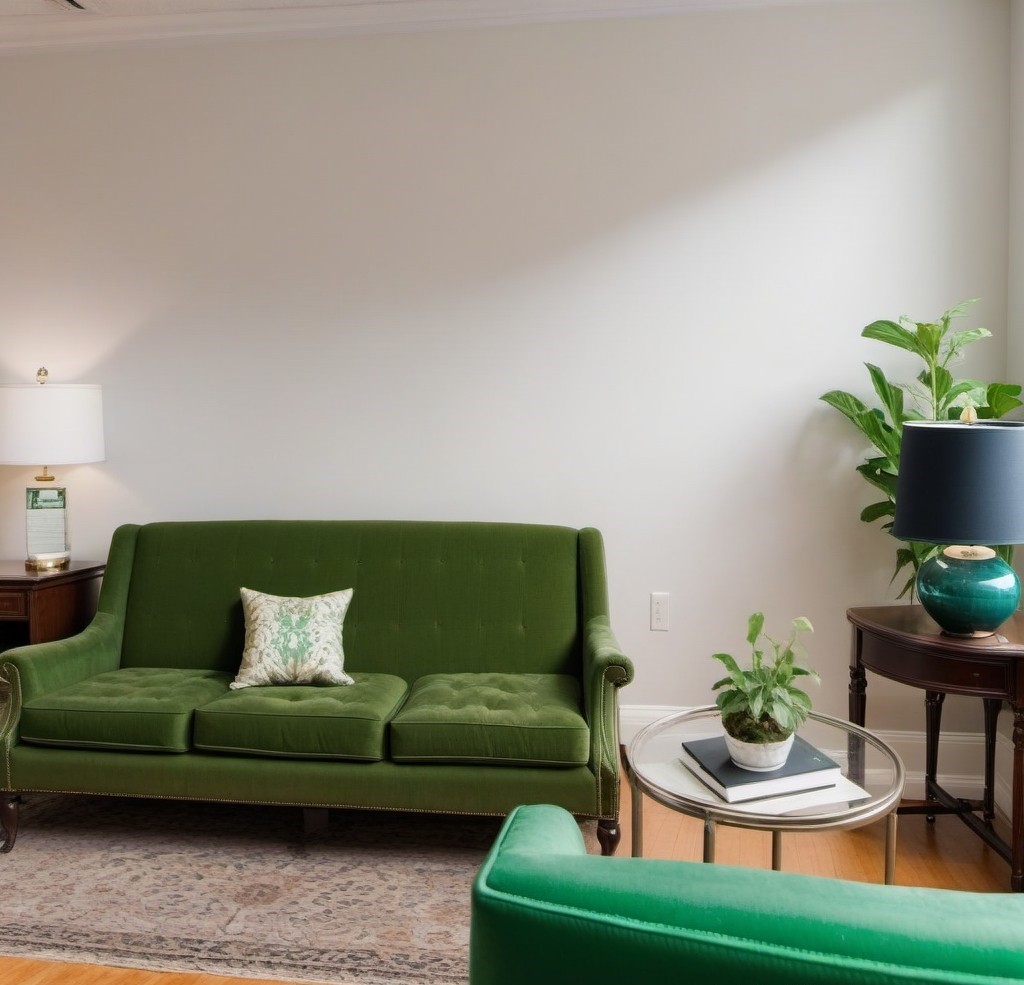}&  \includegraphics[width=0.2\columnwidth]{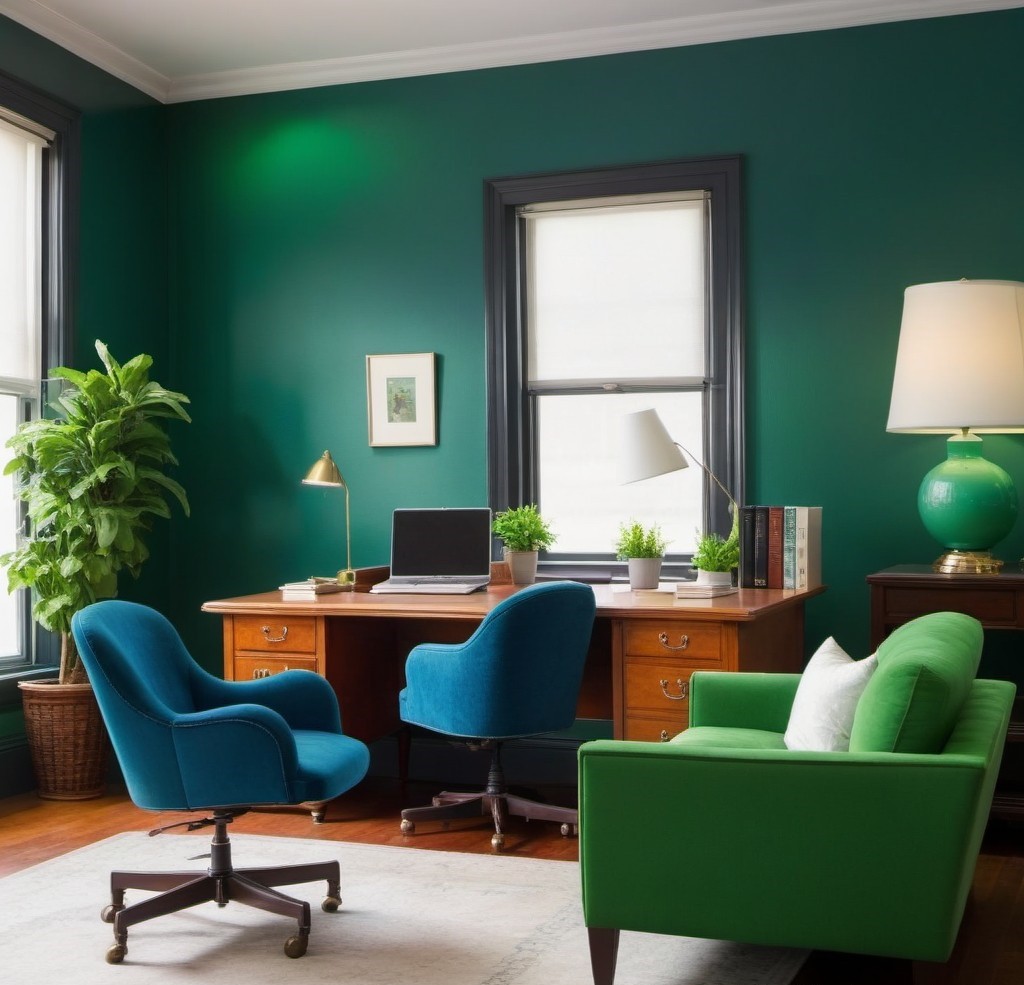}\\

\hline

\multicolumn{4}{|c|}{Prompt: Garden with a fountain in the center and} \\
\multicolumn{4}{|c|}{surrounded by trees and flowers of different colors} \\\hline
\includegraphics[width=0.2\columnwidth]{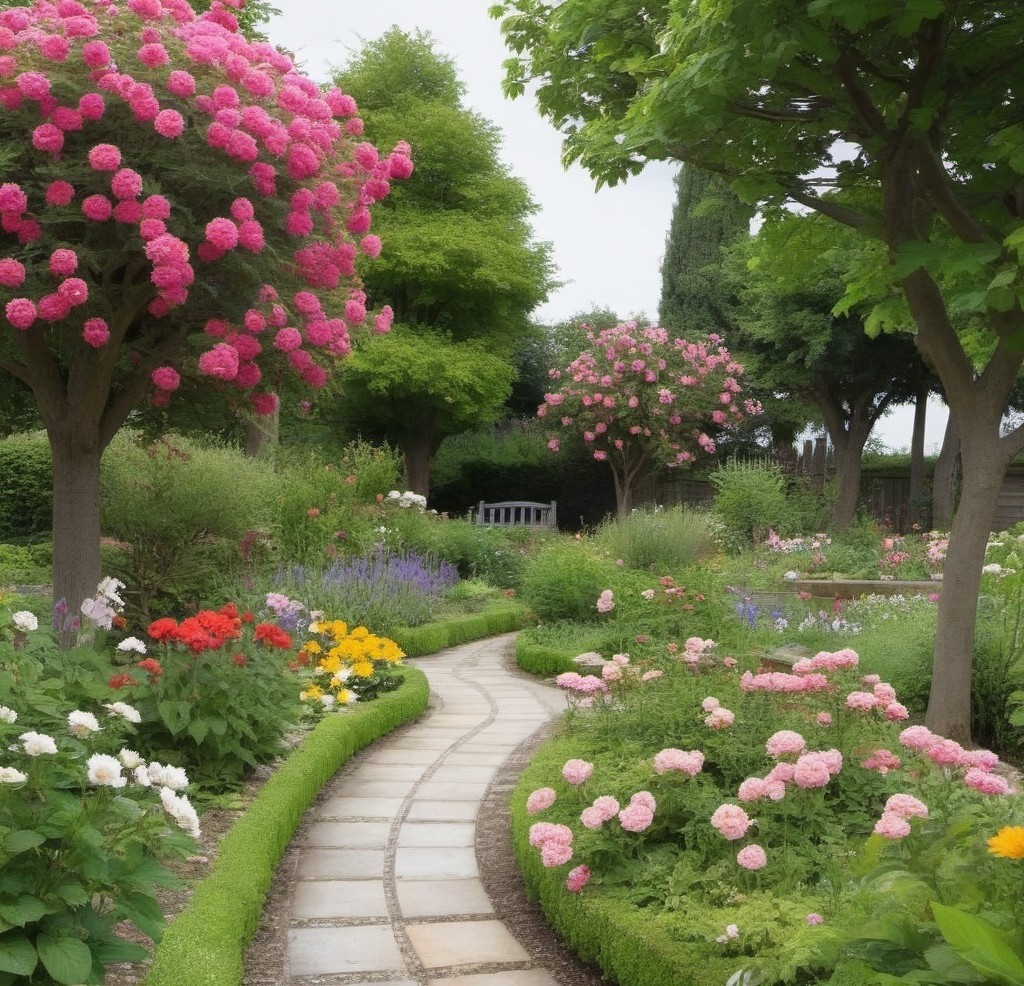} & \includegraphics[width=0.2\columnwidth]{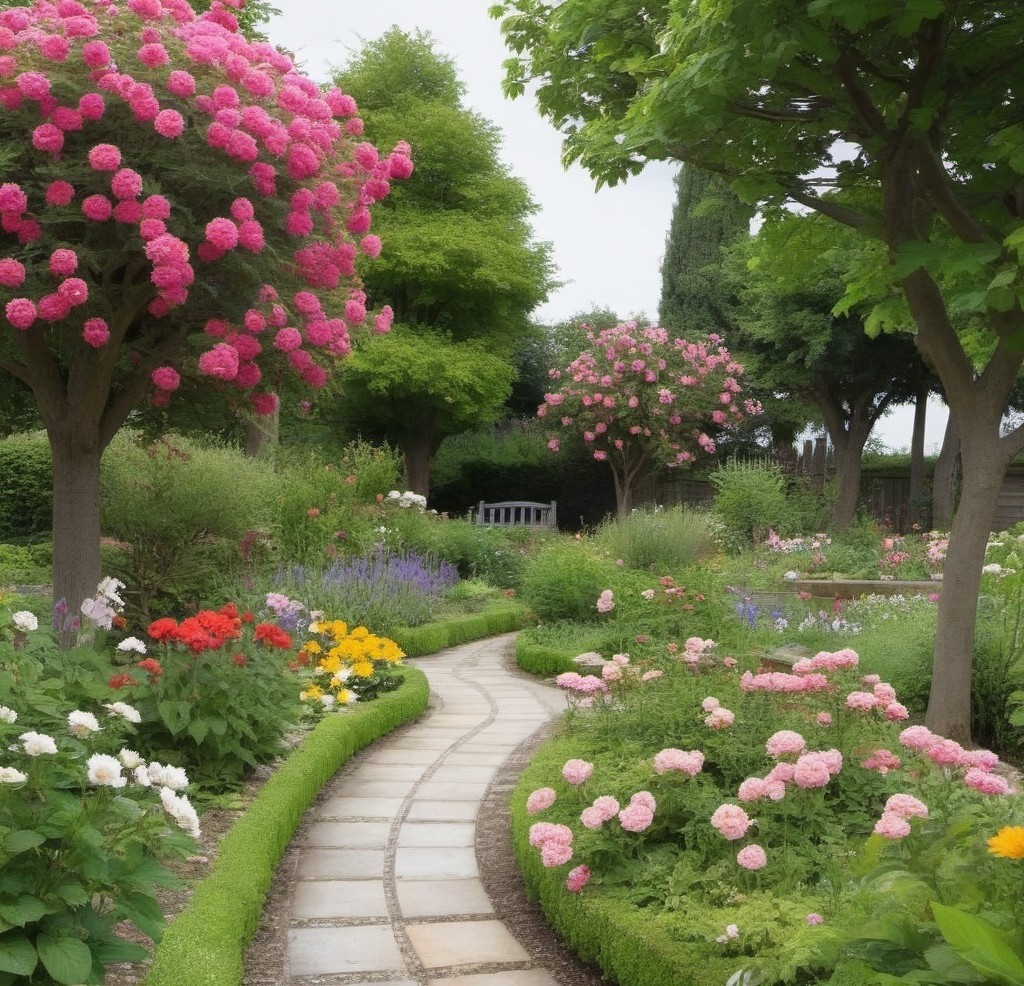} & \includegraphics[width=0.2\columnwidth]{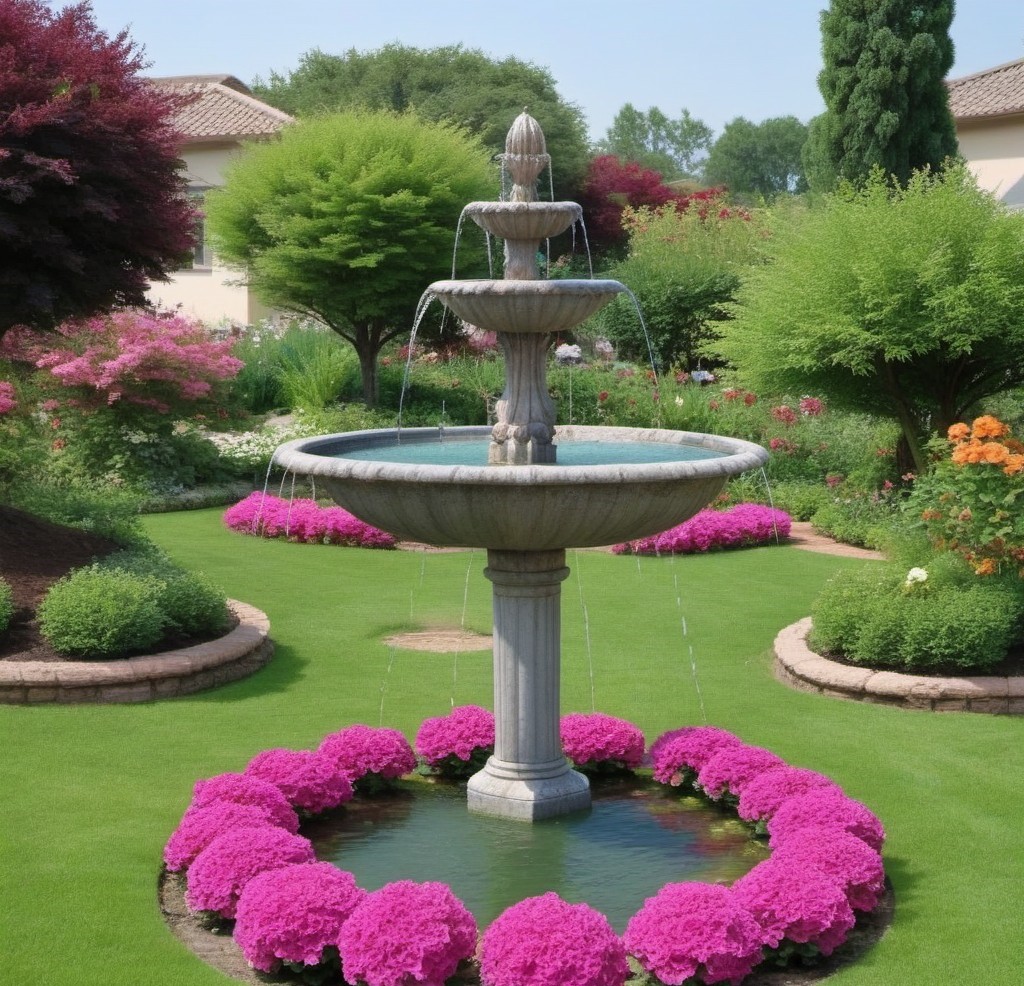}& \includegraphics[width=0.2\columnwidth]{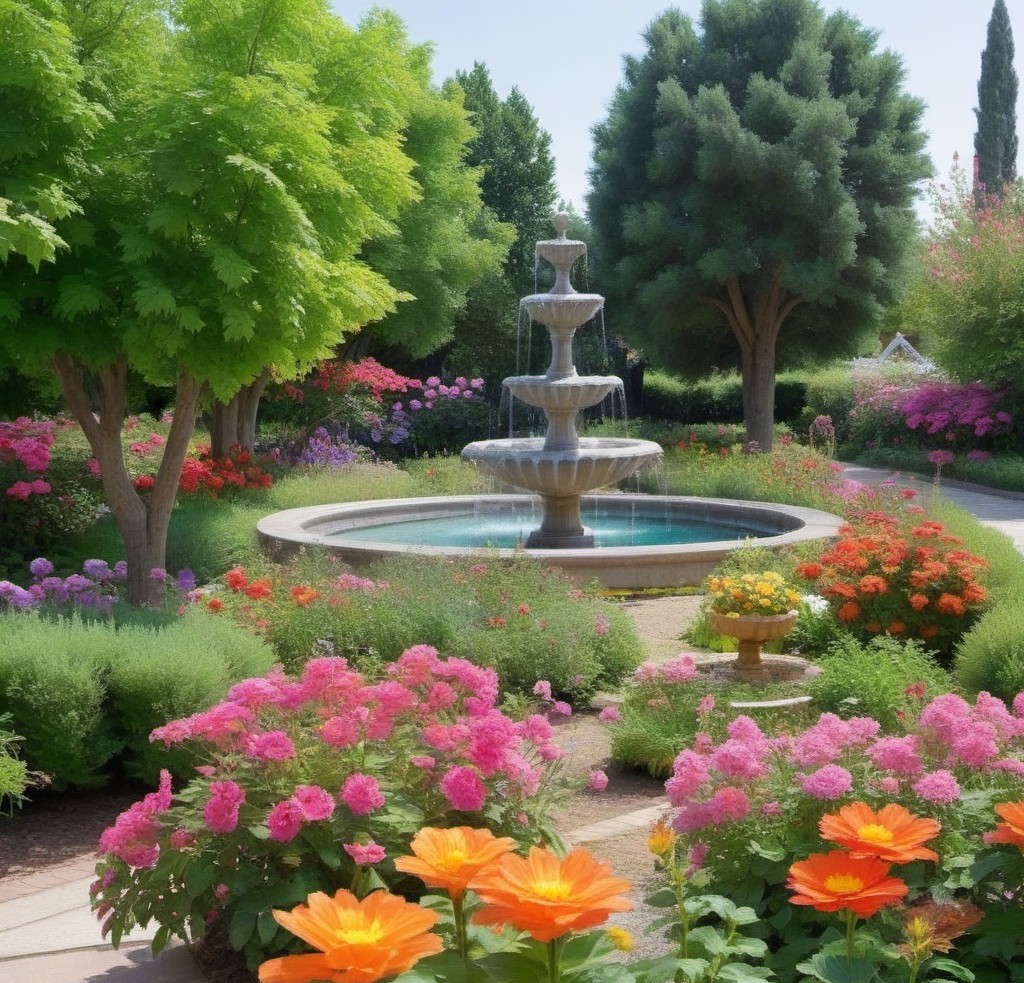}\\

\hline

\end{tabular}
\caption{Prompt-based image generation user results}
\label{tab:imageVAE}

\end{table}

\textbf{(2) Images:} We also apply {\name} to image recommendation.
The motivation is that users are often unable to articulate precisely what images they want, however, given an image they can express their satisfaction. 
We formulate this as a black-box optimization problem in the image space by posing the question: \textit{if a user rates few pictures generated by an AI model, can the model find the ``best" picture to maximize user satisfaction?}
To realize this, we use a pre-trained VAE-based image generator \cite{esser2020taming} (ImGen).
The user issues a crude prompt of what they want, allowing ImGen to display an initial image, $x_{start}$.
The user scores this image as $f(x_{start})$ and {\name} is triggered.
Table \ref{tab:imageVAE} displays $x_{start}$ and the images recommended after $Q=5,15,25$ queries. 
For a subjective measure of optimality, we asked users to describe their ideal image upfront.
{\name}'s recommendation at $Q \geq 15$ seems to match the prompts quite well (more results in Appendix).

\section{Related Work}
\label{sec:related-work}
A body of works in BO has explored kernel learning. 
Closest to {\name} are BOMS \cite{malkomes2016bayesian}, Automatic Statistician (AS) and their variants  \cite{duvenaud2013structure, grosse2012exploiting, kim2016scalable}, and MCMC \cite{gardner2017discovering, abdessalem2017automatic} discussed (and used as baselines) earlier. 
In other work \cite{teng2020scalable}, authors treat the kernel as a random variable and learn its belief from the data. 
\cite{zhen2020learning} introduces kernels with random Fourier features for meta-learning tasks. The kernel features are learned as latent variables of a model to generate adaptive kernels. 
In contrast, {\name} uses variational inference as an auxiliary module to only learn a {\em continuous} latent kernel space; the KerGPR optimization primarily drives the kernel learning. \\
Authors in \cite{kandasamy2015high,gardner2017discovering,mutny2018efficient,wang2018batched} have improved BO performance in high-dimensional spaces by modeling the function via additive kernels. The objective is decomposed into a sum of functions in low-dimensional space. 
{\name}'s comprehensive space of kernels from additive and multiplicative compositions is capable of modeling more complex function structures. 
Finally, \cite{kusner2017grammar}, \cite{gomez2018automatic}, and \cite{gonzalez2015bayesian} perform optimization in a continuous latent space learned by VAEs to circumvent categorical data. Authors of \cite{garrido2020dealing} use one-hot encoding approximations for BO of categorical variables. {\name} borrows from these ideas but applies them to  kernel learning.

\section{Limitations and Conclusion}
\label{sec:limit}

$\blacksquare$ \textbf{Trading Computation for Sample Efficiency:} 
{\name} incurs rounds of computation in estimating the model evidence $\mathcal{L}$. 
However, this does not affect sample efficiency, since KerVAE training is sample-independent.
Thus, {\name}'s advantage is in reducing the sample evaluations of $f(x)$ (e.g., user burden) and not in total CPU cycles. \\
$\blacksquare$ \textbf{Overfitting to Simple Functions:}
As iterations progress, KerGPR might learn a kernel more complex than the actual target function $f$ (see $f_1(x)$ in Table~\ref{tab:synthetic}). 
Choosing a complex kernel expands the surrogate function space, and may need more samples to converge. 
To avoid kernel overfitting, we can regularize the kernel complexity, i.e., the length and norm of the kernel grammar codes. \\
$\blacksquare$ \textbf{Latent Space Interpretability:}
Current latent space learned by KerVAE is abstract and lacks interpretability. An interpretable latent space should offer improvements to KerGPR, facilitating the use of simpler optimizers compared to the expensive Bayesian Optimization in the latent space.

To conclude, we propose {\name}, a kernel learning method for GPR.
We design a continuous latent space of kernels (using a VAE), and optimize the space via an auxiliary GPR to output an optimal kernel $\mathbf{K}^*$.
This optimal kernel better models the structure of the objective function, which ensures sample efficiency.
We show sample applications and believe the ideas could be extended to others.

\section*{Acknowledgements}
We thank Foxconn and NSF (grant 2008338, 1909568, 2148583, and MRI-2018966) for funding this research. We are also grateful to the reviewers for their insightful feedback.
\bibliography{aaai25}
\newpage
\newpage
\newpage
\appendix
\section{Technical Appendix}
\label{sec:append}
\subsection{Bayesian Optimization Background}
Bayesian optimization \cite{frazier2018tutorial} broadly consists of the following two modules:

\begin{itemize}
\item[(1)] \textbf{Surrogate model}: 
A family of functions that serve as candidates for the unknown objective function. 
The functions are commonly drawn from a Gaussian process generated by \textbf{Gaussian Process Regression (GPR)}. 
Given any point of interest $x$, GPR generates a Gaussian posterior distribution for the function values $f(x)$

\item[(2)] \textbf{Acquisition function}: A sampling strategy that prescribes the point at which $f$ should be observed next. 
The GPR posterior model is used to evaluate the function at new points $x'$, and one is picked that maximizes a desired metric. 
This new point $x'$ when observed will maximally improve the GPR posterior.
\end{itemize}

\textbf{Non-parametric model}: 
Gaussian processes \cite{wang2020intuitive} are employed for black-box optimization because they provide a non-parametric mechanism to generate a  probabilistic surrogate for the unknown function $f$. 
Given a set of samples $\mathcal{X} = \{x_1,x_2,\dots,x_K\}$ at which the function $f$ has been observed, i.e., we know $\mathcal{F} = \{f(x_1),f(x_2),\dots,f(x_K)\}$, we can identify an infinite number of candidate functions that match the observed function values.

\textbf{Curse of Dimensionality}: 
The objective function in Eqn. 1 in the main paper, typically lies in a high dimensional space (i.e., $h \in \mathcal{H} \subseteq \mathbf{R}^N, N \ge 500$). 
Bayesian optimization works well for functions of $N<20$ dimensions \cite{frazier2018tutorial}; with more dimensions, the search space $\mathcal{H}$ increases exponentially, and finding the minimum with {\em few} evaluations becomes untenable. 
One approach to reducing the number of queries is to exploit the sparsity inherent in most real-world functions. 

\subsubsection{High dimensional Bayesian Optimization}
We assume our function in Eqn.1 in the main paper is sparse, i.e., there is a low-dimensional space that compactly describes $f$, so $f$ has ``low effective dimensions". 
We consider ALEBO \cite{letham2020re}, a BO method that exploits sparsity to create a low-dimensional embedding space using random projections.

\textbf{ALEBO}\cite{letham2020re}: 
Given a function $f: \mathbf{R}^N \to \mathbf{R}$ with effective dimension $d_f$, ALEBO's linear embedding algorithm uses random projections to transform $f$ to a lower dimensional embedding space.
This transformation must guarantee that the minimum $h^*$ from high dimensional space $\mathcal{H}$ gets transformed to its corresponding minimum $y^*$ in low dimensional embedding space. 

The random embedding is defined by an embedding matrix $\mathbf{B} \in \mathbf{R}^{d \times N}$ that transforms $f$ into its lower dimensional equivalent $f_B(y) = f(h) = f(\mathbf{B}^\dagger y)$, where $\mathbf{B}^\dagger$ is the pseudo-inverse of $\mathbf{B}$. 
Bayesian optimization of $f_B(y)$ is performed in the lower dimensional space $\mathbf{R}^d$.

Our proposed idea \name\ builds on top of ALEBO, but we are agnostic of any specific sparsity method.

\subsection{More details on Evaluation and Results}
\label{sec:eval}

\subsubsection{Synthetic Functions}
\label{sec:app_func}
The synthetic functions employed in {\name}'s performance evaluation are defined as follows,

\begin{itemize}
    \item[(1)]  \textbf{Staircase Functions}: We generated functions that have a discontinuous staircase structure to mimic the user satisfaction scoring function in audio/ visual perception. A user's perception might not change for a range of filters so the score remains the same and might change with sudden jumps for some filter choices, thus leading to a flat shape in some regions and a steep curve in other regions. The staircase structure results in the function having infinite local minima, infinite global minima, and zero gradient regions. These functions are naturally not suited for gradient-based optimization techniques. In this work, we use the function defined in the Eqn below,
\begin{equation}
\begin{aligned}
f_{P1}(\mathbf{x}) = \sum_i^N (\lfloor |x_i + 0.5| \rfloor)^2
\end{aligned}
\label{eqn:P1_stair}
\end{equation}
where, $-100\le x_i \le 100, i=1,2,\dots,N$, $x_i$ is filter $x$ along dimension $i$. Infinite global minima at $f_{min}(\mathbf{x}^*) = 0$, and the minimizers are $-0.5 \le x_i^* < 0.5$ (i.e.,) $x_i^* \in [-0.5,0.5), i=1,2,\dots,N$
\item[(2)] \textbf{Branin Functions}: A commonly used smooth benchmark function in Bayesian optimization research \cite{kim2020benchmark}. \texttt{BRANIN} defined as

    \begin{equation}
\begin{aligned}
f_{B}(\mathbf{x}) = a(x_2 &- bx_1^2 +cx_1 -r)^2\\ &+ s(1-t)cos(x_1) + s
\end{aligned}
\label{eqn:B}
\end{equation}
where, $x_i\in[-5,10], x_2\in[0,15], a=1, b=5.1⁄(4\pi^2), c=5⁄\pi, r=6, s=10, t=1⁄(8\pi)$.

It has three global minima at $f_{min}(\mathbf{x}^*) = 0.397887$, and the minimizers are $\mathbf{x}^*=(-\pi,12.275), (\pi,2.275), (9.42478,2.475)$

\item[(3)] \textbf{Periodic Functions}: We generate functions that exhibit periodicity in its structure: \texttt{MICHALEWICZ} defined in Eqn \ref{eqn:M} to evaluate \name.

\begin{equation}
\begin{aligned}
f_{M}(\mathbf{x}) = -\sum_{i=1}^{d} sin(x_i)sin^{2m}\left(\frac{ix_i^2}{\pi}\right)
\end{aligned}
\label{eqn:M}
\end{equation}
where, $x_i\in[0,\pi], m = 10, i=1,2,\dots.d$. 

The global minimum is at $f_{min}(\mathbf{x}^*) = -9.66015$, and the minimizer is $\mathbf{x}^*=(2.20,1.57)$

\end{itemize}

\textbf{Synthetic Function Evaluation Parameters:}
\label{sec:app_para}
In our synthetic \name\ experiments, we optimize the functions with the following parameters and/or configurations:
\begin{itemize}
    \item[-] f\_evals = 100 function evaluations
    \item[-] r\_init = 5 initial random samples after which the acquisition sampling begins.
    \item[-] $N = 2000, \mathcal{H} \subseteq \mathbf{R}^N$ is the high-dimensional space.
    \item[-] $\mathbf{R}^d, d = 20$ is the low-dimensional embedding space after sparse transformation.
    \item[-] The kernel learning module of \name\ is employed to learn the kernel after every 5 iterations (function observations $\mathcal{D}$) of Function space GPR ($f$GPR). i.e., we begin by using an SE kernel for 5 initial iterations of $f$GPR and is replaced by a newly learned kernel $\mathbf{K}^*$ for the next five iterations and so on.
    \item[-] We restrict the maximum power of the fractional code to 3, i.e., in Eqn 5 in the main paper, 
    $0 \leq a_i+b_i+c_i+d_i+e_i \leq 3$ for $i=1,2,3,\dots$. 
    \item[-] The KerVAE encoder and decoder blocks are identical with three fully connected hidden layers each and ReLU activations. The KerVAE latent space dimensions in 2.     
    \item[-] An SE kernel is used in KerGPR, which runs in the kernel latent space $\mathcal{Z}$ of KerVAE.\\
    
    \textbf{A note on KerGPR's kernel:} We do not learn KerGPR's kernel -- its objective, Model Evidence (Eqn. 4 in main paper) is sufficiently smooth, unlike the objective $f(x)$ of $f$GPR whose discontinuous structure benefits greatly from kernel learning. Kernel learning is useful when the samples are limited (e.g. human users have a limited query budget) whereas KerGPR can sample $\mathcal{Z}$ unrestricted as many times as needed because the human is not in the loop. Hence, KerGPR can model the objective well even with a general SE kernel. Of course, computation increases but KOBO’s main goal is sample efficiency of $f$GPR (i.e., lower user burden).

    \item[-] KerGPR runs for 20 iterations in the kernel latent space $\mathcal{Z}$ to find the current best kernel $\mathbf{K}^*$.
    \item[-] We run 10 random runs of each experiment (each synthetic function: Staircase, Branin, Michalewicz and each algorithm: {\name}, CKS, BOMS, MCMC).
        
\end{itemize}  

\subsubsection{Computation Complexity}
\label{sec:comp}
\begin{itemize}
    \item We use two NVIDIA RTX 3090 cards with 24 GB memory to run all experiments.
    \item Each run of an experiment as outlined in Section \ref{sec:app_para} consumes 65\% of the memory and takes about 2.5 hours to complete.
\end{itemize}

\subsubsection{Baseline details}
\label{sec:app_base}
\textbf{MCMC:}
The MCMC algorithm \cite{gardner2017discovering}\cite{abdessalem2017automatic} is used as a comparison baseline.
The number of possible composite kernel models $\mathbf{k}_{\mathcal{C}}$ is very large. 
This makes computing the model evidence for each kernel prohibitively expensive. 
However, computing a small (i.e., not super-exponential) number of model evidence is tractable. 
Thus, the model evidence is sampled using any MCMC method like Reverse Jump MCMC (RJMCMC) \cite{hastie2012model} or  Metropolis-Hastings \cite{gardner2017discovering}. 

To apply Metropolis-Hastings, the proposal distribution $g(\mathbf{k}'| \mathbf{k})$ is defined as follows; 
given a current kernel model $\mathbf{k}$, we can either add or multiply a chosen base kernel from $\mathcal{B}$. 
We construct the proposal distribution by first choosing whether to add or multiply, each with 50\% probability, and next, picking a base kernel from $\mathcal{B}$ uniformly at random.

Given the current kernel $\mathbf{k}_j$ (initializing $\mathbf{k}_0$ to the final kernel model found in the previous iteration), we sample a proposed model $\mathbf{k}'$ from the proposal distribution $g(\mathbf{k}'| \mathbf{k}_j)$. Next, we compute the model evidence for $\mathbf{k}'$ and use this to compute the Metropolis-Hastings acceptance probability:
\begin{equation}
\begin{aligned}
A(\mathbf{k}'|\mathbf{k}_j) = \text{min}\left(1, \frac{P(\mathcal{F}|\mathcal{D},\mathbf{K}')g(\mathbf{k}_j| \mathbf{k}')}{P(\mathcal{F}|\mathcal{D},\mathbf{K}_j)g(\mathbf{k}'| \mathbf{k}_j)} \right)
\end{aligned}
\label{eqn:mcmc}
\end{equation}
Finally, we update the current state of MCMC to $\mathbf{k}'$ with probability $A(\mathbf{k}'|\mathbf{k}_j)$.

\textbf{CKS:}
The automatic Statistician/ Compositional Kernel Search (CKS) \cite{duvenaud2013structure} method takes advantage of the fact that complex kernels are generated as context-free grammar compositions of positive semi-definite matrices (closed under addition and multiplication) and then the kernel selection is a tree search guided by model evidence. 

The search process begins with the base kernels in $\mathcal{B}$. At each search iteration, the current kernel $\mathbf{K}$ is modified as follows: \vspace{-0.05in}
\begin{itemize}
    \item[(i)] $\mathbf{K}$ can be replaced with $\mathbf{K} + \mathcal{B}$. $\mathcal{B}$ is any base kernel.
    \item[(ii)] $\mathbf{K}$ can be replaced with $\mathbf{K} \times \mathcal{B}$. $\mathcal{B}$ is any base kernel.
    \item[(iii)] If $\mathbf{K}$ is a base kernel, it can be replaced with $\mathcal{B}'$. $\mathcal{B}'$ is another base kernel.
\end{itemize}\vspace{-0.05in}
CKS thus searches over the discrete kernel space $\mathcal{K}$ using a greedy strategy that, at each iteration, chooses the kernel with the highest model evidence. This kernel is then expanded by applying all possible operations listed above. The search process repeats on this expanded list.   

\textbf{BOMS:} The Bayesian Optimization for Model Search (BOMS) \cite{malkomes2016bayesian} method also generates the discrete kernel space $\mathcal{K}$ through kernel compositions. Unlike CKS' greedy strategy, BOMS' search is a meta-learning technique, which, conditioned on observations $\mathcal{D}$ available, establishes similarities among the kernel choices in $\mathcal{K}$ in terms of how they offer explanations for the objective function's structure. In other words, BOMS constructs a kernel between the kernels (a ``kernel kernel”). 

Like {\name}, BOMS treats the model evidence as an expensive black-box function to be optimized as shown in Eqn. 4 in main paper. BOMS then performs BO in $\mathcal{K}$ by defining a Gaussian distribution: $P(g) = \mathcal{N}(g;\mu_g,\mathbf{K}_g)$, where $g$ is the model evidence, $\mu_g$ is the mean of the Gaussian distribution in kernel space, and $\mathbf{K}_g$ is the corresponding covariance. The covariance $\mathbf{K}_g$ is constructed by defining a heuristic similarity measure between two kernels: Hellinger distance. The Hellinger distance is a probability metric that measures the similarity between two distributions $P$ and $Q$:
\begin{equation}
\begin{aligned}
d_H^2(P,Q) &= 1- \frac{|\mathbf{K}_P|^{1/4}|\mathbf{K}_Q|^{1/4}}{\frac{|\mathbf{K}_P+\mathbf{K}_Q|}{2}^{1/2}} \\ &\exp\left\{ -\frac{1}{8}(\mu_P-\mu_Q)^T(\frac{\mathbf{K}_P+\mathbf{K}_Q}{2})^{-1}(\mu_P-\mu_Q)\right\}
\end{aligned}
\label{eqn:helli}
\end{equation}
where $P = \mathcal{N}(\mu_P,\mathbf{K}_P), Q= \mathcal{N}(\mu_Q,\mathbf{K}_Q)$. The Hellinger distance is small when $P$ and $Q$ are highly overlapping i.e., kernels $\mathbf{K}_P,\mathbf{K}_Q$ provide similar modeling of objective function structure. The covariance $\mathbf{K}_g$ for observations $\mathcal{D}$ is defined as,
\vspace{-0.15in}
\begin{equation}
\begin{aligned}
\mathbf{K}_g(\mathbf{K}_a,\mathbf{K}_b;\mathcal{D}) = \sigma^2\exp\left(-\frac{1}{2}\frac{d_H^2(\mathbf{K}_a,\mathbf{K}_b;\mathcal{D)}}{l^2}\right)
\end{aligned}
\label{eqn:kg}
\end{equation}
where, $\sigma,l$ are hyperparameters of the kernel-kernel $\mathbf{K}_g$.

Since the kernel space $\mathcal{K}$ is discrete, it is challenging to explore an infinite set of kernels to advance the BO kernel search. At each iteration, BOMS creates an active candidate set of kernels $\mathcal{C}$ and BO is applied by observing the model evidence of kernels sampled from this set. BOMS traverses the CKS kernel tree and populates the set $\mathcal{C}$ starting with the base kernels. The base kernel with the highest model evidence is retained. New kernels are then composed from the retained kernel (via addition or multiplication) and the "closest" neighbor kernels are added to the set $\mathcal{C}$. Neighbour kernels are those newly composed kernels within a certain Helliger distance from the retained kernel. 

\subsubsection{Additonal Results}
We evaluate some additional results about (1) investigating the impact of the length of Kernel Combiner encoding $r_c$, (2) some more visualizations of {\name}'s ability to learn function structure, (3) analyzing KerVAE's ability to learn a good latent space through the validity of its kernel reconstructions, (4) an ablation study elaborating on the advantages of the Kernel Combiner representations $r=[r_c,r_d]$ against established techniques like one-hot encoding schemes, and (5) results on two other synthetic functions: \texttt{Styblinski-Tang} in 2000 dimensions, and \texttt{Hartmann-6}. 

\begin{figure}[htbp]
\centering
{\includegraphics[width=0.49\columnwidth]{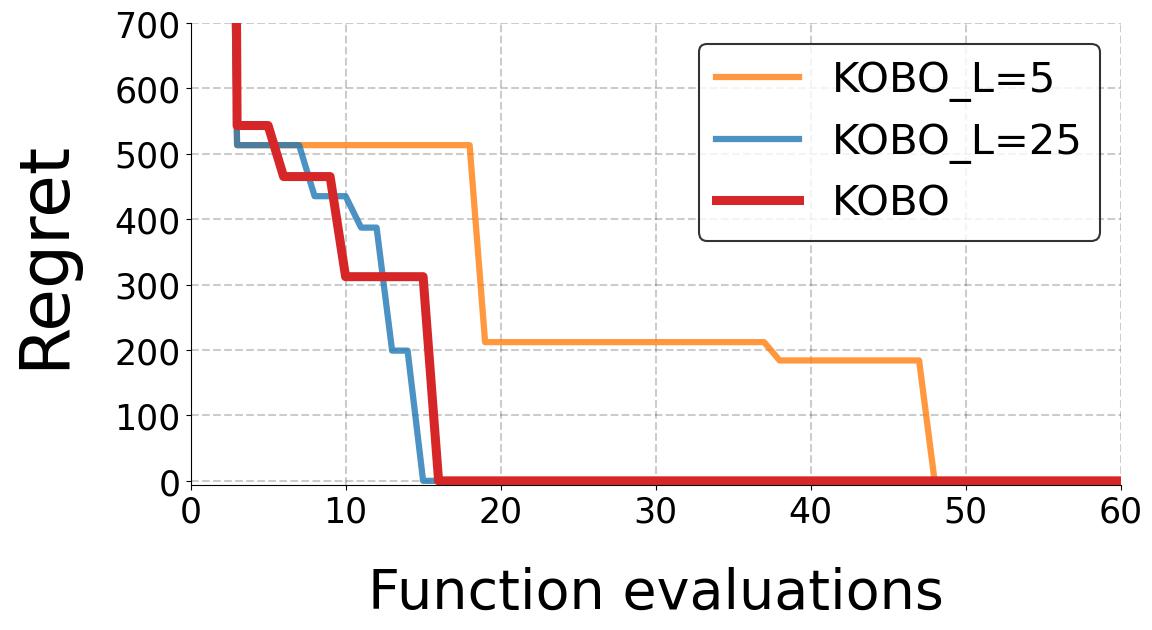}}
{\includegraphics[width=0.49\columnwidth]{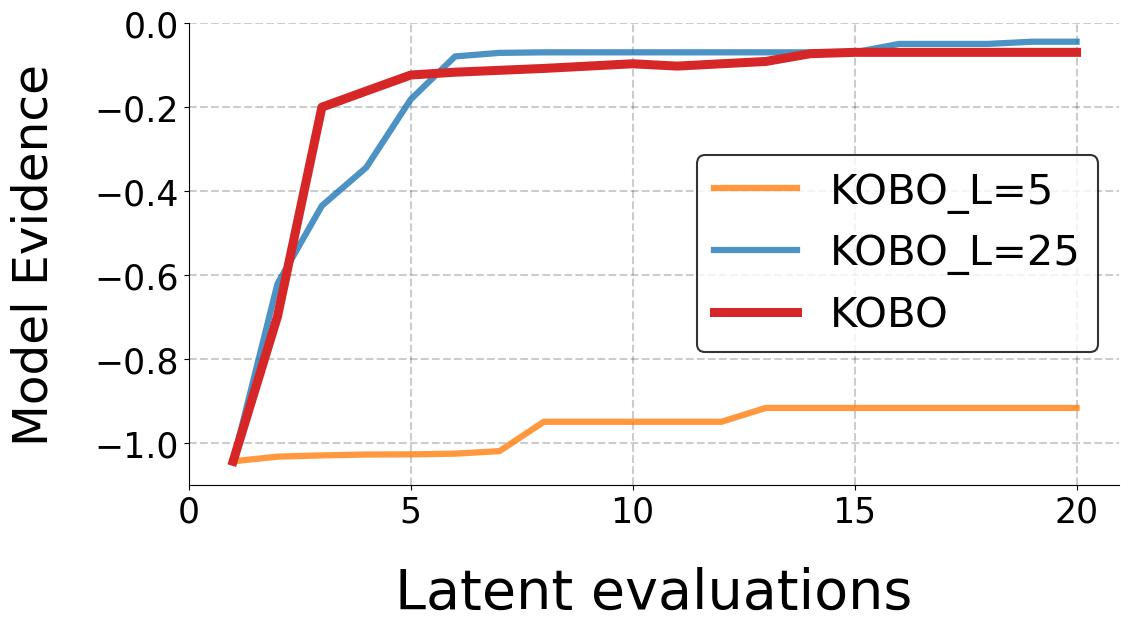}}
\vspace{-0.5em}
\caption{\textit{Comparison of {\name} with $r_c$ lengths $L=5,15,25$:} (a) \texttt{Regret} (b) \texttt{Model Evidence}.}
\label{fig:len}
\vspace{-0.1in}
\end{figure}

\textbf{Length $L$ of Grammar-based Encoding $r_c$:}
A fixed encoding length of $L=15$ (Eqn, 5 in main paper) was used in the kernel generation process of the Kernel Combiner. We arrived at this choice after empirical examination of two additional lengths $L=5$ and $L=25$. Figure \ref{fig:len} shows that increasing the encoding length arbitrarily has diminishing advantage. For the staircase objective function, we observe that a larger encoding length $L=25$ does not offer significant gains over the chosen length $L=15$ and length $L=5$ can not generate kernels with enough complexity to model the objective function structure. This implies that $L=15$ generates a kernel space $\mathcal{K}$ that contains kernels that adequately model the function considered.

\begin{figure*}[htbp]
\centering
{\includegraphics[width=0.65\columnwidth]{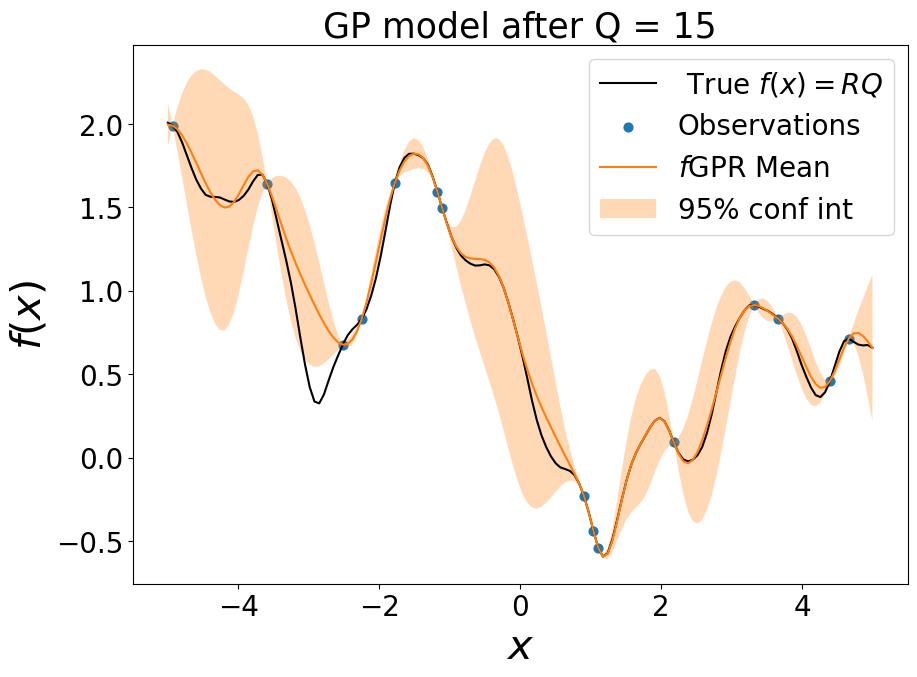}}
\hfill
{\includegraphics[width=0.65\columnwidth]{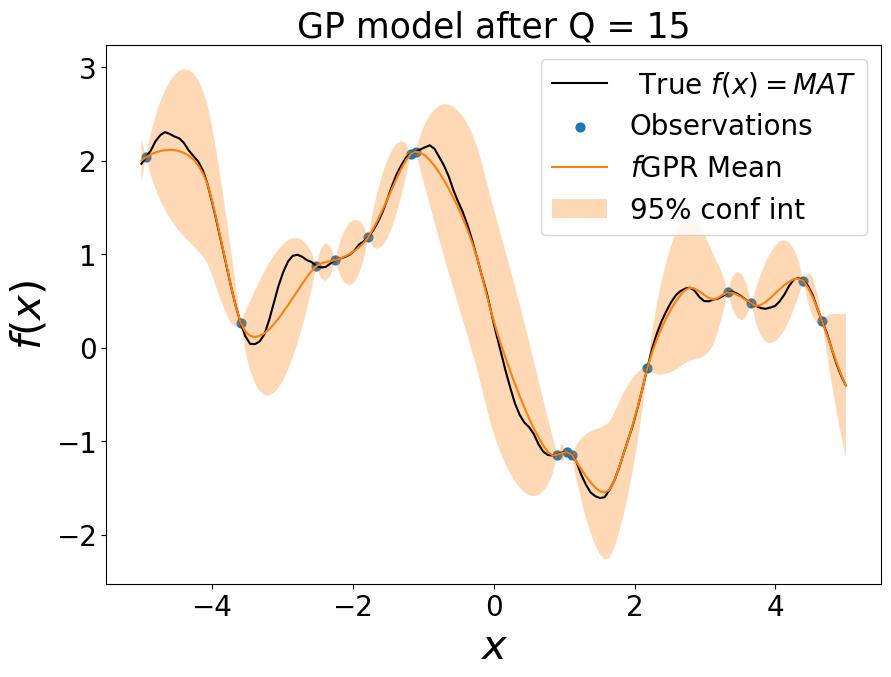}}
\hfill
{\includegraphics[width=0.65\columnwidth]{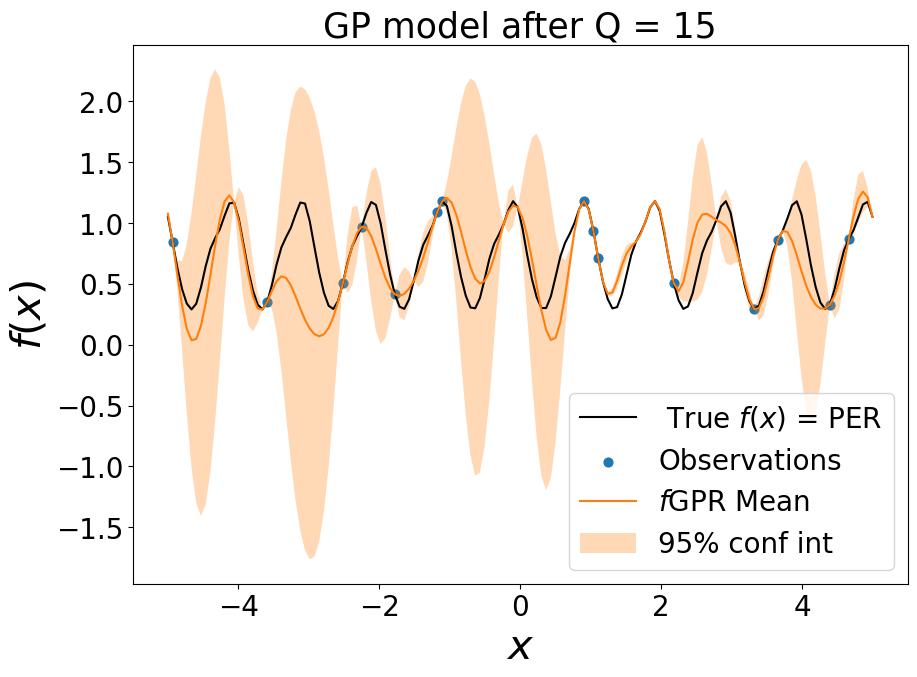}}
\vspace{-0.15in}
\caption{\textit{Function structure modeling with \name:} Objective function drawn from a GPR posterior using (a) Rational Quadratic (b) Mat\'{e}rn, and (c) Periodic kernels. The black line denotes the true function. The orange line denotes the $f$GPR posterior mean function model using $Q=15$ samples.}
\label{fig:R8}
\end{figure*}

\textbf{Visualizing with 1D synthetic functions:}
To visualize kernel learning (as previously done for $CO_2$ emissions data in Section 4 in main paper), we sample 1D functions from GPs employing different kernels $\{\text{PER},\text{RQ},\text{MAT}\}$.
Figure \ref{fig:R8} shows that in $15$ function evaluations, KerGPR learns the exact kernels for each, and consequently, the $f$GPR posterior models the 1D functions almost perfectly.

\begin{figure}[htp]
\centering
{\includegraphics[width=0.8\columnwidth]{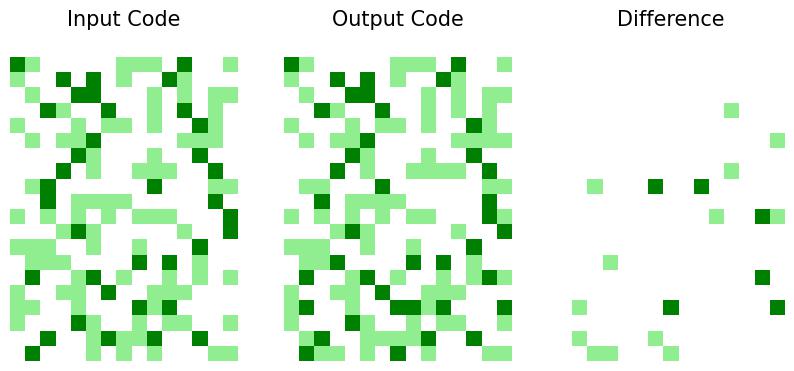}}
\vspace{-0.1in}
\caption{KerVAE reconstruction: (Block 1) Input, (Block 2) Reconstruction, (Block 3) Difference. White squares denote $0$ in code $r_c$, light green denotes $1$, and dark green denotes $2$.}
\vspace{-0.1in}
\label{fig:R4}
\end{figure}

\textbf{KerVAE Reconstruction:}
Figure ~\ref{fig:R4} shows the KerVAE reconstruction results. 
Figure ~\ref{fig:R4}(a) and (b) denote the input kernel code $r_c$ and its reconstruction respectively.
Each row in the input indicates the grammar-based representation $r_c$ of some kernel $\mathbf{k}_{\mathcal{C}}$; the corresponding row in the output shows the KerVAE reconstructed code. 
The $r_c$ codes are of length$=15$ (hence 15 columns in each matrix in Figure \ref{fig:R4}). 
KerVAE generates a near-perfect reconstruction as indicated by the very small difference in Figure \ref{fig:R4}(c). 
This offers confidence that KerVAE's latent space effectively learns the kernel space $\mathcal{K}$. 

\subsubsection{Ablation Studies}
We further elaborate on the advantages of the grammar and data-based representation of Kernel Combiner as illustrated in Figure 5 in main paper. 

\textbf{Proposed kernel encoding vs. One-hot representation:}
Past work has proposed one-hot encoding \cite{garrido2020dealing}\cite{lu2018structured} to encode discrete kernels, while we design the grammar-based encoding in Eqn. 5 in main paper.
Table \ref{tab:encoding} shows both representations for two semantically similar kernels.
The One-hot encoding assigns codes to each base kernel and each operator (e.g., $A=00001,~*=010$, etc.) and concatenates them; consequently, the two one-hot codes are quite different even for similar kernels.
{\name}'s grammar representation preserves the ``continuity'' in code space.
Figure 5(a)\&(b) in main paper visualize this notion of continuity in the code space, i.e., the latent space learned by KerVAE with $r_d=0$.

\begin{table}[!h]
\centering
\vspace{-0.1in}
\caption{Comparing grammar vs. one-hot encoding}
\vspace{-0.05in}
\resizebox{\columnwidth}{!}{%
\begin{tabular}{|c|c|}
\hline &     $\mathbf{k}_{\mathcal{C}_1} = \mathbf{A}^2 * \mathbf{B} + \mathbf{D}$       \\ \hline
Grammar-based  &   $[2, 1, 0, 0, 0, 0, 0, 0, 1, 0, 0, 0, 0, 0, 0]$         \\ \hline
One-hot   &  $[00001,010,00001,010,00010,001,01000,000,00000]$         \\ \hline \hline
    &  $\mathbf{k}_{\mathcal{C}_2} = \mathbf{A}^2 * \mathbf{B} + \mathbf{C} * \mathbf{D}$       \\ \hline
Grammar-based     &   $[2, 1, 0, 0, 0, 0, 0, \mathbf{1}, 1, 0, 0, 0, 0, 0, 0]$         \\ \hline
One-hot     &      
     $[00001,010,00001,010,00010,001,0\mathbf{01}00, 0\mathbf{1}0,0\mathbf{1}000]$\\ \hline
\end{tabular}%
\label{tab:encoding}
}
\vspace{-0.1in}
\end{table}
\begin{figure}[!h]
\centering
{\includegraphics[width=0.49\columnwidth]{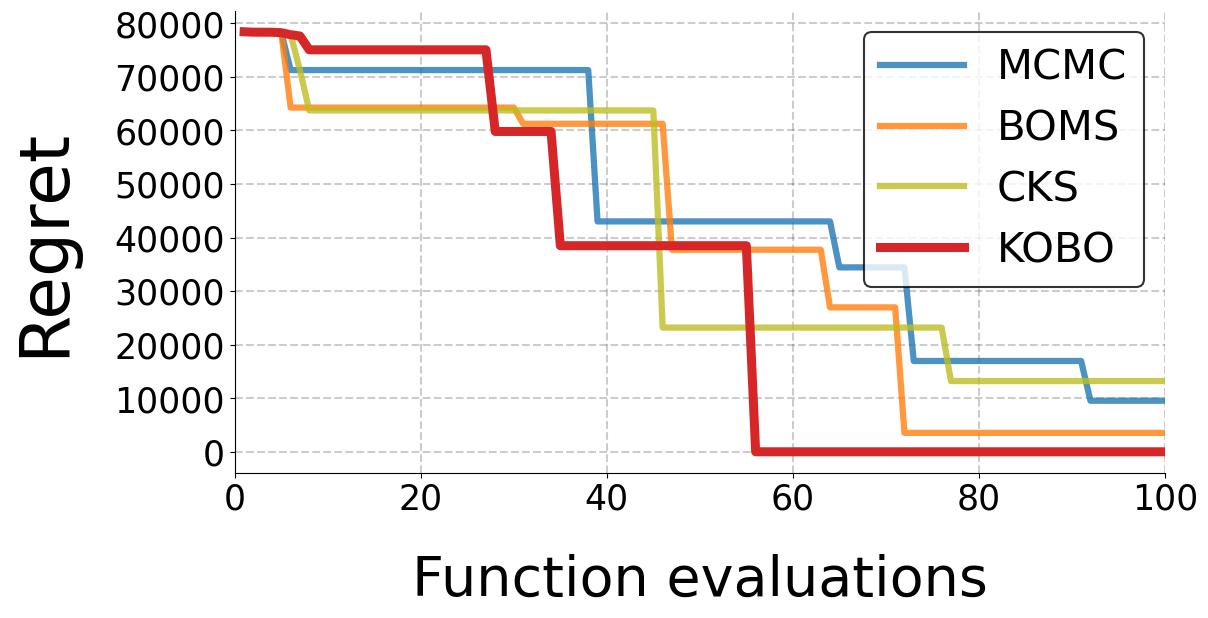}}
{\includegraphics[width=0.49\columnwidth]{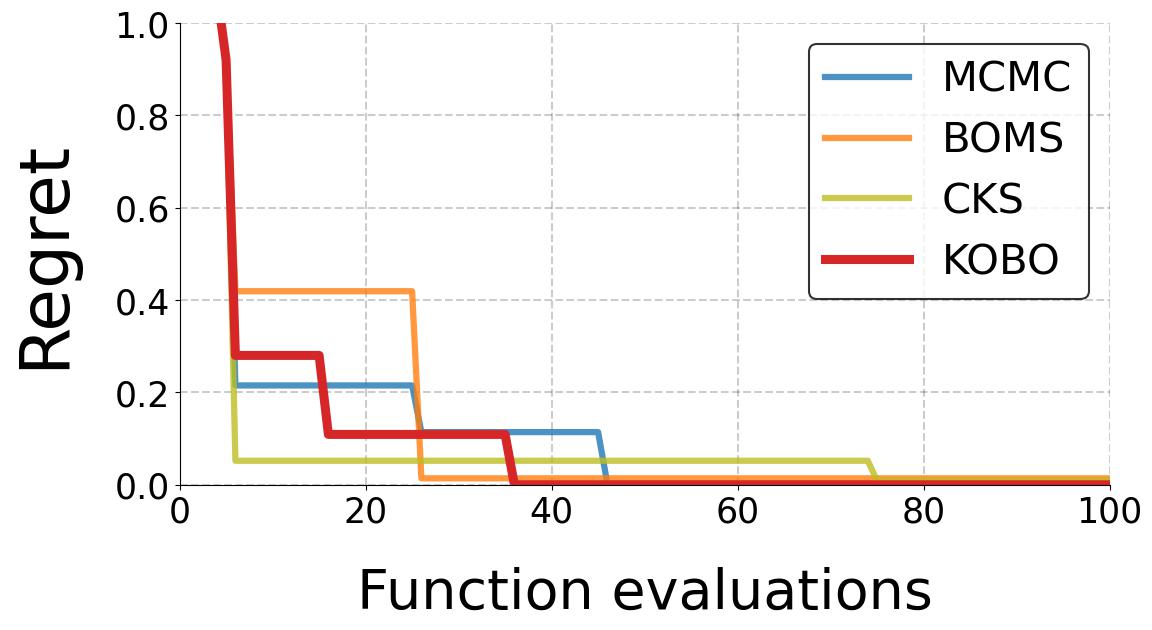}}
{\includegraphics[width=0.49\columnwidth]{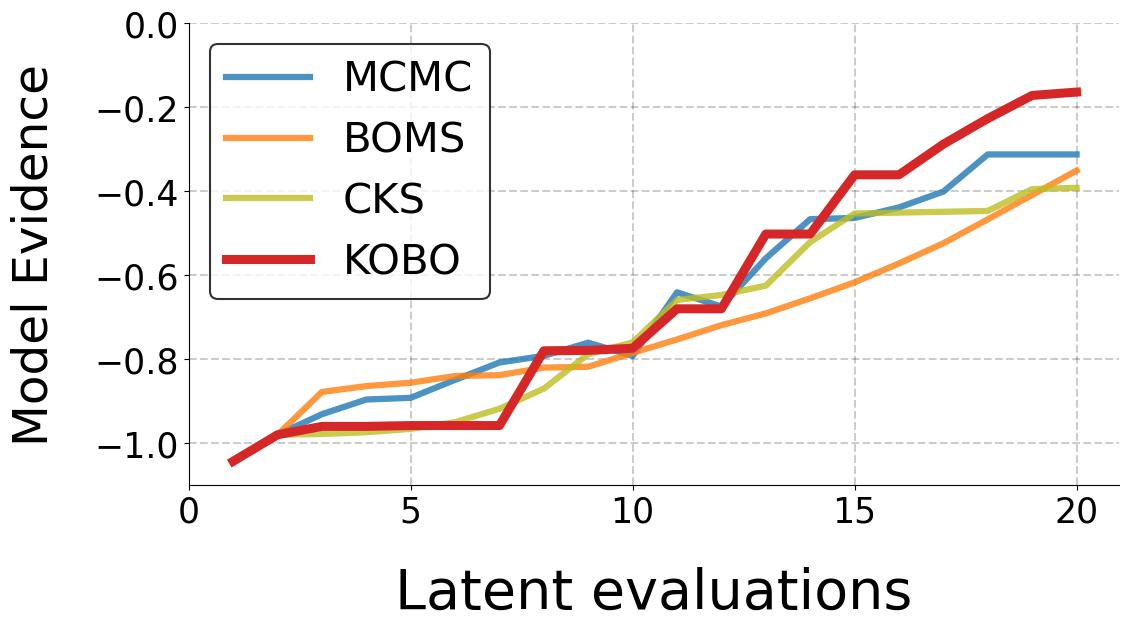}}
{\includegraphics[width=0.49\columnwidth]{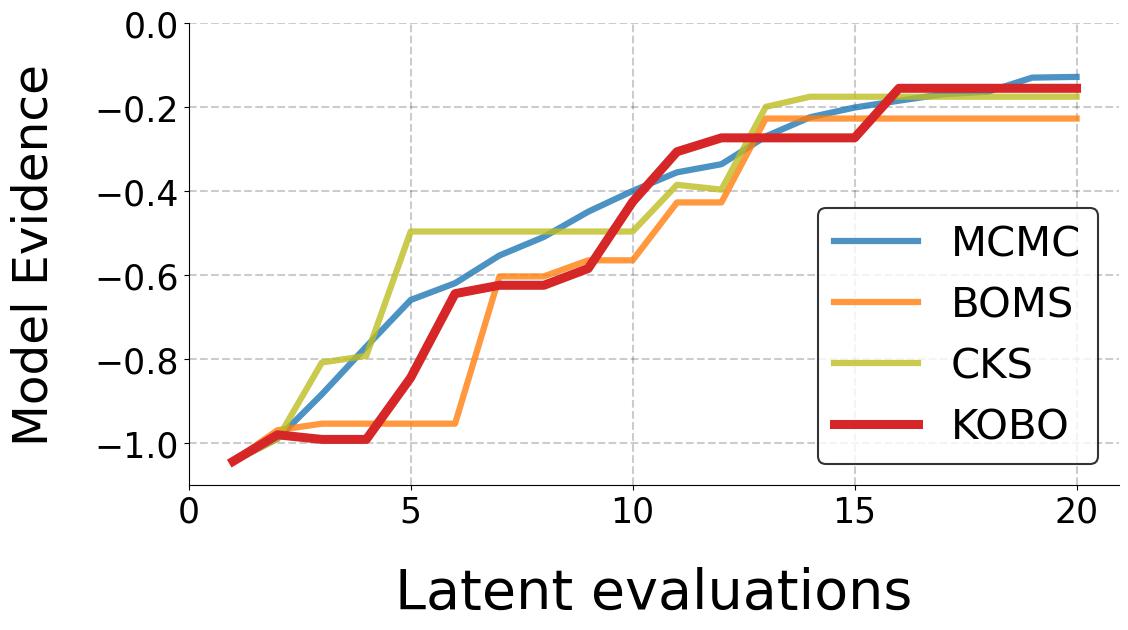}}
\caption{\textit{Comparison of {\name}, MCMC, CKS, and BOMS for Styblinski-Tang (Col 1), and Hartmann-6 (Col 2) functions:} (Row 1) \texttt{Regret} (Row 2) \texttt{Model Evidence}.}
\label{fig:R10}
\end{figure}
\textbf{With and without data-based representation:}
Data-based representation $r_d$ encodes the objective function's structure. 
Consider again the grammar-based codes in Table \ref{tab:encoding} -- they are similar in code space but since they model different kernels, they should be mutually further away in the latent kernel space. 
Figure 5(c)\&(d) visualize this; the latent space that also includes the data context will be tailored to the objective function, ensuring efficiency in KerGPR.

\textbf{{\name}'s performance for \texttt{Styblinski-Tang}, and \texttt{Hartmann-6}:}
Figure \ref{fig:R10}, shows results for two other benchmark functions \texttt{Styblinski-Tang} in 2000 dimensions, and \texttt{Hartmann-6} in 6 dimensions \cite{ssurjano}. These results are similar to those in Figure 7 in main paper. {\name} offers superior performance for \texttt{Styblinski-Tang} over other kernel methods: CKS< BOMS, and MCMC. For Hartmann in 6 dimensions, all methods perform relatively equally. Thus, we highlight the benefits of kernel learning for High-dimensional functions via continuous space optimization using VAEs.

\subsubsection{User Experiment: Audio Personalization}
\label{sec:app_aud}
We consider optimizing high-dimensional black-box functions in real-world applications such as audio personalization with strict sample budgets. For instance, today's hearing aids aim to filter the audio with $h$ so that the user's hearing loss is compensated. We aim to perform hearing aid tuning by estimating a high-resolution personal frequency filter $h^*$ such that the user satisfaction $f(h)$ is maximized if users are willing to listen and rate some audio clips ($Q$ queries) prescribed by {\name} at home.

We recruit $6$ volunteers of $4$ male(s) and $2$ female(s) with normal hearing for the personalization experiment. We apply two types of corruption to the audio played to the volunteers. 

First, we want to emulate hearing loss. We do this by employing the publicly available hearing loss profiles in the NHANES \cite{nhanes} database as the corrupting filter $b_1$. Second, to emulate cheap speakers, we generate random distortions by creating a random corrupting filter $b_2$, each $b_2[j]$ selected independently from [$-30, 30$]dB. A sample speech clip $a$ is filtered with the distorting filter $b_1$ or $b_2$ to obtain the \texttt{corrupted} clip, $r = b_{1:2} * a$. 

The goal of the audio personalization task is to find the filter $\hat{h}^*$ which when applied to the corrupted clip $r$ should make the resulting audio sound similar to the original uncorrupted audio clip, i.e., $r * \hat{h}^* = \hat{a} \approx a$. In other words, the personalization filter $\hat{h}^*$ should counteract the distortion caused by $b_1$ or $b_2$. 

The user satisfaction function $f(h)$ is expensive to evaluate as user feedback is finite and hence has strict sample constraints. As human hearing is not uniform across all audio frequencies \cite{pitch}, a user's audio perception might not change for a range of filters so the satisfaction score remains the same and might change with sudden jumps for some filter choices, thus leading to a satisfaction function $f(h)$ with a flat shape in some regions and a steep curve in other regions, i.e., a staircase shape.  Thus, the audio personalization problem is well suited for kernel learning BO methods like \name.

\textbf{User Querying}: We query the volunteers and use their feedback scores to construct the user satisfaction function. This is done by choosing filters $h_j$ (prescribed by $f$GPR) from the space of all filters $\mathcal{H} \subseteq \mathbf{R}^N$ ($N=4000$). The filter is applied to the audio played to the user, and their score $f(h_j)$ is recorded. Repeating this process, we obtain the black-box user satisfaction function. This function can then be optimized to find $\hat{h}^*$, the personal filter.

\begin{table}[t]
\vspace{-0.1in}
\centering

\begin{adjustbox}{width=1.0\columnwidth}
\begin{tabular}{|c|c|c|c|c|c|}
\hline
$x_{start}$ & $x_{Q=5}$ & $x_{Q=10}$ & $x_{Q=15}$ & $x_{Q=20}$& $x_{Q=25}$\\\hline

\multicolumn{6}{|c|}{Prompt: Office room with a desk, a blue chair, a lamp on the desk, and a green couch on the side} \\\hline

\includegraphics[width=0.35\columnwidth]{office_run2_25.jpg} & \includegraphics[width=0.35\columnwidth]{office_img_5.jpg}& \includegraphics[width=0.35\columnwidth]{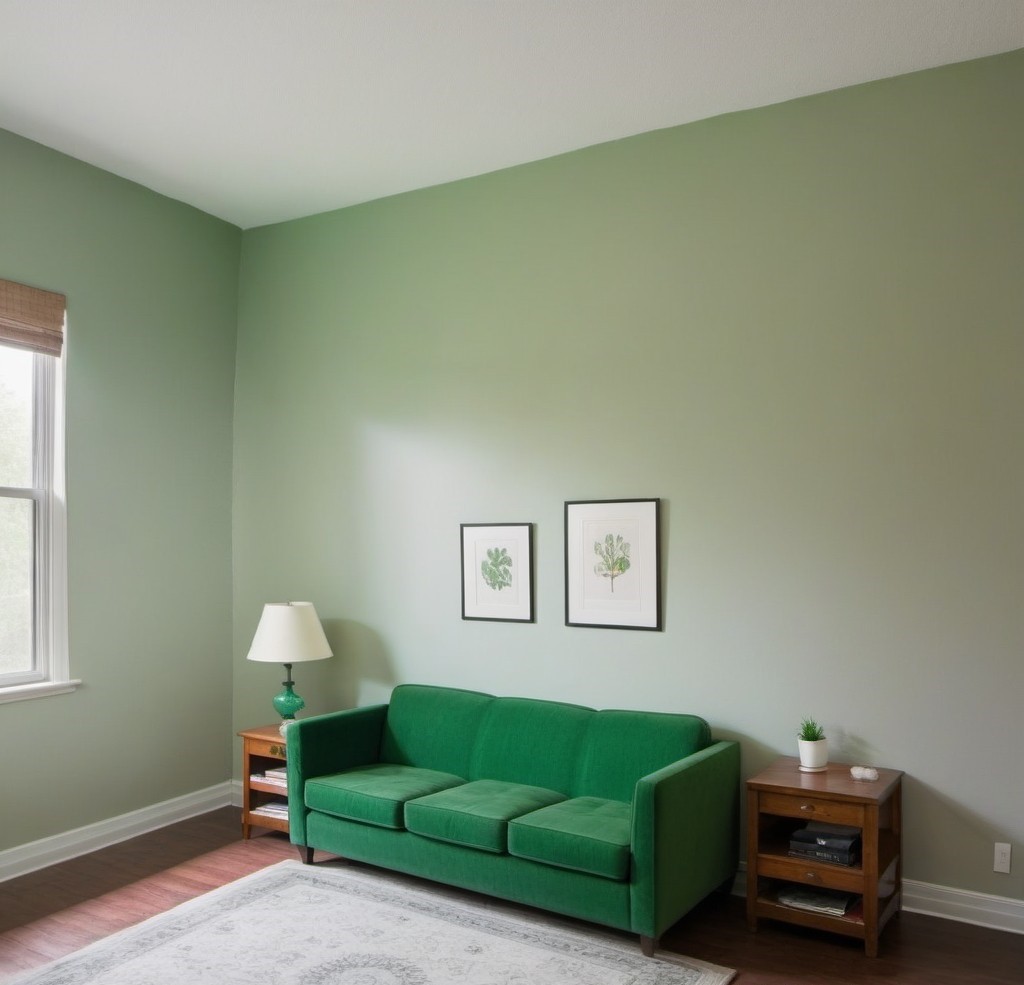} & \includegraphics[width=0.35\columnwidth]{office_img_15.jpg}& \includegraphics[width=0.35\columnwidth]{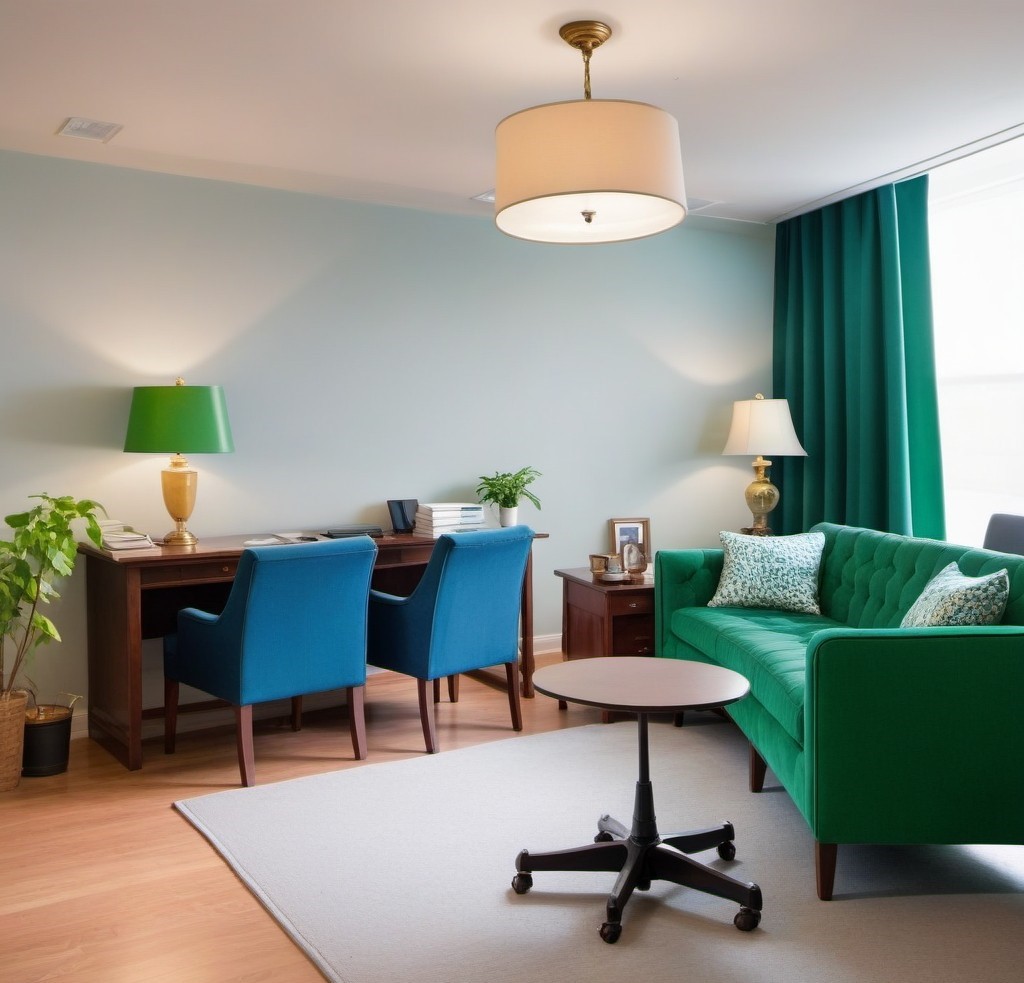}& \includegraphics[width=0.35\columnwidth]{office_img_25.jpg}\\

\hline

\multicolumn{6}{|c|}{Prompt: Garden with a fountain in the center and surrounded by trees and flowers of different colors} \\\hline

\includegraphics[width=0.35\columnwidth]{garden_test.jpg} & \includegraphics[width=0.35\columnwidth]{garden_img_5.jpg}& \includegraphics[width=0.35\columnwidth]{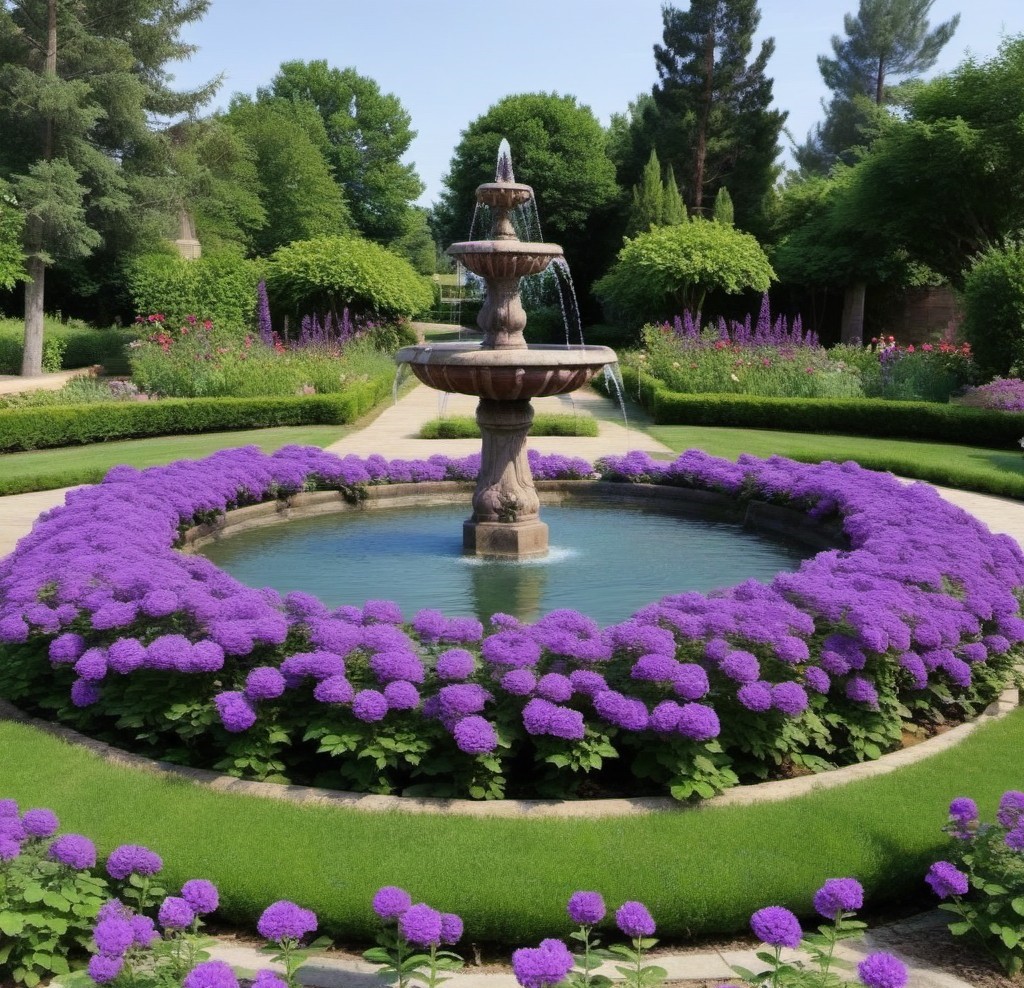} & \includegraphics[width=0.35\columnwidth]{garden_img_15.jpg}& \includegraphics[width=0.35\columnwidth]{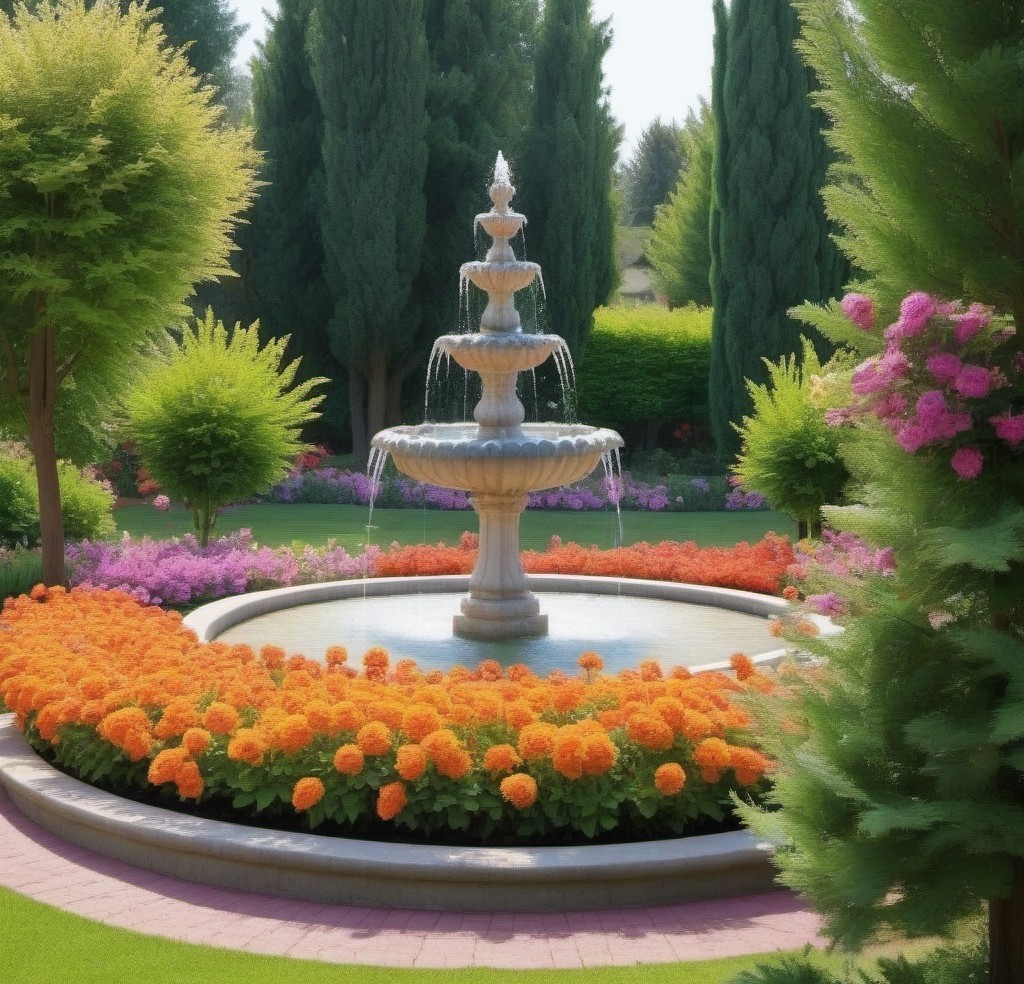}& \includegraphics[width=0.35\columnwidth]{garden_run5_25.jpg}\\

\hline

\multicolumn{6}{|c|}{Prompt: Fair with a darts game booth, a giant wheel, a cotton candy stand, a carousel with children riding and parents standing around} \\\hline

\includegraphics[width=0.35\columnwidth]{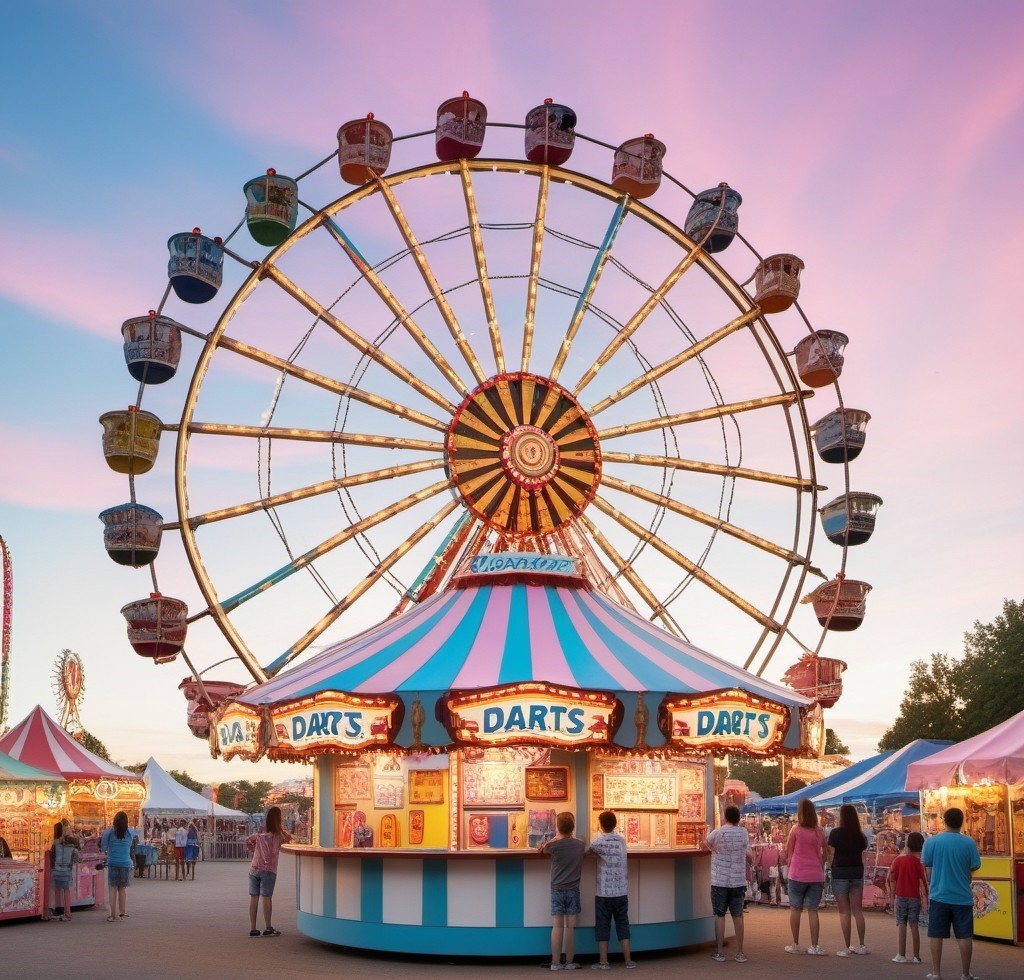} & \includegraphics[width=0.35\columnwidth]{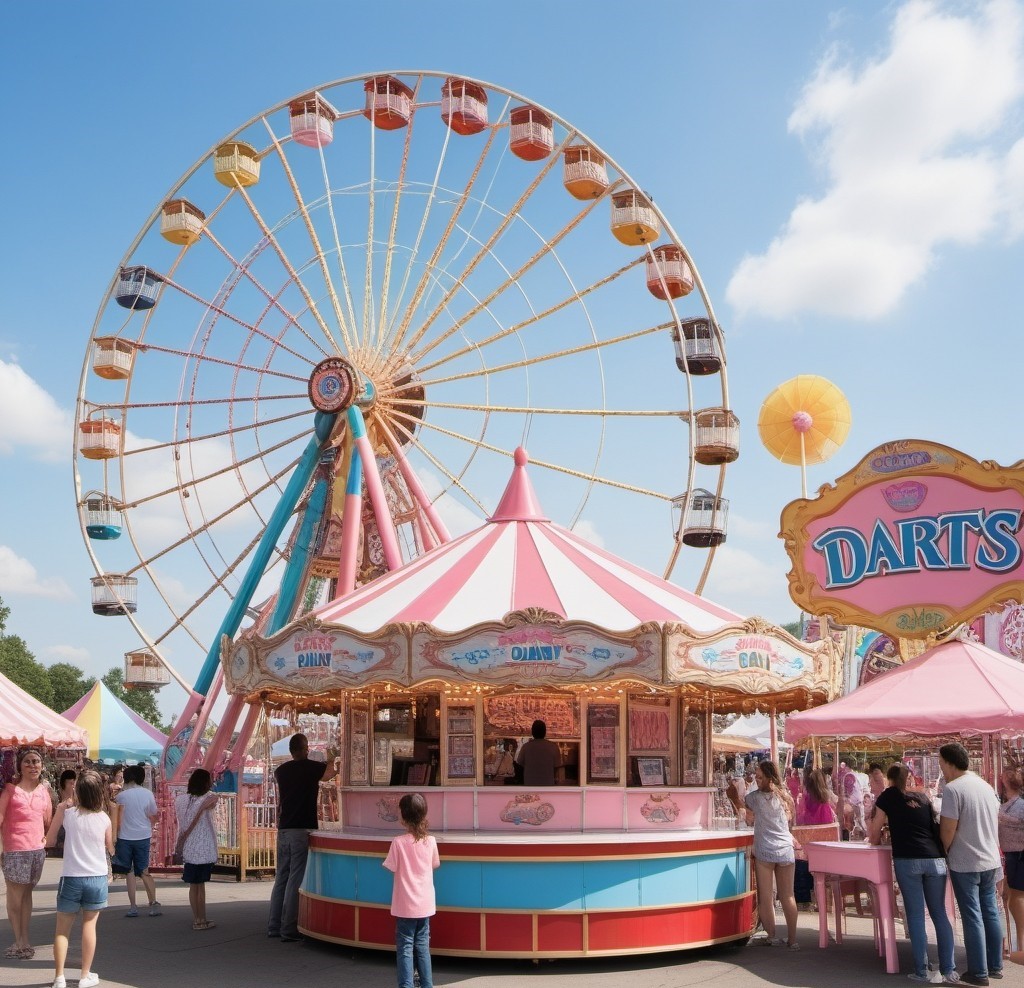}& \includegraphics[width=0.35\columnwidth]{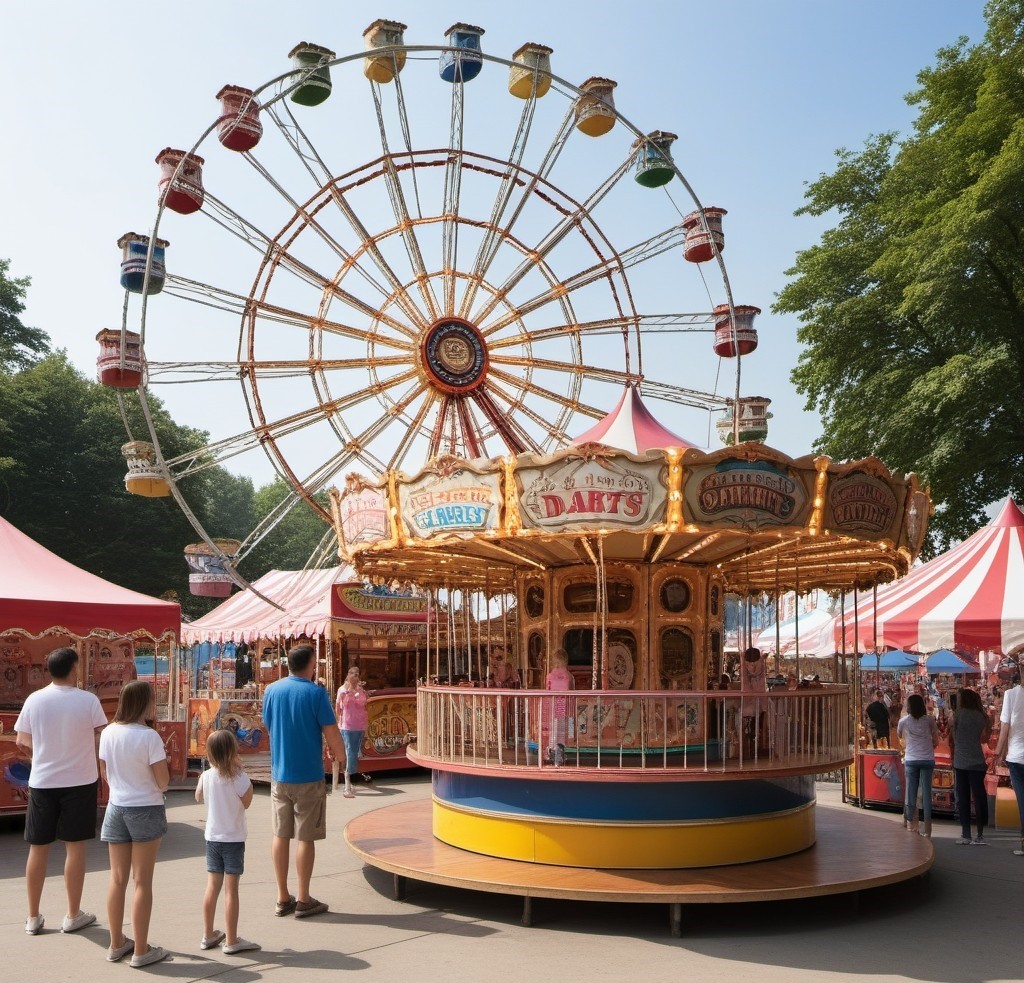} & \includegraphics[width=0.35\columnwidth]{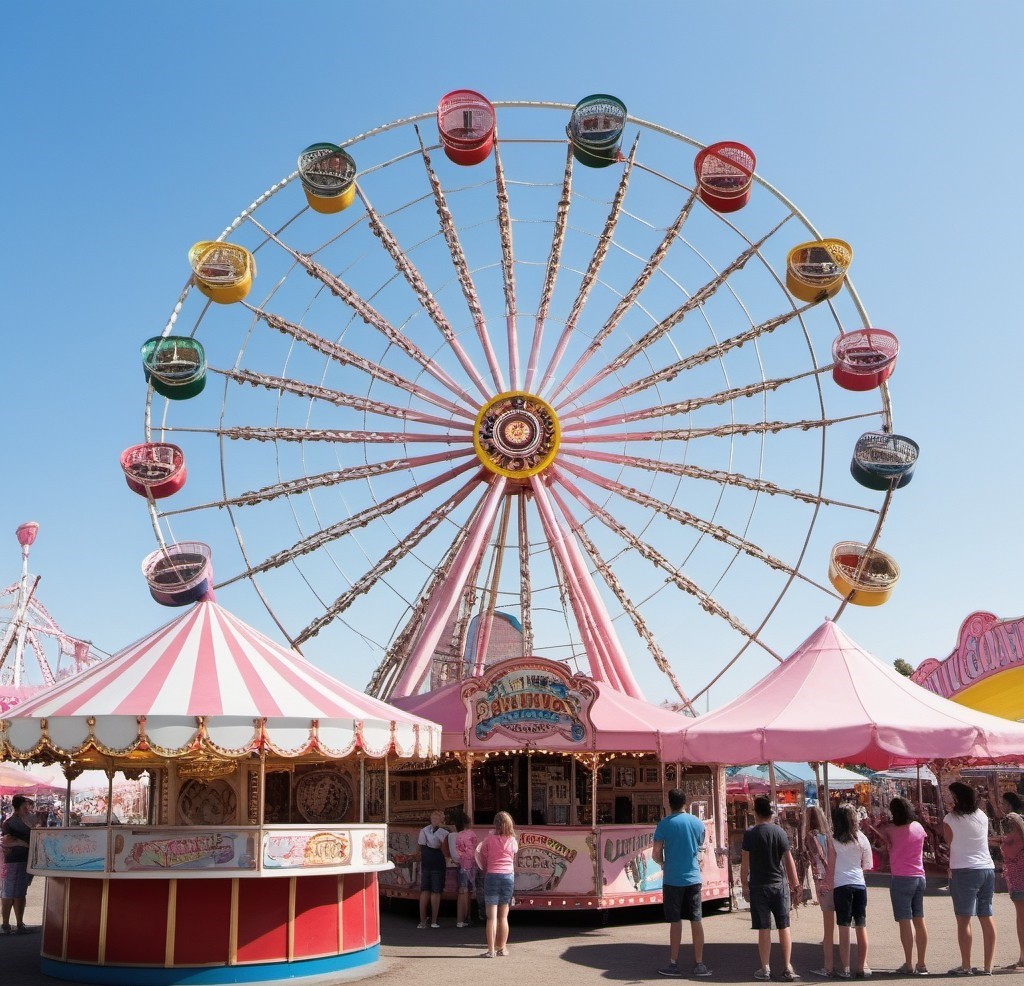}& \includegraphics[width=0.35\columnwidth]{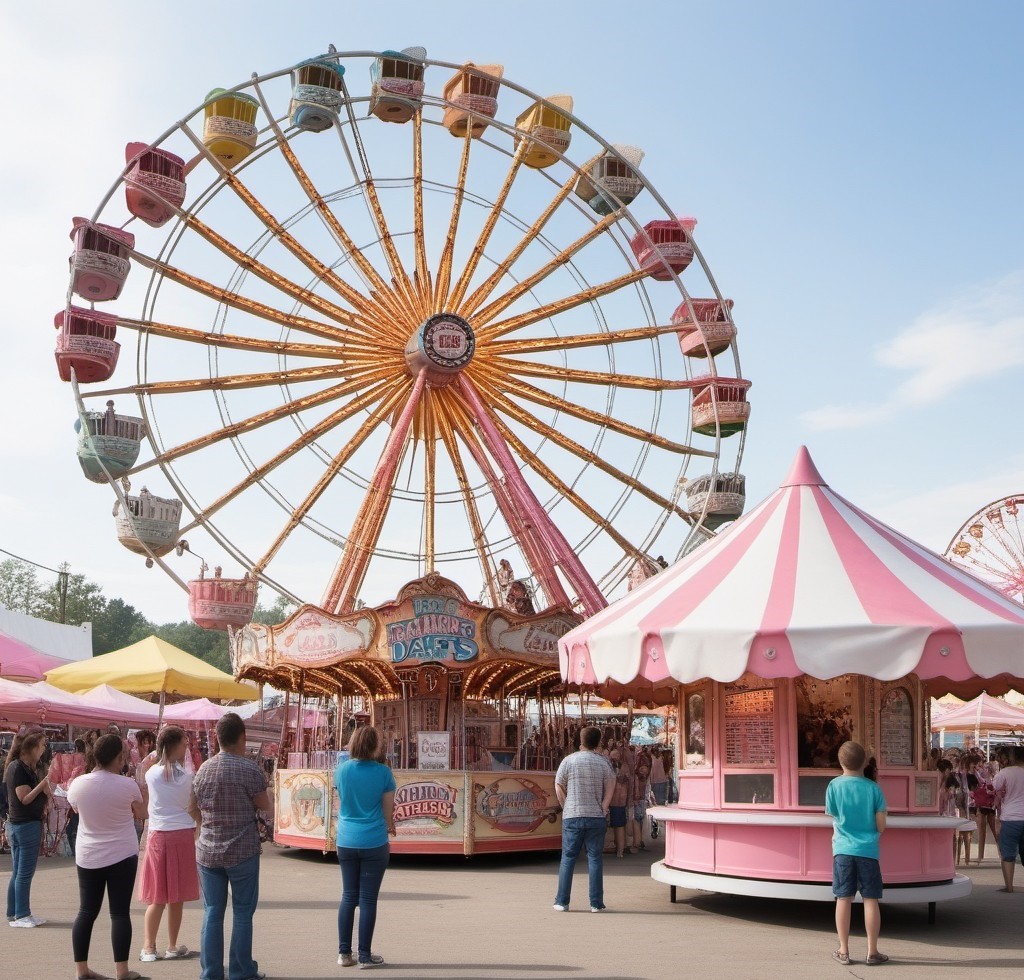}& \includegraphics[width=0.35\columnwidth]{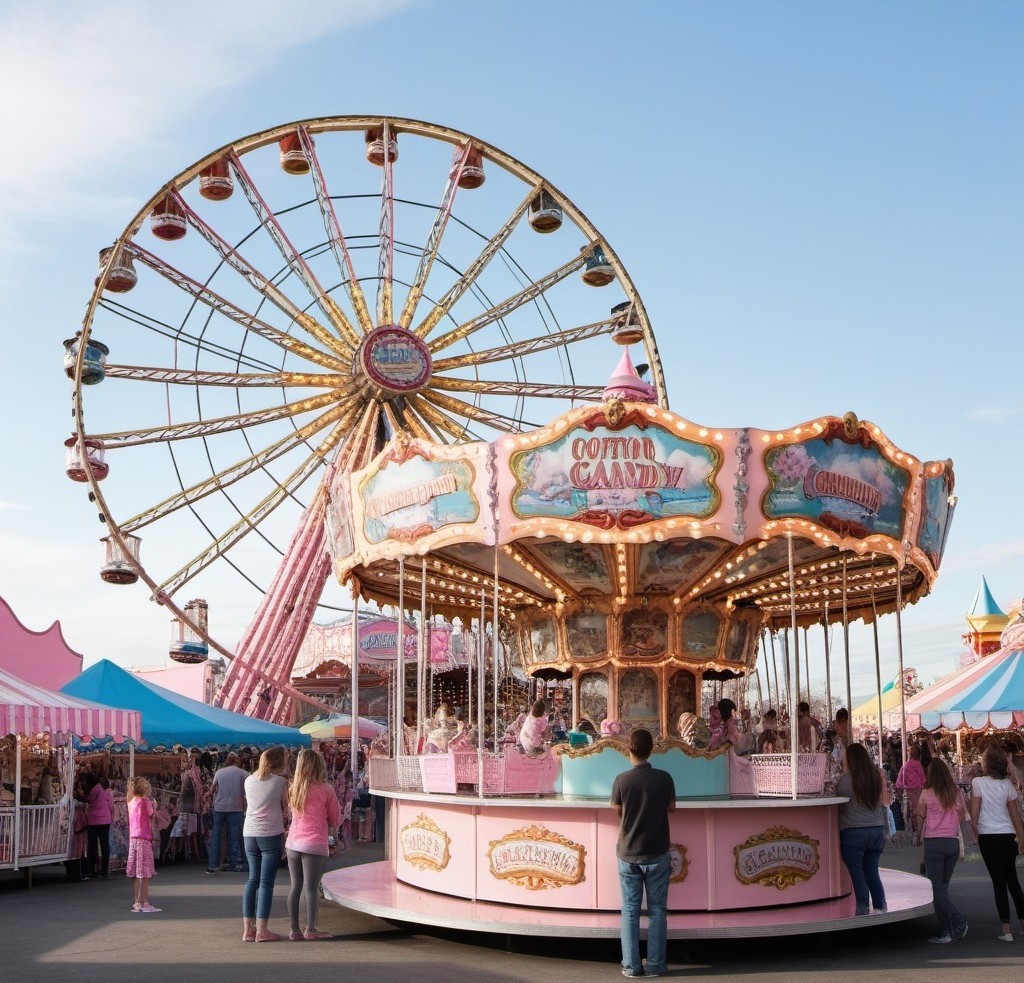}\\

\hline

\multicolumn{6}{|c|}{Prompt: Garden with a bird nest, flowers, a greenhouse, a sitting bench, and a statute with water} \\\hline

\includegraphics[width=0.35\columnwidth]{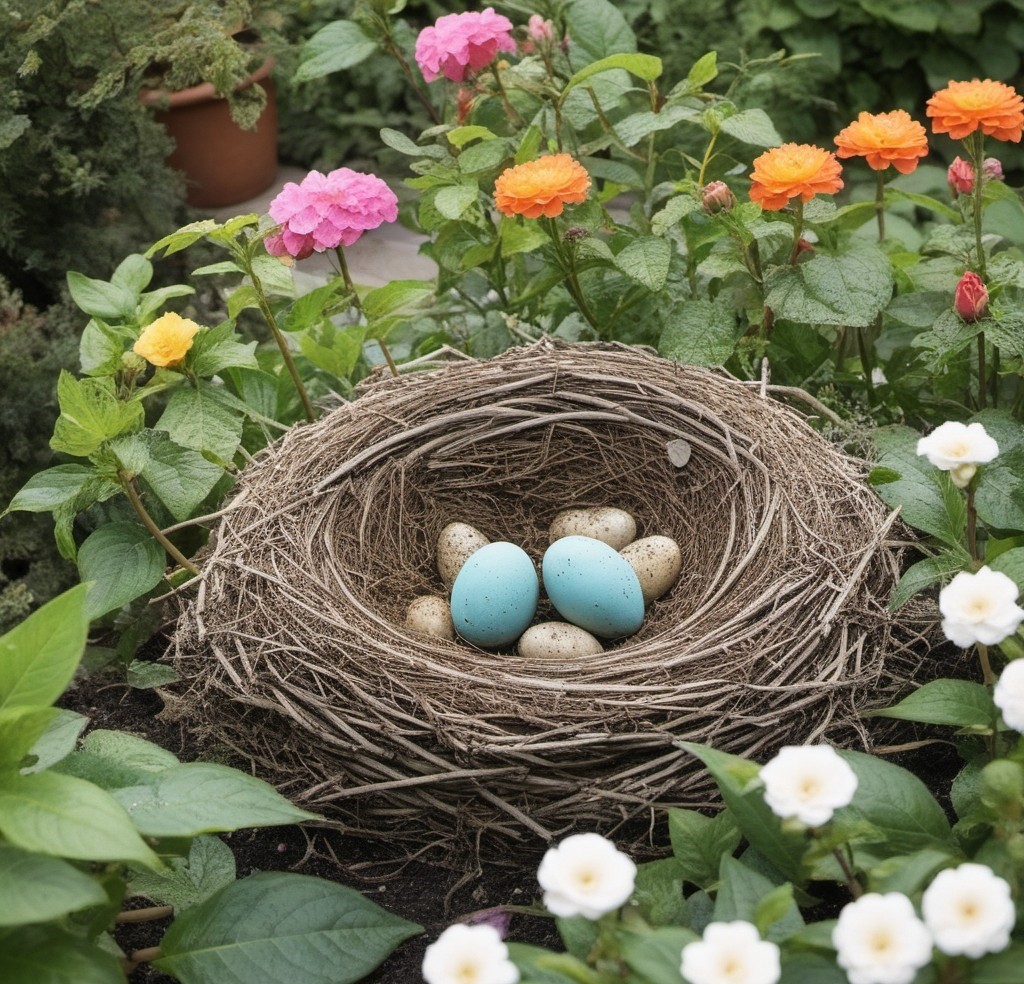} & \includegraphics[width=0.35\columnwidth]{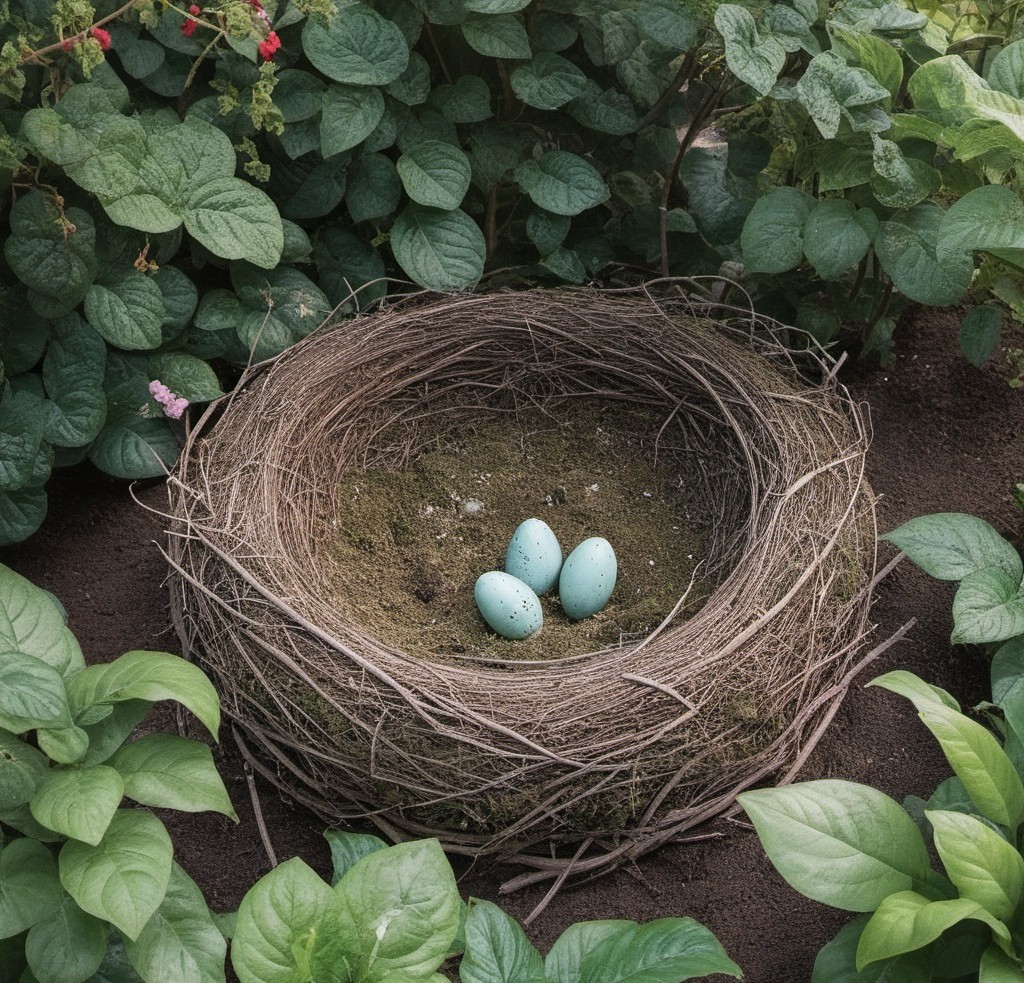}& \includegraphics[width=0.35\columnwidth]{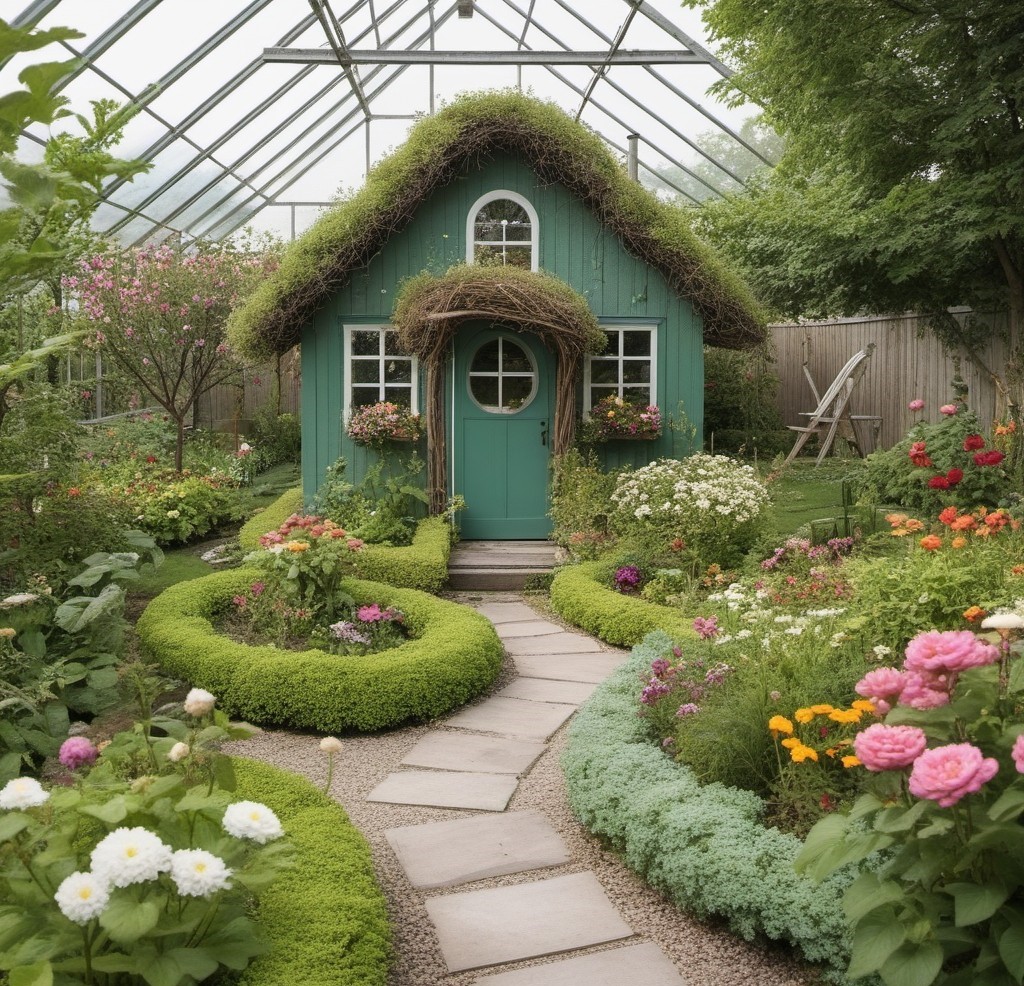} & \includegraphics[width=0.35\columnwidth]{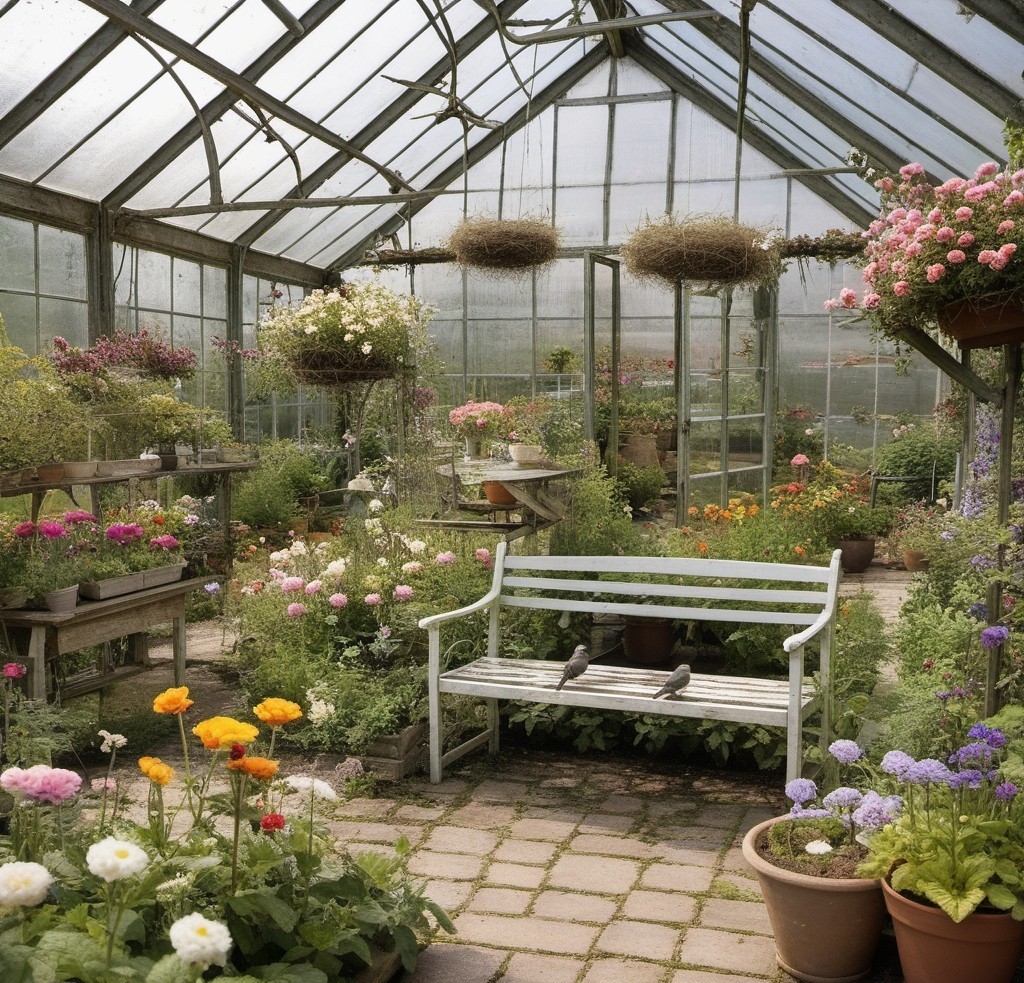}& \includegraphics[width=0.35\columnwidth]{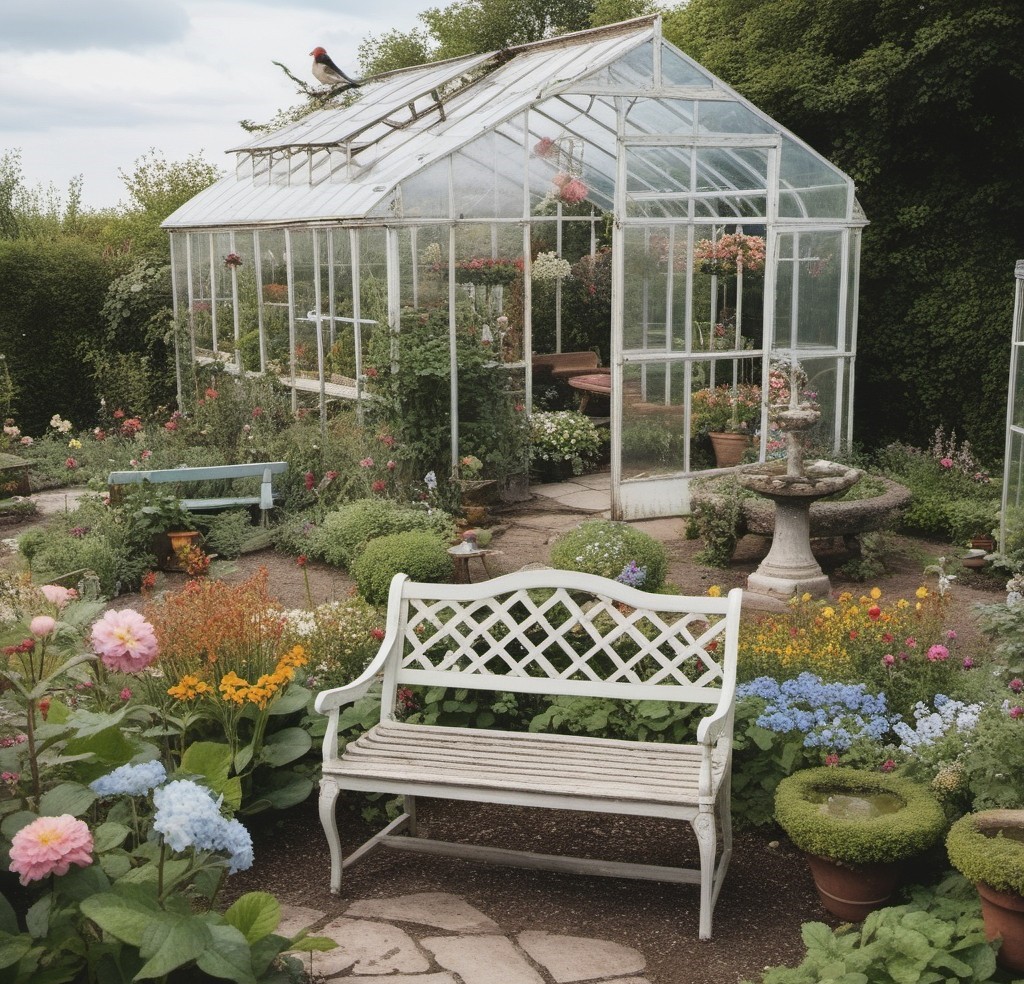}& \includegraphics[width=0.35\columnwidth]{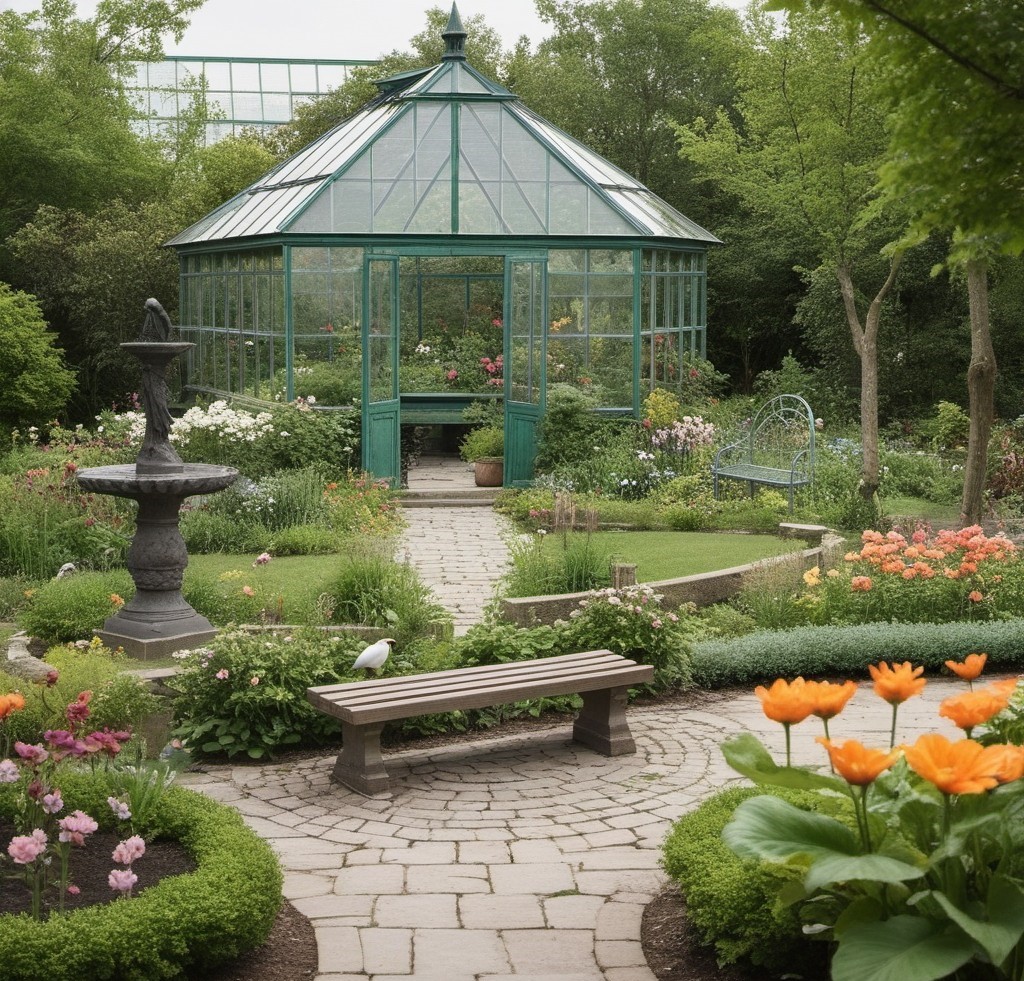}\\

\hline

\multicolumn{6}{|c|}{Prompt: Deep waterfall, a cloudy sky, rain, and kids jumping into the waterfall wearing red clothes} \\\hline

\includegraphics[width=0.35\columnwidth]{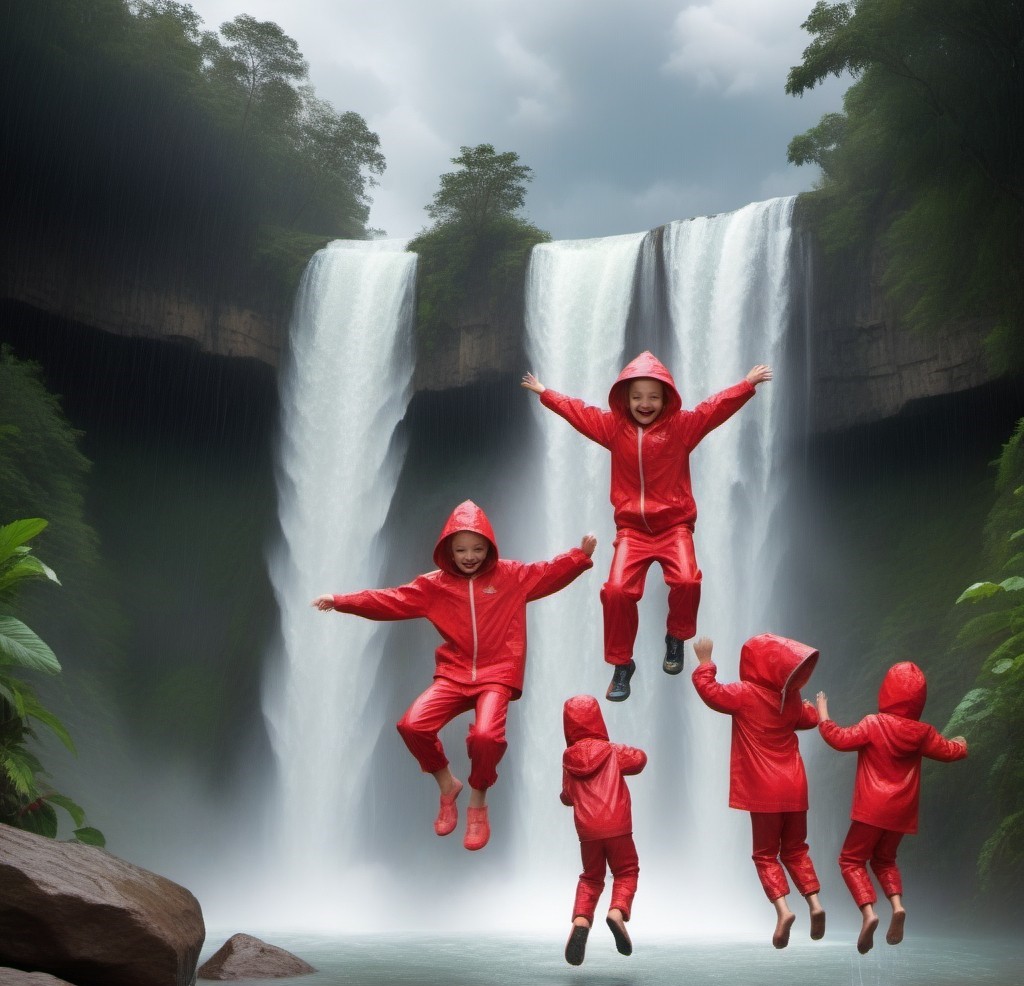} & \includegraphics[width=0.35\columnwidth]{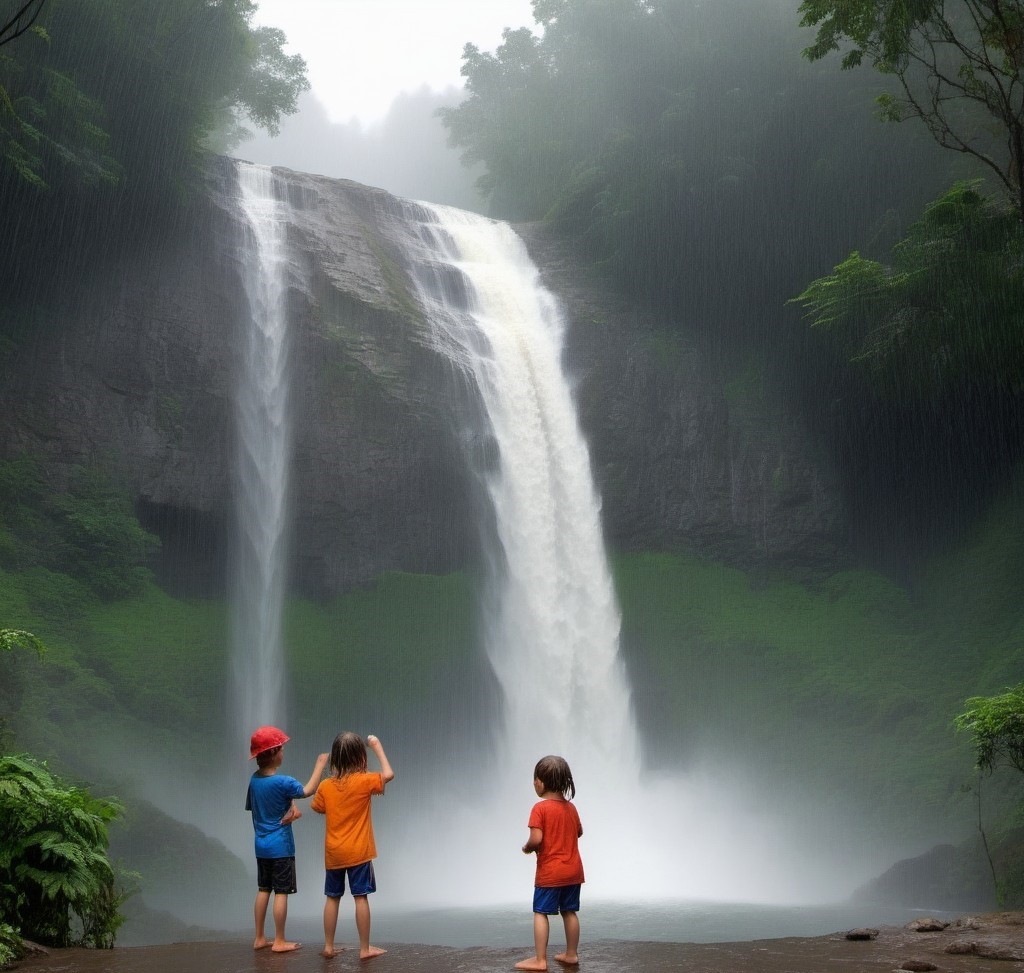}& \includegraphics[width=0.35\columnwidth]{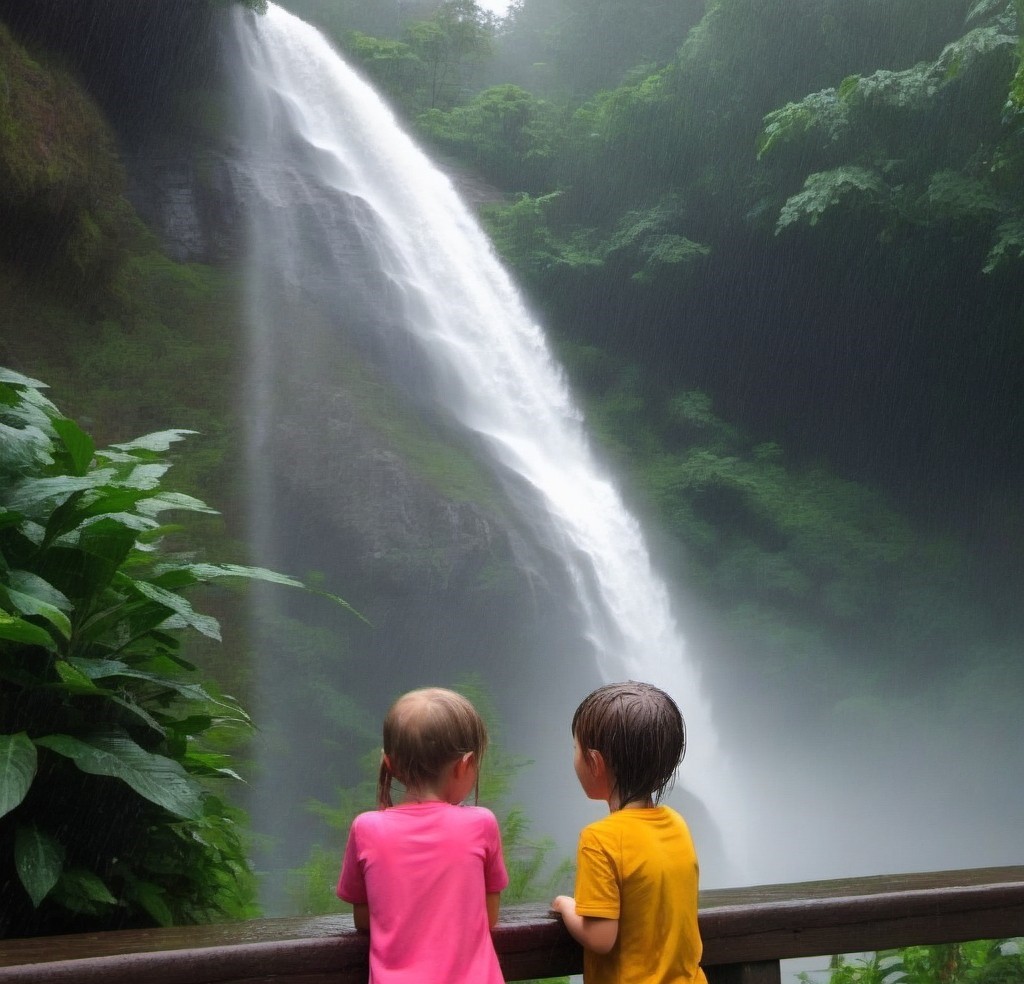} & \includegraphics[width=0.35\columnwidth]{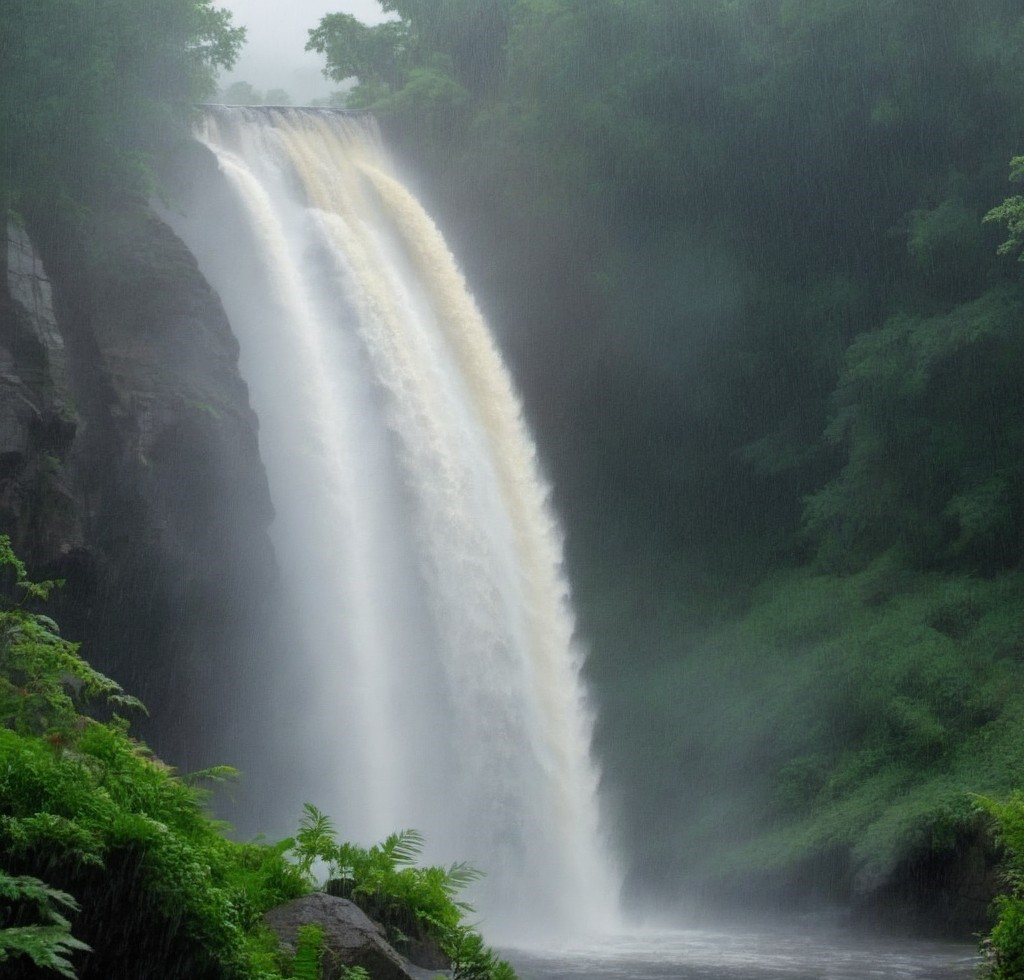}& \includegraphics[width=0.35\columnwidth]{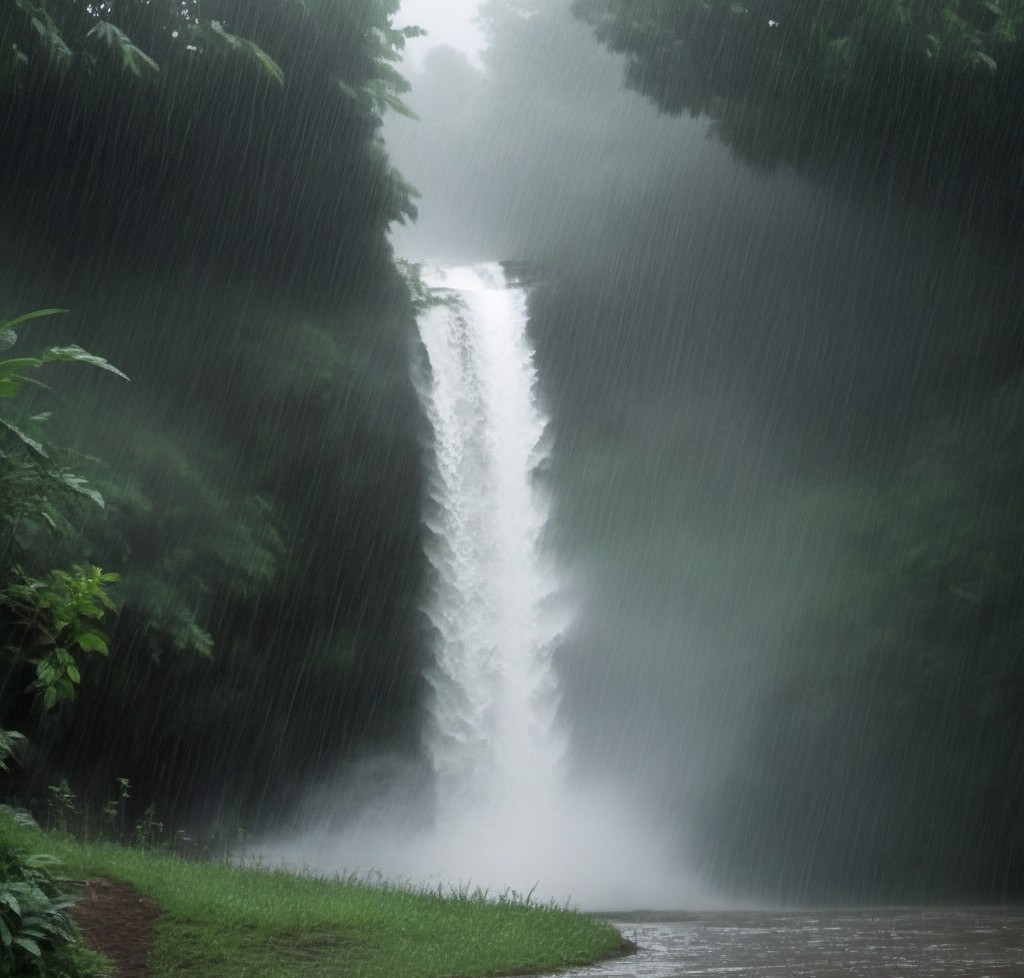}& \includegraphics[width=0.35\columnwidth]{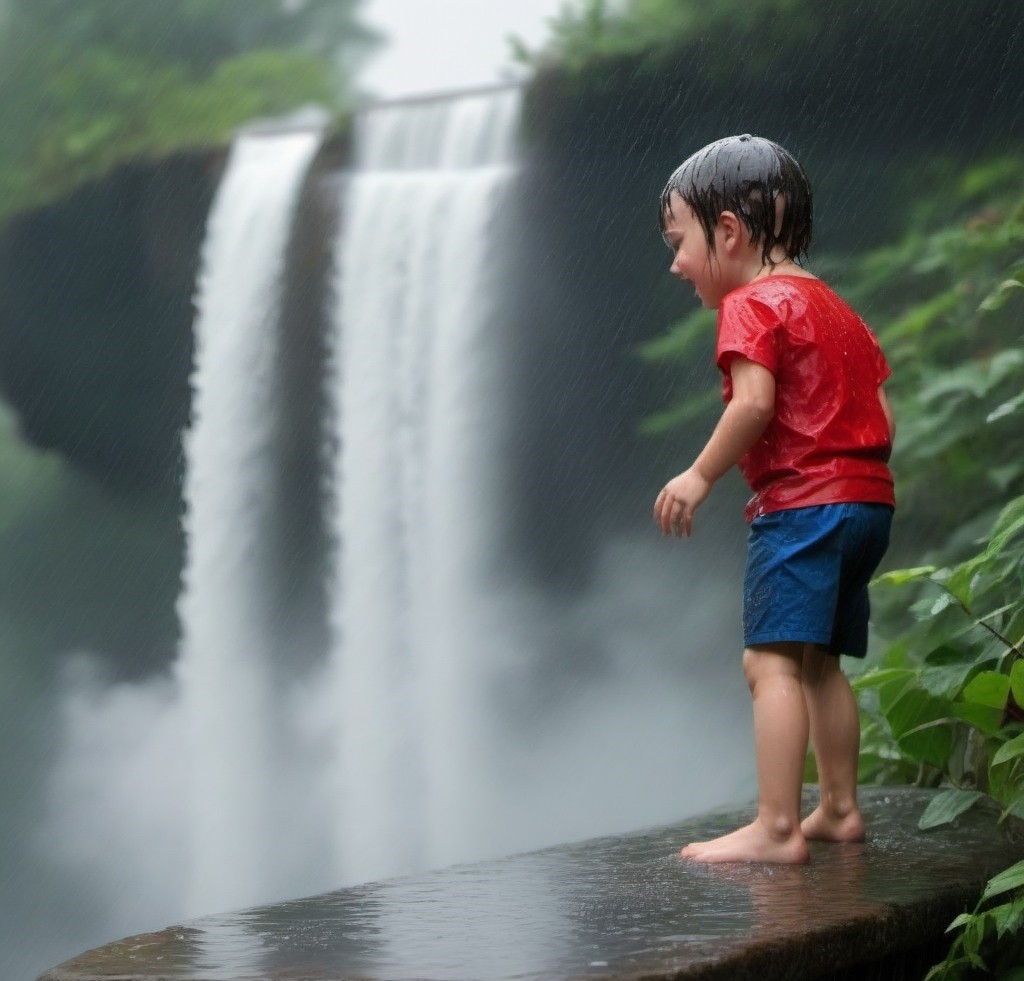}\\

\hline

\multicolumn{6}{|c|}{Prompt: Picnic in a park with a red and white checkered blanket, a wicker basket, picnic sandwiches, drinks, and two children} \\\hline

\includegraphics[width=0.35\columnwidth]{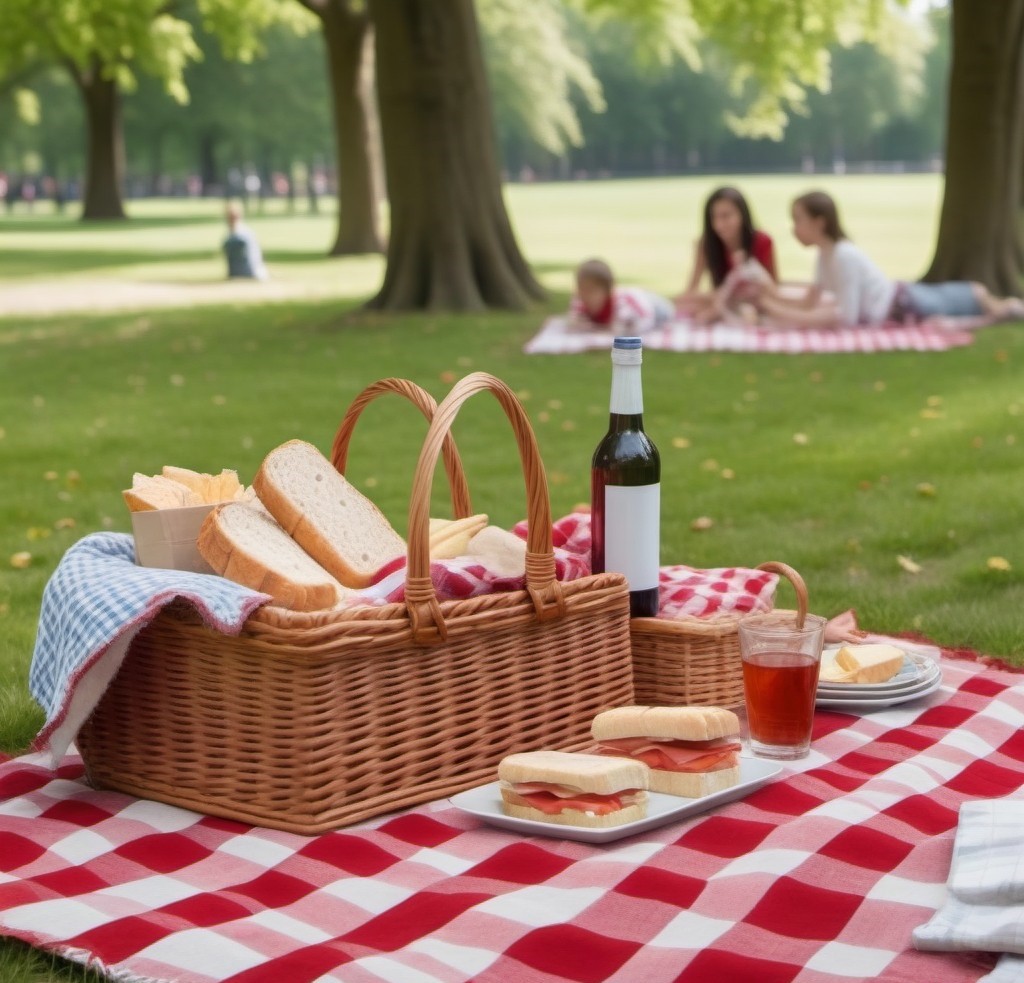} & \includegraphics[width=0.35\columnwidth]{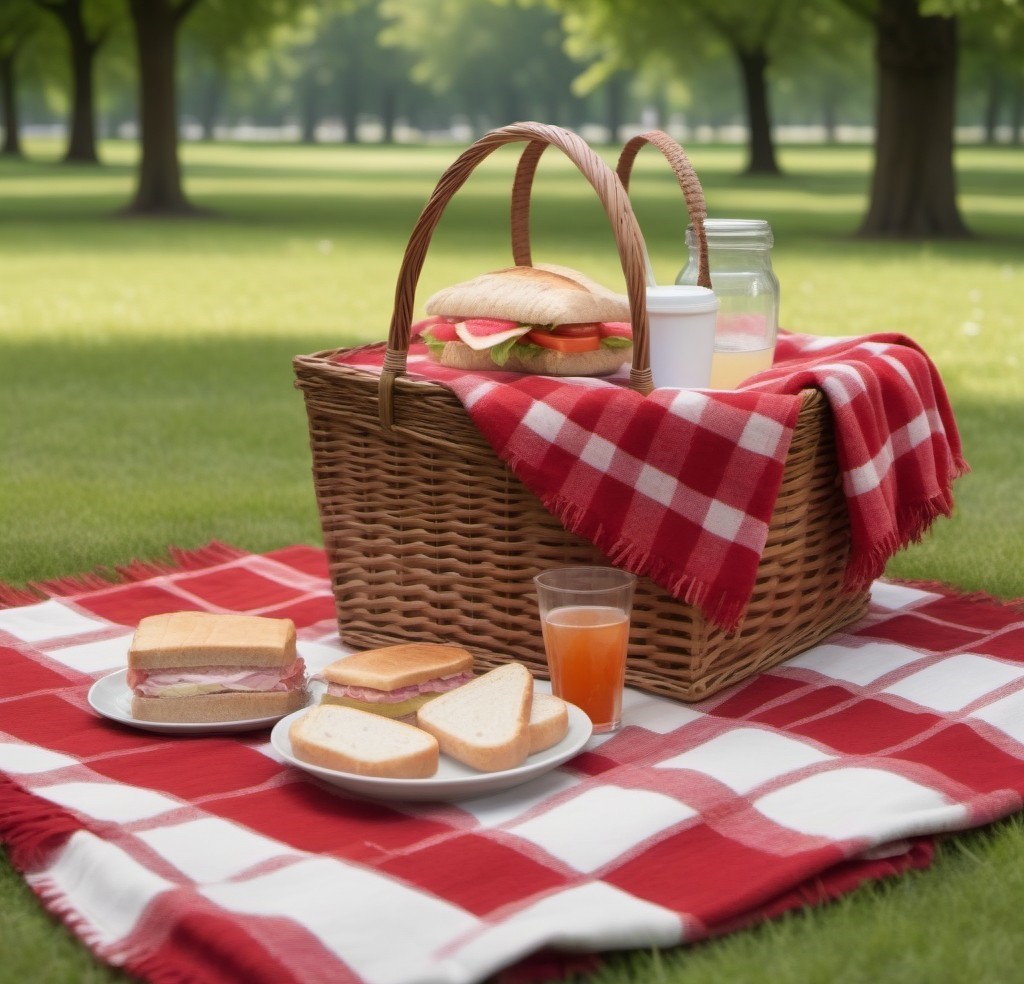}& \includegraphics[width=0.35\columnwidth]{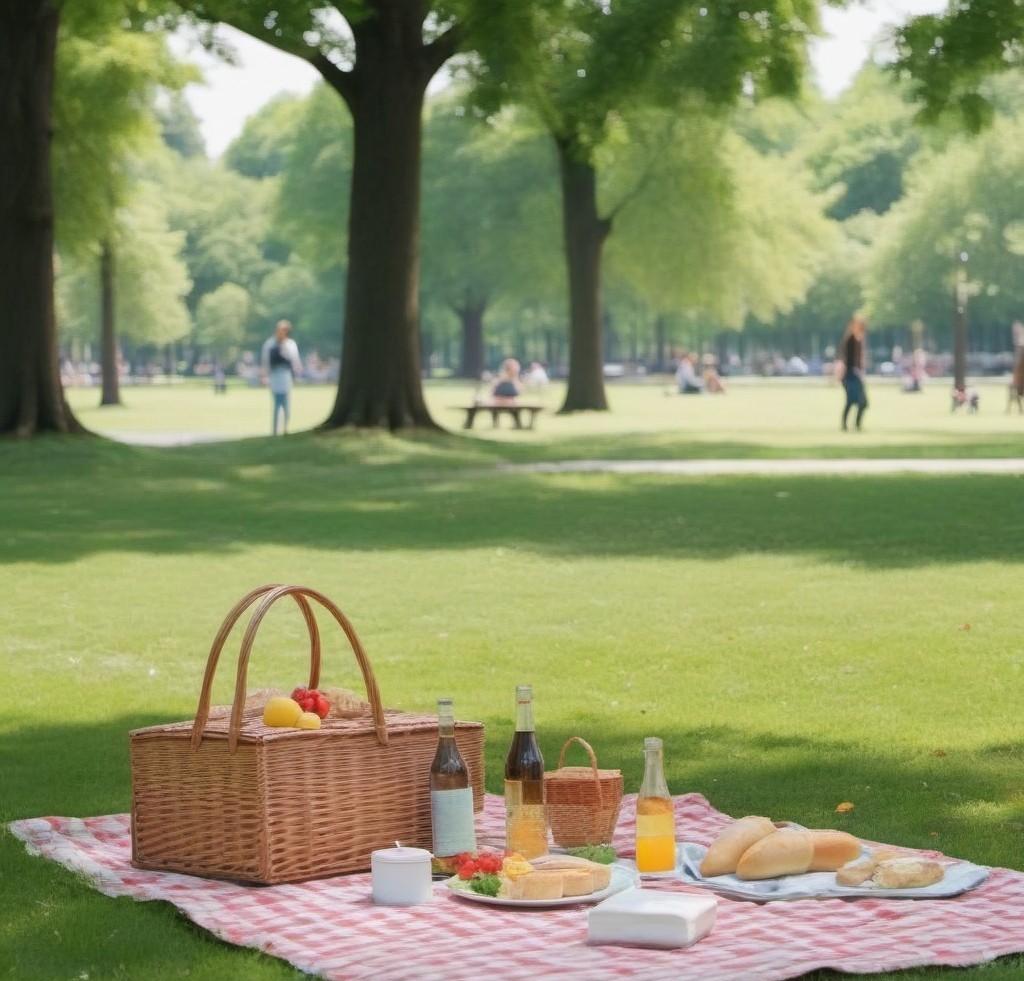} & \includegraphics[width=0.35\columnwidth]{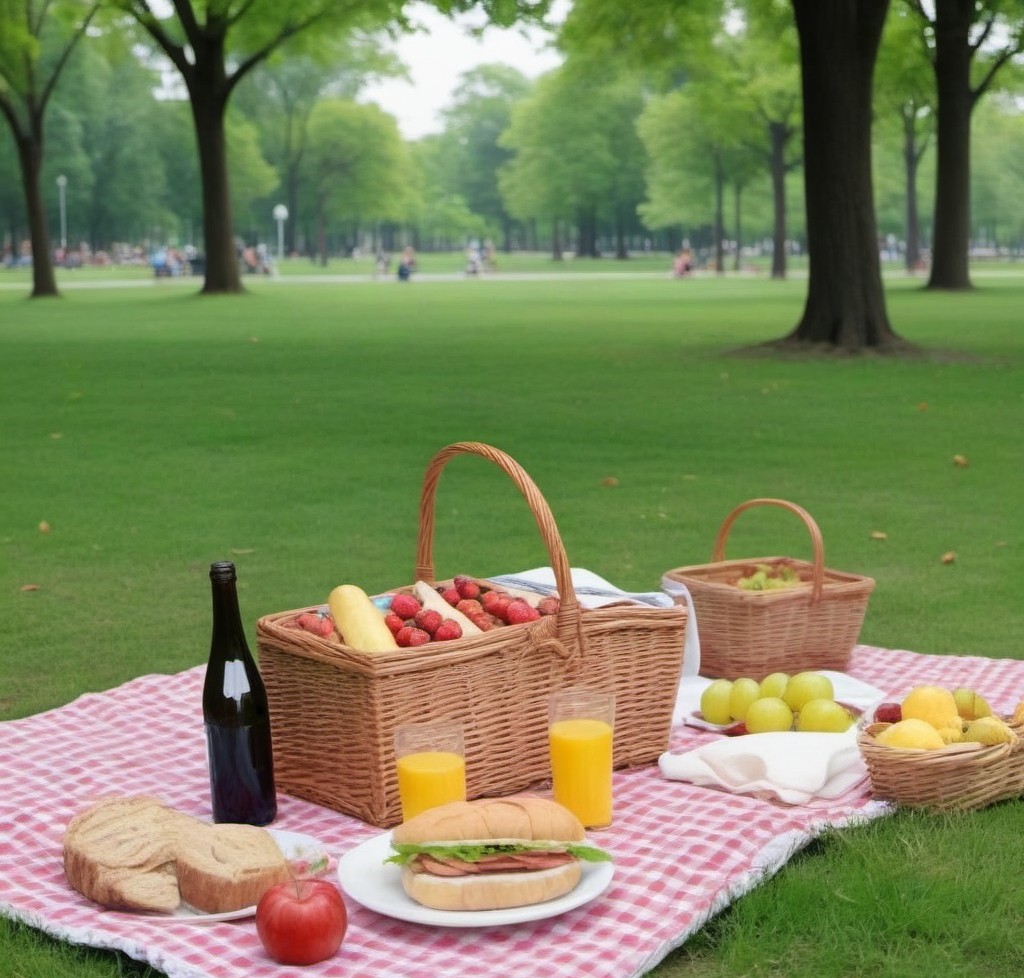}& \includegraphics[width=0.35\columnwidth]{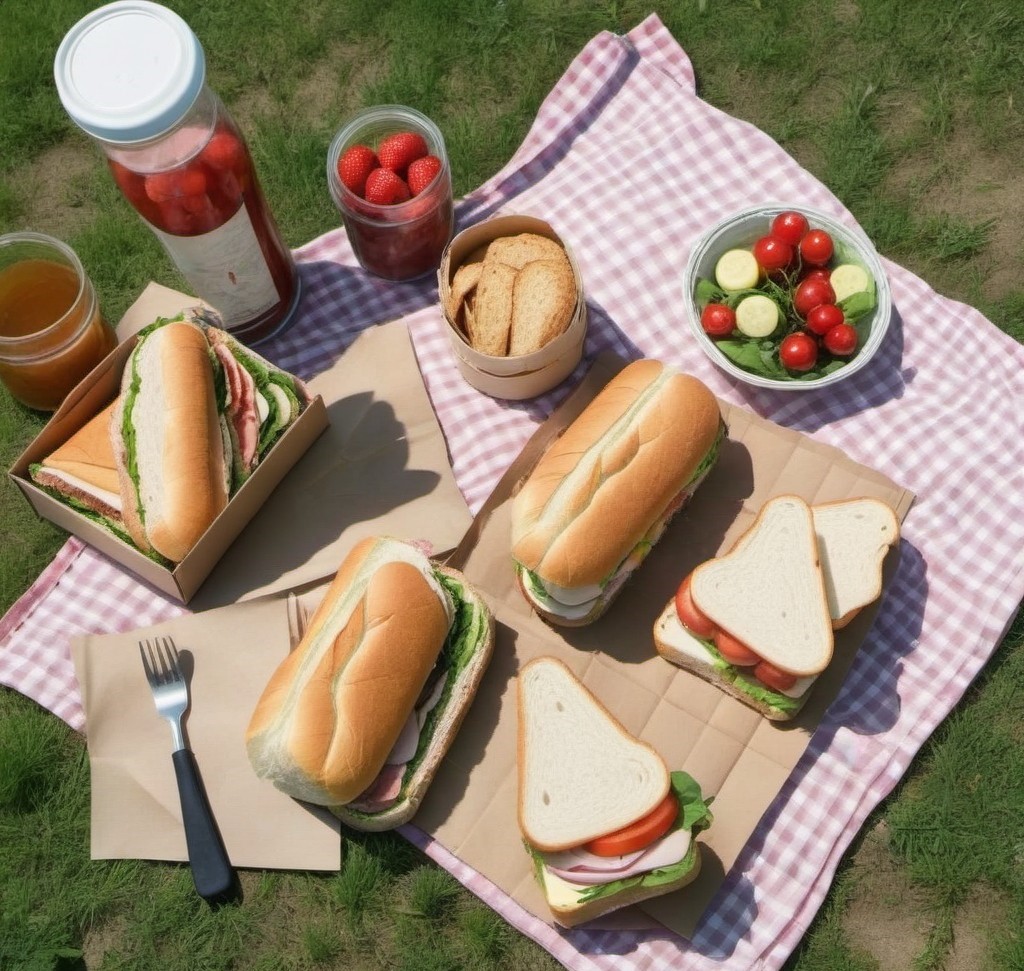}& \includegraphics[width=0.35\columnwidth]{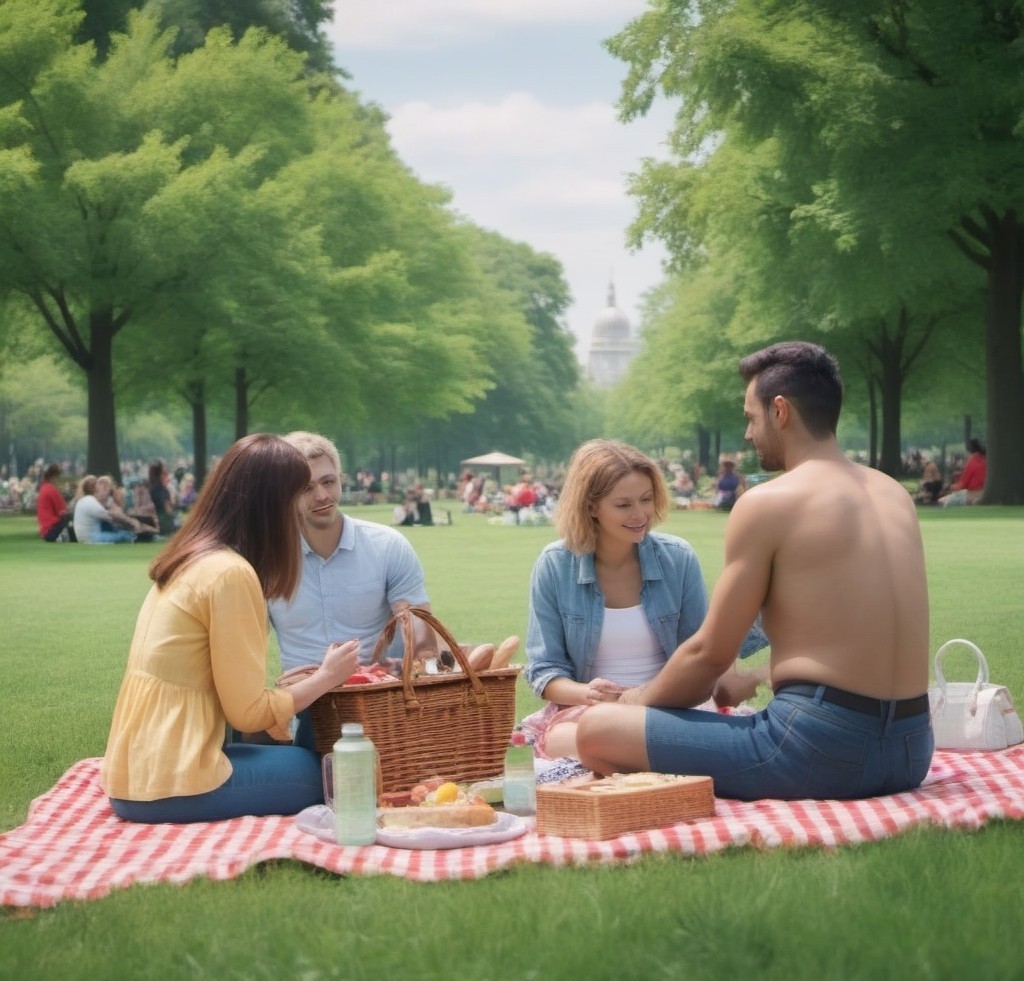}\\

\hline

\end{tabular}
\end{adjustbox}
\caption{Prompt-based image generation user results}
\label{tab:image}
\vspace{-0.2in}

\end{table}

\subsubsection{User Experiment: Prompt-based Image Generation}
\label{sec:app_img}

Generating images that satisfy a particular prompt is a popular application for image generation models. Oftentimes, users are unable to articulate precisely what images they want generated, at most they can look at an image provided and offer their satisfaction of whether it meets their taste. Moreover, even if the requirement can be structured as a textual prompt, the prompt may be too long for a prompt-based generator to handle (generators only focus on some parts of the prompt). 

Prompt engineering has been one avenue of research for writing effective prompts. In this section, we explore augmenting broad image generative models --- that may or may not be trained to handle text-based prompts --- to produce/recommend images that fit a user's requirement (or ``prompt'')\footnote{We provide a user prompt here to illustrate if {\name} has satisfied the user requirement which is difficult to textualize. This "prompt" is the user's requirement that they want to see in the images and score accordingly.}
This prompt-based image recommender can be formulated as a black-box optimization problem by posing the question as: \textit{``if a user rates few pictures generated by the model, can the model find the ``best" picture from the space of all images?" }

A user thinks of a particular prompt say "I want to see pictures of a garden with a fountain". A pre-trained image generator (ImGen) generates images by sampling its latent space. A GPR running on the image latent space prescribes latent samples $x$ and their corresponding image reconstructions $y$ which are rated by the user. Eventually, GPR would determine the optimal latent sample $x^*$ and the corresponding image that best fits the prompt $y^*$. Several images may receive similar scores from the user implying that the user satisfaction function is a staircase (similar to audio personalization); a great candidate for \name.

For the experiment, we use a pre-trained generator \cite{esser2020taming} (ImGen). We create an approximate prompt for the user's requirement (maybe too long) and generate the starting image $x_{start}$ by feeding the prompt to ImGen. The user scores this image (starting observation of $f$GPR). {\name} then optimizes the user satisfaction on ImGen's latent space $\mathcal{I}$ (user scores several images $y$ prescribed by {\name} from ImGen). Table \ref{tab:image} displays the starting image $x_{start}$ and the image that is the ``best" for the corresponding prompt after $Q=5,10,15,20,25$ (left to right) queries as determined by {\name}. For rows 1-4, {\name} quickly identifies images that fit the user prompt within $Q=25$ queries. We can observe that the starting image $x_{start}$  from the generator does not satisfy the user requirement entirely but the images from {\name} get closer and closer to the user's tastes after $Q=5,10,15,20,25$ queries. However, for rows 5\&6, we can see that the generator is doing a pretty good job of matching the prompt ($x_{start}$) and {\name} cannot seem to refine this further even with user queried information.

\clearpage

\end{document}